\documentclass[review]{elsarticle}
\graphicspath{ {./figures/} }
\usepackage{hyperref}
\usepackage{float}
\usepackage{verbatim} 
\usepackage{apalike}

\usepackage{ragged2e}  

\usepackage[ruled,vlined]{algorithm2e}

\usepackage{algpseudocode}
\usepackage{parskip}
\usepackage{subfig}
\usepackage{pifont}
\usepackage{float}

\usepackage{booktabs}
\usepackage{colortbl}
\usepackage{xcolor}
\usepackage{graphicx}

\usepackage{lastpage}
\usepackage{enumitem}
\usepackage{tikz}
\usepackage{tcolorbox}
\usepackage{multirow} 

\usepackage{amsmath}
\usepackage{amssymb}
\usepackage{fullpage}
\usepackage{float}

\restylefloat{figure}
\floatstyle{plaintop} 
\restylefloat{table}

\newenvironment{biography}[1]{%
  \noindent\textbf{#1}~~\ignorespaces
}{%
  \par\vspace{1em}
}

\journal{Expert Systems with Applications}

\biboptions{authoryear}

\begin{document}
\begin{frontmatter}

\title{BiTA: Bidirectional Gated Recurrent Unit-Transformer Aggregator in a Temporal Graph Network Framework for Alert Prediction in Computer Networks}

\author[label1]{Zahra Makki Nayeri}
\ead{zmakki@shahroodut.ac.ir}

\author[label1,cor1]{Mohsen Rezvani}
\ead{mrezvani@shahroodut.ac.ir}

\cortext[cor1]{Corresponding author: Mohsen Rezvani, 
Email: mohsen.rezvani@gmail.com}

\address[label1]{Faculty of Computer Engineering, 
Shahrood University of Technology, Shahrood, Iran}

\begin{abstract}
Proactive alert prediction in computer networks is critical for mitigating evolving cyber threats and enabling timely defensive actions. Temporal Graph Neural Networks (TGNs) provide a principled framework for modeling time-evolving interactions; however, existing TGN-based methods predominantly rely on unidirectional or single-mechanism temporal aggregation, which limits their ability to capture recursive, multi-scale temporal patterns commonly observed in real-world attack behaviors.
In this paper, we propose BiTA, a Bidirectional Gated Recurrent Unit--Transformer Aggregator for temporal graph learning. Rather than introducing a deeper or higher-capacity model, BiTA redesigns the temporal aggregation function within the TGN framework by jointly encoding bidirectional sequential dependencies and long-range contextual relations over each node’s temporal neighborhood. This aggregation strategy enables complementary temporal reasoning at different scales while preserving the original TGN memory and message-passing structure.
We evaluate BiTA on real-world alert datasets, demonstrating significant improvements in key performance metrics such as area under the curve, average precision, mean reciprocal rank, and per-category prediction accuracy when compared to state-of-the-art temporal graph models. BiTA outperforms baseline methods under both transductive and inductive settings, highlighting its robustness and generalization capabilities in dynamic network environments. BiTA is a scalable and interpretable framework for real-time cyber threat anticipation, paving the way toward more intelligent and adaptive intrusion detection systems.
\end{abstract}


\end{frontmatter}

\section{Introduction}
\label{introduction}

In modern computer networks, timely and accurate prediction of alert events is a fundamental requirement for mitigating the impact of increasingly sophisticated cyber threats. As attack campaigns become multi-stage, adaptive, and temporally correlated, traditional rule-based defense systems struggle to provide proactive and context-aware protection. This challenge has motivated the adoption of data-driven approaches, particularly Temporal Graph Neural Networks (TGNs), which offer a principled way to model the evolving interactions among network entities over time.

Temporal GNNs are well suited for capturing both structural and temporal dependencies in dynamic graphs. However, despite their success, a critical limitation lies in how historical interactions are aggregated over time. In the original TGN framework, incoming messages for each node are aggregated using simple heuristic operators such as mean or last. While computationally efficient, these aggregation schemes are non-learnable and temporally myopic: they discard the internal temporal structure of interaction sequences, treat all historical messages as equally informative, and fail to model recursive or long-range temporal dependencies. As a result, the expressive power of TGN is fundamentally constrained by a static aggregation mechanism, even though its memory update module is recurrent.

This limitation is particularly severe in cybersecurity scenarios. Network attacks often exhibit delayed, recurrent, and bidirectional temporal patterns, where early benign-looking events can only be correctly interpreted after subsequent alerts occur. Unidirectional temporal encoders, such as GRU or LSTM-based models commonly used in existing temporal aggregation schemes, are inherently unable to retrospectively refine historical representations using future context. Prior studies have shown that such models struggle to capture complex temporal dependencies in dynamic graphs \cite{wu2019graph, rossi2020temporal}, motivating alternative designs such as non-local recurrent mechanisms \cite{fu2019non}, temporal convolutions \cite{bai2018empirical}, and attention-based temporal models \cite{xu2020inductive}. Nevertheless, how to endow TGN with a learnable, expressive, and temporally aware aggregation mechanism remains an open problem.

To address this challenge, we propose BiTA, a Bidirectional GRU–Transformer Aggregator, which fundamentally redesigns the message aggregation component of TGN. Instead of relying on static operators, BiTA transforms aggregation into a learnable temporal modeling process. For each node, incoming messages are first temporally ordered and encoded using a Bidirectional GRU, enabling the model to capture fine-grained sequential dependencies in both forward and backward temporal directions. The resulting hidden representations are then processed by a Transformer encoder, which applies multi-head self-attention to selectively emphasize the most informative historical interactions and model long-range temporal dependencies. The final aggregated representation is subsequently used for memory updating within the original TGN framework.

Importantly, BiTA preserves the core architecture of TGN while addressing its most restrictive component. By replacing heuristic aggregation with a bidirectional and attention-aware temporal aggregator, the proposed framework significantly enhances the expressive capacity of TGN without sacrificing its inductive capabilities. This design allows historical node representations to be dynamically refined based on both past and future interaction context, a property that is crucial for accurate alert prediction in evolving network environments.

In practical settings such as Security Operation Centers (SOCs) and Computer Emergency Response Teams (CERTs), analysts are overwhelmed by massive volumes of heterogeneous alerts, where timely prioritization and accurate categorization are essential to prevent alert fatigue and missed threats \cite{oguntoyinbo2025mitigating, siyan2024machine, kearney2023combating, riebe2021cysecalert}. Beyond predicting whether an alert will occur, understanding its category and potential propagation path is equally critical. The proposed framework supports both alert existence prediction and fine-grained category prediction, enabling security operators to anticipate attack escalation and take informed mitigation actions.

The main contributions of this work are summarized as follows:
\begin{itemize}
\item We identify the static message aggregation mechanism of TGN as a key bottleneck and propose BiTA, a novel learnable temporal aggregation module that replaces heuristic operators such as \emph{mean} and \emph{last}.
\item We design a bidirectional and attention-aware aggregation pipeline based on BiGRU and Transformer encoders, enabling the modeling of recursive, delayed, and long-range temporal dependencies in dynamic alert streams.
\item We integrate BiTA into the TGN framework without modifying its memory update mechanism, preserving inductive generalization while significantly enhancing temporal expressiveness.
\item Extensive experiments on real-world network alert datasets demonstrate consistent and substantial improvements over state-of-the-art temporal graph models under both transductive and inductive settings.
\end{itemize}

The remainder of this paper is organized as follows. 
Section~\ref{sec:related_work} reviews the research problem and relevant prior work. 
Section~\ref{sec:model} presents the proposed model and its architecture. 
Section~\ref{sec:BiTA-Alert} describes the adaptation of the BiTA framework to the alert prediction task in computer networks. 
Section~\ref{sec:experiments} details the experimental setup and reports the evaluation results. 
Section~\ref{sec:discussion} discusses the findings, limitations, and future works of the proposed approach. 
Finally, Section~\ref{sec:conclusion} concludes the paper and outlines future research directions.

\section{Related Works}
\label{sec:related_work}
Prior work on network alert prediction has explored sequential neural models~\cite{ansari2020shallow, ansari2022gru} and graph-based representations~\cite{nayeri2024alert, nayeri2026alert}. 
This section focuses on temporal graph models, with particular emphasis on attention-based and transformer-inspired architectures for modeling dynamic interactions.

\paragraph{Temporal Graph Models Perspective}
More recently, temporal graph neural networks (TGNs) have been proposed to model time-evolving interactions by explicitly incorporating temporal dynamics into graph representations. DyRep \cite{trivedi2019dyrep} introduces a dynamic representation learning framework that jointly models node embeddings and event intensities through temporal point processes, enabling the prediction of future interactions in dynamic networks. Similarly, memory-based models such as EdgeBank \cite{poursafaei2022towards} store historical interactions to provide strong and efficient baselines for temporal link prediction, demonstrating that recency and interaction frequency play a critical role in forecasting future edges.

\paragraph{Attention-based and Transformer-inspired Temporal Graph Models Perspective}
Recent studies have explored the integration of Transformer mechanisms into temporal graph learning to enhance long-range dependency modeling and scalability. 
TIDFormer \cite{peng2025tidformer} introduces a Transformer-based architecture for temporal interaction modeling with improved efficiency, while TF-TGN \cite{huang2024retrofitting} incorporates attention mechanisms into the TGN framework to strengthen temporal representation learning. 
EasyDGL \cite{chen2024easydgl} focuses on scalable and modular temporal graph processing, enabling efficient deployment of complex models, and TransformerG2G \cite{varghese2024transformerg2g} extends Transformer-based message passing to graph-to-graph interaction settings.
Despite their effectiveness, these approaches primarily rely on either attention-based aggregation or unidirectional temporal encoding, whereas BiTA explicitly combines bidirectional temporal modeling with learnable recursive aggregation tailored for alert prediction.

\section{Model Architecture}
\label{sec:model}
Before describing the proposed architecture, we first define the temporal graph formulation used in our framework.

\subsection{Problem Formulation}

We represent the system as a temporal graph:
\[
G_t = (V, E_t),
\]
where:
\begin{itemize}
    \item $V = \{v_1, v_2, \dots, v_N\}$ is the set of nodes, each representing an entity in the system (e.g., a host, a process, or a user).
    \item $E_t \subseteq V \times V$ is the set of temporal edges at time $t$, where each edge corresponds to an interaction event between two nodes.
\end{itemize}

Each temporal interaction event is represented as:
\[
e_{uv}^{(t)} = (u, v, \mathbf{x}_{uv}, t, c_{uv}),
\]
where:
\begin{itemize}
    \item $u, v \in V$ are the interacting nodes.
    \item $\mathbf{x}_{uv}$ denotes the edge features (e.g., communication attributes).
    \item $t$ is the timestamp of the interaction.
    \item $c_{uv}$ is the edge category or label associated with the interaction.
\end{itemize}

\paragraph{Learning Objective.}  
Given a sequence of past interaction events 
\[
\mathcal{E}^{(<t)} = \{ e_{uv}^{(t_1)}, e_{uv}^{(t_2)}, \dots, e_{uv}^{(t_{k})} \mid t_1 < t_2 < \dots < t_k < t \},
\] 
our model aims to simultaneously predict:
\begin{enumerate}
    \item \textbf{Link prediction:} Whether a future edge between two nodes $(i,j)$ at time $t$ will occur.
    \item \textbf{Category prediction:} The category $c_{ij}$ of the edge if it occurs.
\end{enumerate}

\subsection{Proposed Method}

The proposed BiGRU-Transformer Aggregator (BiTA) is designed to enhance temporal message aggregation by incorporating both past and future contextual information. Traditional TGNs typically aggregate messages from neighboring nodes in a sequential and causal manner, using only past messages \cite{rossi2020temporal}. However, this approach may lead to incomplete or myopic representations of temporal dynamics, especially in complex scenarios such as alert prediction in cyber networks.

To address this limitation, BiTA introduces a bidirectional gated recurrent unit (GRU) transformer-based aggregation mechanism. Specifically, for each node, BiTA constructs a sequence of its incoming messages ordered temporally. This sequence is then passed through a bidirectional GRU transformer encoder, which allows each message to attend not only to preceding messages but also to succeeding ones. As a result, BiTA can capture rich temporal dependencies and long-range interactions that are otherwise inaccessible in standard forward-only models.

Unlike causal transformers, which restrict attention flow to prior tokens to preserve temporal consistency, BiTA is applied only at the aggregation stage and not during prediction. Thus, it does not violate causal constraints of real-time inference. In inductive and transductive settings, BiTA consistently improves representation quality by producing temporally enriched embeddings before memory updates and prediction steps.

The advantages of BiTA are twofold: (1) improved awareness of temporal context around each interaction, and (2) the ability to model latent temporal patterns that span across multiple events. These capabilities are particularly beneficial in scenarios involving stealthy or delayed cyber attacks, where current alerts are better understood in the context of both past and future interactions.

\begin{table}[H]
\centering
\caption{Notation summary used throughout the paper.}
\label{tab:notation}
\begin{tabular}{ll}
\toprule
\textbf{Symbol} & \textbf{Description} \\
\midrule
$u, v, i, j$ & Node indices in the temporal graph \\
$e_{uv}$ & Feature vector of the edge $(u,v)$ \\
$t$ & Timestamp of an interaction event \\
$(u, v, e_{uv}, t)$ & Temporal interaction tuple \\
$m_{uv}^{(t)}$ & Raw message constructed from interaction $(u,v)$ at time $t$ \\
$\text{TE}(t)$ & Time encoding function \\
$\tilde{m}_{uv}^{(t)}$ & Time-enriched message \\
$\mathcal{M}_v^{(t)}$ & Sequence of messages received by node $v$ up to time $t$ \\
$\mathbf{x}_{ij,t}$ & Time-encoded input for edge $(i,j)$ at time $t$ \\
$\mathbf{H}_{ij,t}^{\text{temp}}$ & BiGRU hidden states for edge $(i,j)$ at time $t$ \\
$\mathbf{z}_{ij,t}^{\text{temp}}$ & Temporal embedding of edge $(i,j)$ \\
$\mathbf{e}_{ij,t}$ & Projected embedding (input to Transformer) \\
$\mathbf{H}_{t}^{\text{ctx}}$ & Transformer contextual hidden states at time $t$ \\
$\mathbf{z}_{ij,t}^{\text{ctx}}$ & Final contextual embedding of edge $(i,j)$ \\
$\hat{y}_{ij,t}^{\text{link}}$ & Predicted probability of link existence \\
$\hat{y}_{ij,t}^{\text{cat}}$ & Predicted edge category distribution \\
$\mathcal{L}_{\text{link}}$ & Link prediction loss (binary cross-entropy) \\
$\mathcal{L}_{\text{cat}}$ & Category prediction loss (focal loss) \\
$\mathcal{L}$ & Final joint loss function \\
\bottomrule
\end{tabular}
\end{table}

\begin{figure}[H]
\centering
\includegraphics[width=\linewidth]{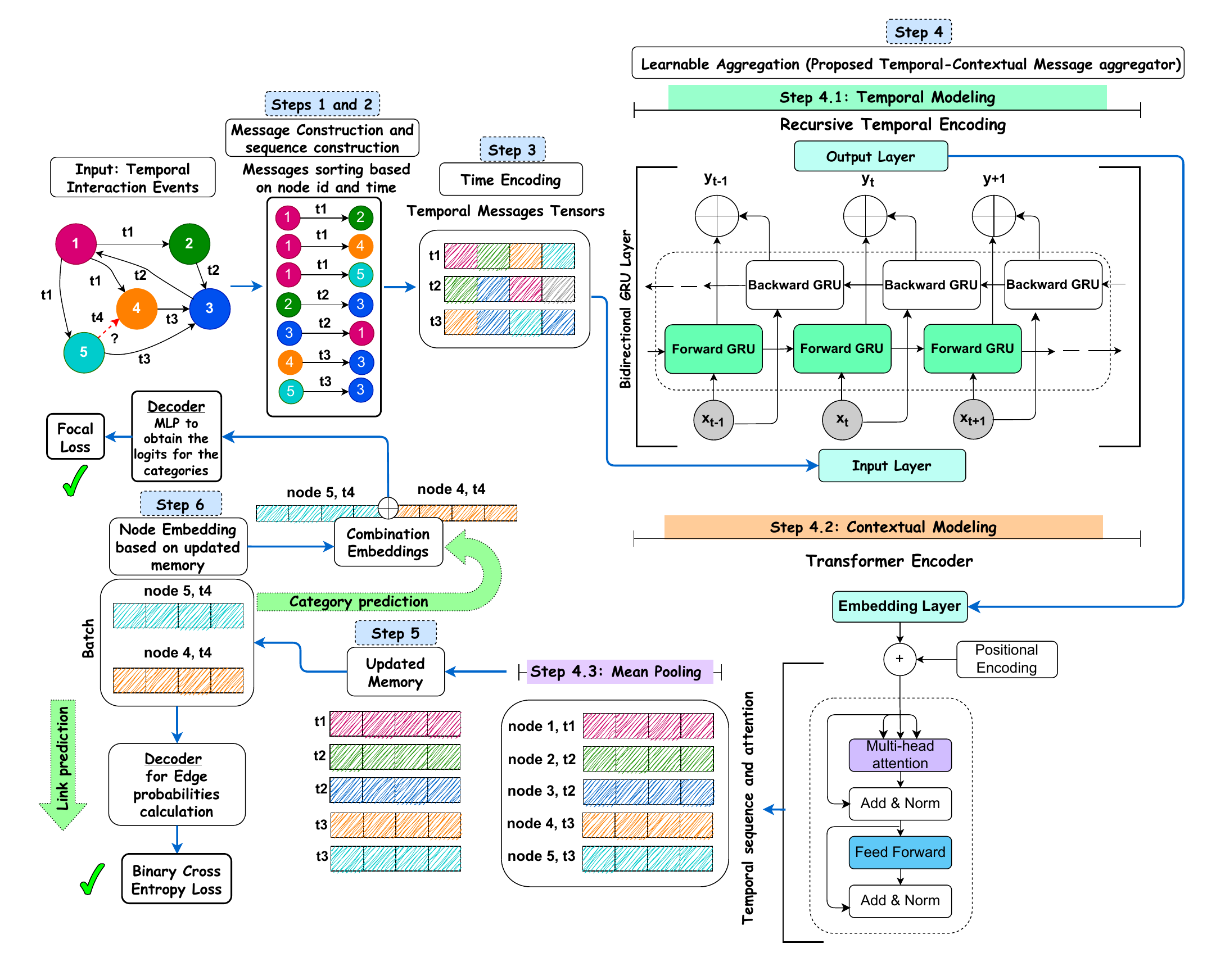} 
\caption{Overview of the proposed BiTA module integrated into TGN.
Unlike conventional TGNs where message aggregation is performed using heuristic operators (e.g., mean or last),
BiTA realizes aggregation through unified temporal--contextual representation learning.
Incoming messages are jointly encoded via a BiGRU and a Transformer to model bidirectional temporal dependencies
and global contextual interactions.
The final mean pooling operation serves only as a readout mechanism, not as the aggregation itself.}
\label{BiTA_Flow}
\end{figure}

\textbf{\textit{BiTA Architectural Overview}}

The BiTA framework processes temporal graph interactions through a sequential pipeline consisting of message construction, temporal encoding, bidirectional sequence modeling, and contextual attention. Unlike conventional TGN aggregators that rely on heuristic pooling operations (e.g., last or mean), BiTA performs aggregation through learned representations that encode the internal structure of message sequences.

As illustrated in Figure \ref{BiTA_Flow}, the architecture of the BiTA can be described in the following sequential components:

\paragraph{\textbf{Input Representation:}}

Each temporal interaction between nodes $u$ and $v$ at time $t$ is represented as a tuple $(u, v, e_{uv}, t)$, where $e_{uv} \in \mathbb{R}^{d_e}$ denotes the edge feature vector. The complete notation is summarized in Table~\ref{tab:notation}.

\paragraph {\textbf{Step 1. Message Construction:}}

For each interaction event, we construct a temporal message using a learnable message function:
\begin{equation}
    m_{uv}^{(t)} = \text{MSG}(u, v, e_{uv}, t),
    \label{eq:message_construction}
\end{equation}
{where $\text{MSG}: \mathcal{V} \times \mathcal{V} \times \mathbb{R}^{d_e} \times \mathbb{R} \rightarrow \mathbb{R}^{d_m}$ maps interaction tuples to message embeddings of dimension $d_m$.

\paragraph{\textbf{Step 2. Temporal Message Sequence Formation:}}
For each target node $v$, all incoming messages up to the current time $t$ are collected and sorted chronologically to form an ordered sequence:
\begin{equation}
    \mathcal{M}_v^{(t)} = [m_{v}^{(t_1)}, m_{v}^{(t_2)}, \ldots, m_{v}^{(t_n)}], \quad t_1 < t_2 < \cdots < t_n \leq t,
    \label{eq:message_sequence}
\end{equation}
where $n = |\mathcal{M}_v^{(t)}|$ denotes the number of messages received by node $v$ up to time $t$.
\paragraph{\textbf{Step 3. Temporal Encoding:}}
To explicitly incorporate timing information, each message is augmented with a continuous time encoding:
\begin{equation}
    \mathbf{x}_{ij,t} = \mathbf{m}_{ij,t} + \text{TimeEnc}(t),
    \label{eq:time_encoding}
\end{equation}
where $\text{TimeEnc}: \mathbb{R} \rightarrow \mathbb{R}^{d_m}$ is a learnable temporal encoding function (e.g., Bochner's time encoding~\cite{xu2020inductive}).
\paragraph{\textbf{Step 4. Temporal-Contextual Representation Learning as Aggregation:}}
Given the temporally ordered message sequence $\mathcal{M}_v^{(t)}$, BiTA performs aggregation by learning a sequence-level representation that captures both bidirectional temporal dependencies and global contextual interactions. The core innovation of BiTA lies in its three-stage representation learning strategy. Rather than applying static pooling operations, BiTA learns sequence-level representations that encode both temporal dynamics and contextual dependencies through the following substeps:

\paragraph{\textbf{Step 4.1. Temporal Modeling (BiGRU):}}
To capture sequential dependencies in the message stream, we employ a BiGRU. The BiGRU processes the temporally ordered message sequence in both forward and backward directions:
\begin{equation}
    \mathbf{H}_{ij,t}^{\text{temp}} = \text{BiGRU}(\mathbf{x}_{ij,t}),
    \label{eq:bigru_encoding}
\end{equation}
where $\mathbf{x}_{ij,t}$ represents the time-encoded input sequence (from Equation~\ref{eq:time_encoding}), and $\mathbf{H}_{ij,t}^{\text{temp}}$ denotes the sequence of BiGRU hidden states encoding bidirectional temporal context.
The temporal representation for edge $(i,j)$ at time $t$ is then extracted as:
\begin{equation}
    \mathbf{z}_{ij,t}^{\text{temp}} = \mathbf{H}_{ij,t}^{\text{temp}}[-1],
    \label{eq:temporal_representation}
\end{equation}
where $[-1]$ denotes the final hidden state, capturing the complete temporal context from both forward and backward passes.
\paragraph{\textbf{Step 4.2. Contextual Modeling (Transformer):}}
To enable the model to selectively attend to relevant past interactions regardless of their temporal distance, we apply a Transformer encoder with multi-head self-attention. First, the temporal representations are projected into the Transformer's embedding space:
\begin{equation}
    \mathbf{e}_{ij,t} = W_e \mathbf{z}_{ij,t}^{\text{temp}} + b_e,
    \label{eq:transformer_projection}
\end{equation}
where $W_e \in \mathbb{R}^{d_{\text{trans}} \times 2d_h}$ and $b_e \in \mathbb{R}^{d_{\text{trans}}}$ are learnable parameters.
The projected embeddings are then processed through the Transformer encoder across all edges at time $t$:
\begin{equation}
    \mathbf{H}_{t}^{\text{ctx}} = \text{Transformer}(\{\mathbf{e}_{ij,t} \mid (i,j) \in E_t\}),
    \label{eq:transformer_encoding}
\end{equation}
where $E_t$ denotes the set of edges with interactions up to time $t$. The Transformer produces contextualized representations through self-attention mechanisms, and the final contextual embedding for edge $(i,j)$ is obtained as:
\begin{equation}
    \mathbf{z}_{ij,t}^{\text{ctx}} \in \mathbf{H}_{t}^{\text{ctx}},
    \label{eq:contextual_representation}
\end{equation}
where $\mathbf{z}_{ij,t}^{\text{ctx}} \in \mathbb{R}^{d_{\text{trans}}}$ encodes both temporal and contextual information.
\paragraph{\textbf{Step 4.3. Mean Pooling:}}
To obtain a fixed-size aggregate representation compatible with the TGN memory update interface, we apply mean pooling as a readout function. For a node $v$ that receives messages from multiple edges, the aggregated message is computed as:

\begin{equation}
    \bar{h}_v^{(t)} = \frac{1}{|\mathcal{N}_v(t)|} \sum_{(i,j) \in \mathcal{N}_v(t)} \mathbf{z}_{ij,t}^{\text{ctx}},
    \label{eq:readout_pooling}
\end{equation}
where $\mathcal{N}_v(t)$ denotes the set of edges incident to node $v$ up to time $t$, and $|\mathcal{N}_v(t)|$ is the number of such edges.

\textbf{\textit{Remark on Aggregation versus Readout:}} It is crucial to emphasize that this mean pooling operation in Step 4.3 serves solely as a \emph{readout mechanism} to summarize the learned representations. The aggregation itself is performed through the learnable temporal and contextual encoding stages in Steps 4.1 and 4.2 (BiGRU and Transformer). This fundamentally differs from standard TGN aggregation, where pooling operations directly define the aggregation behavior without intermediate representation learning.

\textbf{\textit{Comparison with Standard TGN Aggregation:}}
In the original TGN framework~\cite{rossi2020temporal}, message aggregation is defined by heuristic operators applied directly to raw messages:
\begin{equation}
    \bar{m}_v^{(t)} =
    \begin{cases}
    m_v^{(t_n)} & \text{(last aggregation)} \\[0.5em]
    \displaystyle\frac{1}{n} \sum_{k=1}^{n} m_v^{(t_k)} & \text{(mean aggregation)}
    \end{cases}
    \label{eq:standard_tgn_aggregation}
\end{equation}
In this paradigm, the aggregation behavior is entirely determined by the choice of pooling operator, with no learned representation of temporal dependencies or contextual relationships.
In contrast, BiTA decouples aggregation from pooling through a two-level architecture:
\begin{enumerate}
    \item \textbf{Aggregation Layer:} Temporal and contextual dependencies are learned through BiGRU and Transformer encoders, producing sequence-level representations.
    \item \textbf{Readout Layer:} Mean pooling is applied solely to summarize the already-aggregated representations into a single vector.
\end{enumerate}
This design enables BiTA to capture bidirectional temporal dependencies and long-range interactions that are fundamentally inaccessible to memoryless heuristic aggregation schemes.

\paragraph{\textbf{Step 5. Memory Update (Optional):}}
If a memory module is employed, the memory state is updated using the aggregated representation:
\begin{equation}
    \text{Mem}_v^{(t)} = \text{MemUpd}(\text{Mem}_v^{(t-1)}, \bar{h}_v^{(t)}),
    \label{eq:memory_update}
\end{equation}
where $t-1$ denotes the previous timestamp, and $\text{MemUpd}$ is a recurrent update function (e.g., GRU or RNN).
\paragraph{\textbf{Step 6. Node Embedding Update:}}
Node embeddings are updated by combining previous states with new aggregated messages:
\begin{equation}
    \mathbf{h}_i(t) = \text{Update}(\mathbf{h}_i(t-\Delta t), \mathbf{m}_i(t)),
    \label{eq:node_embedding_update}
\end{equation}
where $\Delta t$ represents the time elapsed since the last update, and $\text{Update}$ is a learnable function that integrates temporal information into node representations.

\paragraph{\textbf{Joint Prediction Framework:}}
After obtaining contextualized edge embeddings $\mathbf{z}_{ij,t}^{\text{ctx}}$, BiTA performs two prediction tasks simultaneously:

\paragraph{\textbf{Category Prediction:}}
A multi-class classifier predicts the category of each interaction:
\begin{equation}
    \hat{y}_{ij,t}^{\text{cat}} = \text{softmax}(W_c \mathbf{z}_{ij,t}^{\text{ctx}} + b_c),
    \label{eq:category_prediction}
\end{equation}
where $W_c \in \mathbb{R}^{K \times d_{\text{trans}}}$ and $b_c \in \mathbb{R}^{K}$ are learnable parameters, with $K$ denoting the number of categories.

To address class imbalance, we employ the focal loss:
\begin{equation}
    \mathcal{L}_{\text{cat}} = - \sum_{k=1}^K \alpha_k (1 - \hat{y}_{ij,t,k}^{\text{cat}})^{\gamma} y_{ij,t,k} \log \hat{y}_{ij,t,k}^{\text{cat}},
    \label{eq:focal_loss}
\end{equation}
where $\alpha_k$ are class weights, $\gamma$ is the focusing parameter, and $y_{ij,t,k}$ is the one-hot encoded ground truth.

\paragraph{\textbf{Link Prediction:}}
A binary classifier predicts whether a future interaction will occur:
\begin{equation}
    \hat{y}_{ij,t}^{\text{link}} = \sigma(W_\ell \mathbf{z}_{ij,t}^{\text{ctx}} + b_\ell),
    \label{eq:link_prediction}
\end{equation}
where $W_\ell \in \mathbb{R}^{1 \times d_{\text{trans}}}$, $b_\ell \in \mathbb{R}$, and $\sigma$ denotes the sigmoid function.

The corresponding binary cross-entropy loss is:
\begin{equation}
    \mathcal{L}_{\text{link}} = - \left( y_{ij,t} \log \hat{y}_{ij,t}^{\text{link}} + (1 - y_{ij,t}) \log (1 - \hat{y}_{ij,t}^{\text{link}}) \right).
    \label{eq:bce_loss}
\end{equation}

\paragraph{\textbf{Joint Optimization Objective:}}
The final training objective combines both tasks:
\begin{equation}
    \mathcal{L} = \mathcal{L}_{\text{link}} + \lambda \mathcal{L}_{\text{cat}},
    \label{eq:joint_loss}
\end{equation}
where $\lambda$ is a hyperparameter balancing the two objectives (set to 1 in our experiments).

\paragraph{\textbf{Algorithmic Implementation:}}
Algorithm~\ref{alg:bita_aggregation} provides the complete procedural implementation of the BiTA aggregation mechanism.

\paragraph{\textbf{Computational Complexity Analysis:}}
The computational complexity of BiTA is dominated by two components:

\begin{itemize}
    \item \textbf{BiGRU Encoding:} $\mathcal{O}(n \cdot d_m \cdot d_h)$, where $n$ is the sequence length, $d_m$ is the message dimension, and $d_h$ is the hidden dimension.
    
    \item \textbf{Transformer Encoding:} $\mathcal{O}(n^2 \cdot d_{\text{trans}})$, dominated by the self-attention mechanism's quadratic complexity in sequence length.
\end{itemize}

For typical interaction sequences where $n \ll |\mathcal{V}|$, the overhead introduced by BiTA remains manageable while providing substantial improvements in representation quality.

\begin{algorithm}[H]
\caption{BiTA Message Aggregation via BiGRU-Transformer}
\label{alg:bita_aggregation}
\KwIn{
    Temporal interaction batch $\mathcal{I} = \{(u_i, v_i, t_i, e_i)\}_{i=1}^{N}$ \\
    Node memory states $\{\text{Mem}_u\}_{u \in \mathcal{V}}$ \\
    Message function $\text{MSG}(\cdot)$ \\
    Time encoding function $\text{TimeEnc}(\cdot)$ \\
    BiGRU encoder $f_{\text{BiGRU}}$, Transformer encoder $f_{\text{Trans}}$
}
\KwOut{
    Aggregated hidden states $\{\bar{h}_v\}_{v \in \mathcal{V}}$ for memory update
}

\tcp{Step 1: Message Construction}
\ForEach{interaction $(u_i, v_i, t_i, e_i) \in \mathcal{I}$}{
    $m_{u_iv_i}^{(t_i)} \gets \text{MSG}(u_i, v_i, e_i, t_i)$ \\
    Assign message to edge $(u_i, v_i)$ at time $t_i$
}

\tcp{Step 2 \& 3: Message Sequence Formation and Time Encoding}
\ForEach{edge $(i,j)$ with interactions up to time $t$}{
    Gather and sort messages: $\mathcal{M}_{ij}^{(t)} = [m_{ij}^{(t_1)}, m_{ij}^{(t_2)}, \ldots, m_{ij}^{(t_k)}]$ \\
    Apply time encoding: $\mathbf{x}_{ij,t} \gets \mathbf{m}_{ij,t} + \text{TimeEnc}(t)$
}

\tcp{Step 4.1: Temporal Modeling (BiGRU)}
\ForEach{edge $(i,j)$ at time $t$}{
    $\mathbf{H}_{ij,t}^{\text{temp}} \gets f_{\text{BiGRU}}(\mathbf{x}_{ij,t})$ \tcp*{Bidirectional temporal encoding}
    $\mathbf{z}_{ij,t}^{\text{temp}} \gets \mathbf{H}_{ij,t}^{\text{temp}}[-1]$ \tcp*{Extract final hidden state}
}

\tcp{Step 4.2: Contextual Modeling (Transformer)}
\ForEach{edge $(i,j)$ at time $t$}{
    $\mathbf{e}_{ij,t} \gets W_e \mathbf{z}_{ij,t}^{\text{temp}} + b_e$ \tcp*{Project to Transformer space}
}
$\mathbf{H}_{t}^{\text{ctx}} \gets f_{\text{Trans}}(\{\mathbf{e}_{ij,t} \mid (i,j) \in E_t\})$ \tcp*{Self-attention across edges}
\ForEach{edge $(i,j) \in E_t$}{
    Extract $\mathbf{z}_{ij,t}^{\text{ctx}}$ from $\mathbf{H}_{t}^{\text{ctx}}$
}

\tcp{Step 4.3: Readout via Mean Pooling}
\ForEach{node $v \in \mathcal{V}$ receiving messages}{
    Gather edges incident to $v$: $\mathcal{N}_v(t) = \{(i,j) \mid j = v \text{ or } i = v\}$ \\
    $\bar{h}_v^{(t)} \gets \frac{1}{|\mathcal{N}_v(t)|} \sum_{(i,j) \in \mathcal{N}_v(t)} \mathbf{z}_{ij,t}^{\text{ctx}}$ \tcp*{Mean pooling}
}

\Return{$\{\bar{h}_v^{(t)}\}$ for memory and embedding updates}
\end{algorithm}

\section{BiTA for Alert Prediction}
\label{sec:BiTA-Alert}

\subsection{System Model}

We consider a cyber-security monitoring system that observes a stream of time-stamped alerts generated from network traffic. Each alert corresponds to a directed interaction between two entities, representing an attack attempt from a source IP address (attacker) to a destination IP address (victim).

The system models these interactions as a temporal graph, where nodes denote IP addresses and edges represent attack events enriched with categorical and numerical features (e.g., protocol type, port, and alert category). Interactions arrive sequentially over time and are processed in chronological order.

Given the historical sequence of interactions observed up to time $t$, the objective of the system is to predict future attacker--victim interactions and, when applicable, their associated attack categories. The system operates under both transductive and inductive settings, allowing prediction over previously seen as well as unseen nodes.

\subsection{Threat Model and Assumptions}

To enable focused and analytically tractable modeling of temporal attacker--victim interactions, we adopt a bounded threat model consistent with the characteristics of the evaluated datasets. The following assumptions define the adversarial setting and operational scope considered in this study:
\begin{enumerate}
\item \textbf{Limited-Adversary Knowledge:}
The adversary is assumed to have no access to the prediction framework, including the model architecture, learned parameters, or inference outcomes. As a result, adversarial behavior cannot be adapted to exploit the learning process, ensuring that observed patterns reflect inherent attack dynamics rather than strategic manipulation.

\item \textbf{Unidirectional Attack Flows:}
Each interaction corresponds to a one-way attack from a source (attacker) to a destination (victim). Bidirectional or retaliatory behaviors are not explicitly modeled. This abstraction simplifies edge semantics while remaining consistent with the dataset annotation scheme.

\item \textbf{Mutually Exclusive Roles per Event:}
Within a single interaction, an IP address assumes a unique role, either attacker or victim, but never both simultaneously. This dataset-driven constraint avoids ambiguity in role assignment and provides clear supervision signals during training.

\item \textbf{Closed-Set Attack Classes:}
The prediction task is restricted to attack categories explicitly labeled in the dataset. The identification of previously unseen or zero-day attacks lies outside the scope of this work and is deferred to future studies on open-set or zero-shot learning.

\item \textbf{Bounded Entity Universe with Inductive Generalization:}
The model operates over the set of IP addresses observed in the dataset and is evaluated under both transductive and inductive settings. In the inductive case, the framework demonstrates the ability to generalize to nodes not seen during training. Nevertheless, the prediction space remains bounded to dataset-defined entities due to labeling and evaluation constraints. Extending the model to fully open-world scenarios with dynamically emerging entities is left for future work.
\end{enumerate}

These assumptions are not intended to capture the full complexity of real-world cyber adversaries. Instead, they are designed to isolate the fundamental challenge of learning temporal patterns in dynamic interaction graphs. Future extensions may relax these constraints by incorporating adaptive adversaries, overlapping node roles, or open-world learning mechanisms.

\subsection{Design Goals}

Based on the system and threat models described above, the design of BiTA is guided by the following objectives:

\begin{itemize}
    \item \textbf{Temporal Dependency Modeling:}
    Effectively capture both short-term and long-range temporal dependencies in attacker--victim interactions.

    \item \textbf{Expressive Message Aggregation:}
    Go beyond heuristic aggregation operators (e.g., mean or last) by learning aggregation functions that preserve temporal order and contextual information.

    \item \textbf{Context-Aware Interaction Modeling:}
    Enable interactions to influence each other through global contextual reasoning over historical message sequences.

    \item \textbf{Inductive Generalization:}
    Support prediction for previously unseen nodes without retraining, which is critical in dynamic network environments.

    \item \textbf{Scalability to Streaming Data:}
    Operate efficiently on large-scale temporal interaction streams typical of real-world cyber-security systems.
\end{itemize}

\subsection{Alert Data}
For clarity, Table~\ref{Alert_Table1} and Figure~\ref{sample} present a representative example of the alert records and their corresponding graph-based representation. Each alert record contains several key features that characterize the temporal and semantic properties of network interactions:

\begin{itemize}
\item \textbf{Detect Time}: Timestamp indicating when the alert was generated, enabling temporal ordering of events.
\item \textbf{Flow Count}: Traffic intensity metric representing the volume of network flows associated with the alert.
\item \textbf{Source/Target IP}: IP addresses of the communicating entities, defining the nodes in the temporal graph.
\item \textbf{Port and Protocol}: Service-level identifiers specifying the communication channel and transport protocol.
\item \textbf{Category}: Attack classification label (e.g., Reconnaissance, Availability DoS, DDoS) characterizing the threat type.
\end{itemize}

\begin{table}[H]
\centering
\caption{A sample of typical alert data.} 
\label{Alert_Table1}
\scriptsize
\renewcommand{\arraystretch}{1.1} 
\setlength{\tabcolsep}{4pt} 
\scriptsize 
\begin{tabular}{
    >{\centering\arraybackslash}p{4cm} 
    >{\centering\arraybackslash}p{1.5cm} 
    >{\centering\arraybackslash}p{1.8cm} 
    >{\centering\arraybackslash}p{1.8cm} 
    >{\centering\arraybackslash}p{1cm} 
    >{\centering\arraybackslash}p{1.2cm} 
    >{\centering\arraybackslash}p{2cm}
}
\toprule
\textbf{Detect Time} & \textbf{Flow Count} & \textbf{Source IP} & \textbf{Target IP} & \textbf{Port} & \textbf{Protocol} & \textbf{Category} \\
\midrule
2019-03-11T00:05:00+02:00 & 17094 & 185.192.59.136 & 142.252.135.136 & 22 & TCP & Recon scan \\
2019-03-12T00:45:00+02:00 & 5113 & 78.234.46.141 & 142.252.32.63 & 443 & TCP & Availability Dos \\
2019-03-14T00:25:00+02:00 & 39 & 78.234.46.141 & 142.252.32.63 & 22 & UDP & Anomaly.Traffic \\

\bottomrule
\end{tabular}
\end{table}

\begin{figure}[H]
\centering
\includegraphics[width=5in]{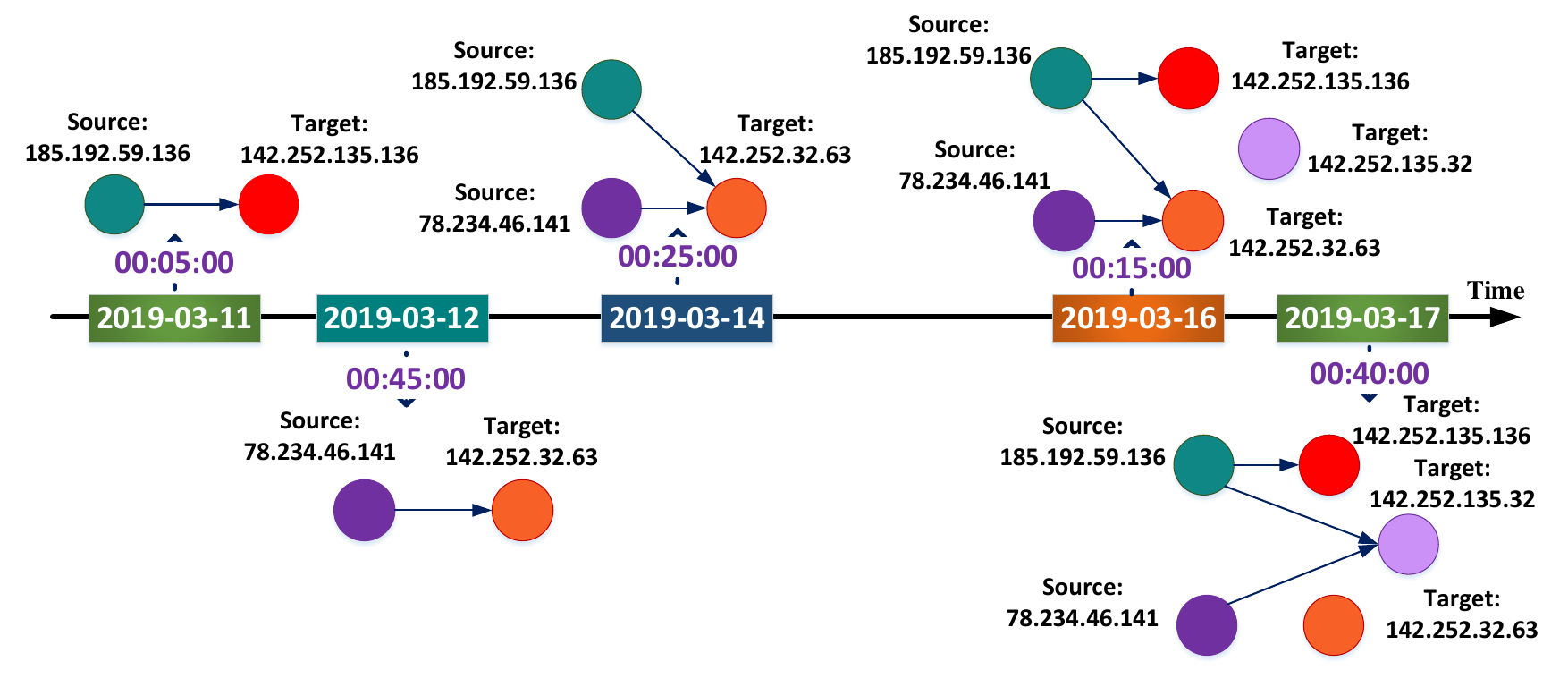}
\caption{Sample of a graph-based representation of alert data.}
\label{sample}
\end{figure}

These features collectively enable the construction of a temporal graph representation where nodes represent network entities (IP addresses), edges represent alert interactions, and temporal/semantic attributes encode the dynamic evolution of security events. Detailed dataset characteristics, preprocessing procedures, and experimental configurations are presented in Section~\ref{sec:experiments}.

\subsection{Graph Modeling}

\textbf{\textit{Definition 1. Attack Graph:}} 
The attack graph \( G = (V, E) \) is constructed from alert records. The node set \( V \) comprises attacker IPs and victim IPs, with no connections allowed within the same type (i.e., no attacker-to-attacker or victim-to-victim links). 
Directed edge \( e(u, v) \) is created from an attacker node \( u \) to a victim node \( v \) when an attack is observed, thus capturing the asymmetry of attacks where interactions are initiated by attackers. 

To enrich predictive capabilities, nodes and edges are assigned feature vectors. Node features include targeted port information, reflecting services or vulnerabilities exploited in attacks.
Edge features capture the protocol, attack type, and interaction intensity between attacker and victim.

The graph presents the following properties:

\textbf{\textit{1. Bipartite Graph:}}  
A graph \( G = (V, E) \) is bipartite since \( V \) can be partitioned into two disjoint sets \( V_A \) (attackers) and \( V_V \) (victims), such that:
\[
E \subseteq \{(u, v) \mid u \in V_A, v \in V_V\}
\]
\( V_A \) contains nodes representing attackers launching alerts such as Reconscan, Anomaly Traffic, Availability DoS, and Availability DDoS.
\( V_V \) contains nodes representing victims targeted by these alerts.
%
%
\textbf{\textit{2. Multigraph:}}  
The graph \( G = (V, E) \) is a multigraph since multiple edges are allowed between the same attacker \( u \in V_A \) and victim \( v \in V_V \). Each edge represents a unique type of alert or attack instance. Formally:
\[
E = \{(u, v, i) \mid u \in V_A, v \in V_V, i \in I\}
\]
where \( i \) is the identifier of the alert type (e.g., Reconscan, DoS). For example:
\[
(A_1 \to V_1, Reconscan), \quad (A_1 \to V_1, DoS)
\]
\textbf{\textit{3. Dynamic Graph:}}
A dynamic graph \( G_t = (V_t, E_t) \) is time-variant, where the set of nodes \( V_t \) and edges \( E_t \) evolve over time based on alert timestamps. Formally:
\[
E_t = \{(u, v, i, t) \mid u \in V_A, v \in V_V, i \in I, t \in T\}
\]
An edge \( (A_1, V_1, DoS, t_1) \) indicates a DoS attack from \( A_1 \) to \( V_1 \) at time \( t_1 \).
The graph evolves dynamically as new alerts are added with their respective timestamps.
\begin{figure}[H]
\centering
\includegraphics[width=4in]{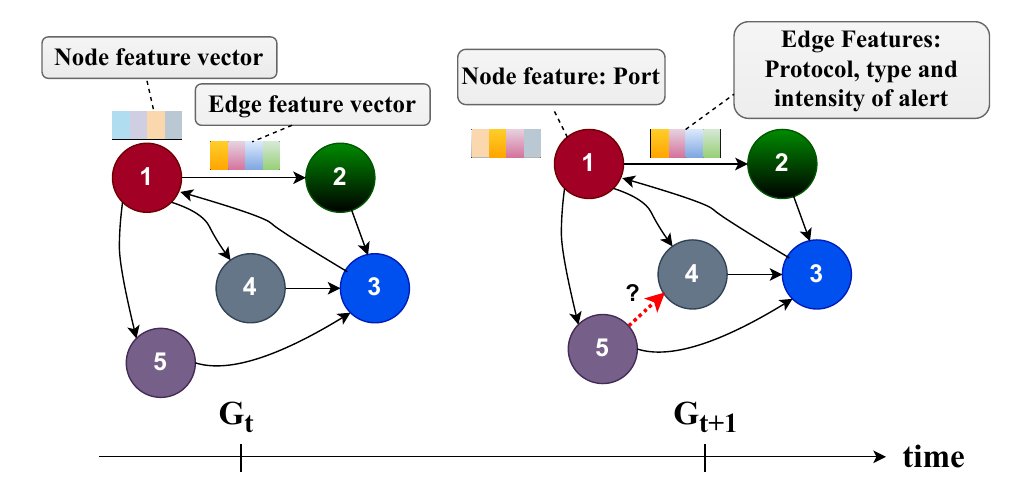}
\caption{The graph evolution in time.}
\label{graph evolution}
\end{figure}

\subsection{Transform Attack Prediction to Link Prediction}
As we defined, the attack prediction task can transform into a temporal link prediction problem in an attack graph. Specifically, given the state of the attack graph \( G_t \) at time \( t \), the goal is to predict future edges in \( G_{t+1} \), indicating potential future attacks. Figure~\ref{graph evolution} illustrates this process, where a red edge in \( G_{t+1} \) represents a predicted attack derived from the previous structure in \( G_t \).

\subsection{Example of BiTA framework}
Consider a small temporal graph with three nodes involved in network alerts. The interaction events, including attacker, victim, edge features, time, and attack category, are summarized in Table~\ref{tab:example_attacks}.
\begin{table}[H]
\centering
\caption{Example temporal interactions with attacker, victim, and attack category.}
\label{tab:example_attacks}
\begin{tabular}{cccccc}
\toprule
Event & Attacker & Victim & Time & Edge Feature & Attack Category \\
\midrule
1 & A & B & $t_1$ & 0.2 & DoS \\
2 & C & D & $t_2$ & 0.5 & Scan \\
3 & A & D & $t_3$ & ? & ? \\
\bottomrule
\end{tabular}
\end{table}
\begin{figure}[H]
\centering
\vspace{5ex}%
\includegraphics[width=7in]{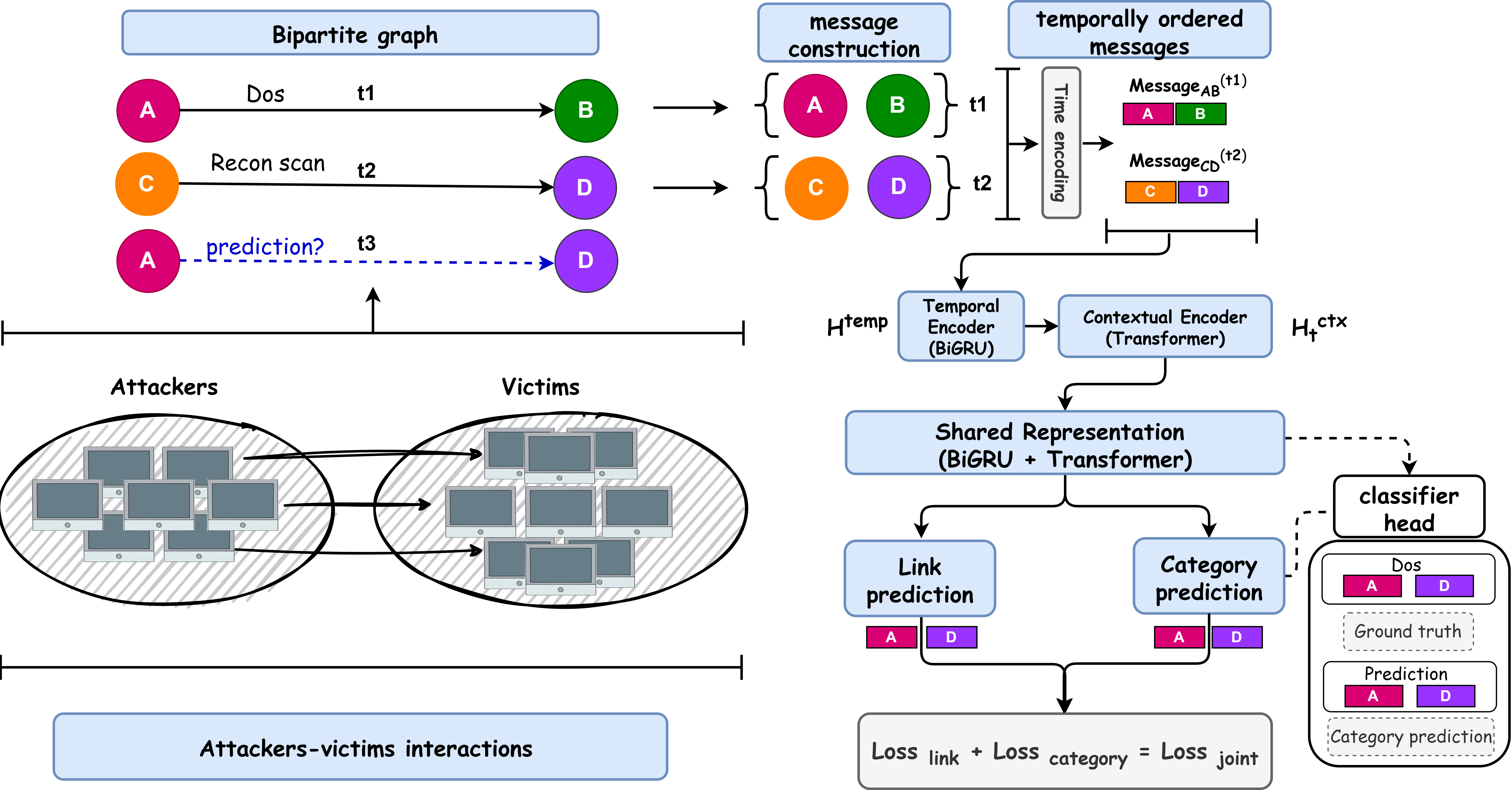}
\caption{Example of the BiTA framework for link prediction and category prediction}
\label{example}
\end{figure}

\paragraph{Step 1: Message Construction}
Each interaction is converted into a message:
\[
m_{AB}^{(t_1)} = \text{MSG}(\text{Attacker}=A, \text{Victim}=B, e_{AB}=0.2, t_1), 
\]
\[
m_{CD}^{(t_2)} = \text{MSG}(\text{Attacker}=C, \text{Victim}=D, e_{CD}=0.5, t_2)
\]
\paragraph{Step 2: Time Encoding}
Messages are enriched with temporal encoding:
\[
\tilde{m}_{AB}^{(t_1)} = m_{AB}^{(t_1)} + \text{TE}(t_1), \quad
\tilde{m}_{CD}^{(t_2)} = m_{CD}^{(t_2)} + \text{TE}(t_2)
\]
\paragraph{Step 3: BiGRU Temporal Encoding}
The temporally ordered messages are passed through a BiGRU:
\[
\mathbf{H}_{v}^{\text{temp}} = \text{BiGRU}([\tilde{m}_{AB}^{(t_1)}, \tilde{m}_{CD}^{(t_2)}])
\]
\paragraph{Step 4: Transformer Contextual Encoding}
The BiGRU outputs are processed by a Transformer to capture long-range dependencies:
\[
\mathbf{H}_{t}^{\text{ctx}} = \text{Transformer}(\mathbf{H}_{v}^{\text{temp}})
\]
\paragraph{Step 5: Joint Prediction}
Finally, the model predicts both link existence and attack category:
\[
\hat{y}_{AD}^{\text{link}} = \sigma(W_\ell \mathbf{z}_{AD,t_3}^{\text{ctx}} + b_\ell), \quad
\hat{y}_{AD}^{\text{cat}} = \text{softmax}(W_c \mathbf{z}_{AD,t_3}^{\text{ctx}} + b_c)
\]
This example demonstrates how the model jointly considers temporal and contextual information to predict potential future interactions and classify their attack categories. To provide a clearer intuition, Figure~\ref{example} visualizes the example in the form of a graph.

\subsection{Causality and Real-Time Inference}
\label{subsec:causality}

Strict causality is a fundamental requirement for temporal graph learning, ensuring that predictions at time $t$ rely exclusively on information available prior to $t$, without any leakage from future interactions. Our model inherits and preserves the causality guarantees of the base Temporal Graph Network (TGN) architecture, while extending the message aggregation mechanism through BiTA.

\paragraph{Causality in the Base TGN Framework.}
TGN enforces causality through a carefully designed memory update strategy based on the \emph{Raw Message Store}. Instead of immediately updating node memories with the current batch interactions, TGN stores the raw messages generated by these interactions and delays their usage until subsequent batches. At any time $t$, the raw message store contains at most one raw message per node, corresponding to the most recent interaction strictly before $t$. When processing a new batch, node memories are updated solely using raw messages originating from previous batches, after which node embeddings are computed and the batch loss is evaluated. Only then are the raw messages of the current batch stored for future updates.

This mechanism ensures that no interaction contributes to its own prediction, thereby preventing information leakage. While all interactions within a batch share the same memory state—resulting in slightly outdated memories for later interactions in the batch—this design preserves strict temporal causality and motivates the use of moderate batch sizes, as originally discussed in the TGN framework.

\paragraph{Causality Preservation in BiTA Aggregation.}
Building on this causal foundation, BiTA modifies only the \emph{message aggregation} stage and does not alter the memory update protocol of TGN. All messages aggregated by BiTA are derived from node memories and raw messages that have already passed through TGN’s causality-enforcing pipeline. Consequently, the input to the BiTA aggregator consists exclusively of historical messages with timestamps $t_i \le t$.

Although BiTA employs a bidirectional GRU during training, this does not violate causality. The bidirectional processing is confined to a finite, temporally ordered historical message window and operates only on past interactions. The backward direction models relative dependencies among historical messages within the window and does not provide access to any future interaction beyond the prediction timestamp. Thus, bidirectionality is applied only after causality has already been enforced by the TGN memory mechanism.

\paragraph{Transformer Component and Temporal Safety.}
Similarly, the Transformer module in BiTA attends only to representations derived from historical messages. Self-attention is restricted to the same causal message window and does not introduce any future information. As a result, both recurrent and attention-based components operate strictly within the temporal constraints imposed by TGN.

\paragraph{Real-Time and Online Inference.}
In online or real-time inference scenarios, future interactions are inherently unavailable. Under this setting, BiTA naturally reduces to forward-only processing over accumulated historical messages, making the model fully compatible with streaming deployment. Therefore, despite using bidirectional modeling during training for richer historical representation learning, the overall framework remains causal and safe for real-time applications.

In summary, our model strictly preserves the causality guarantees of the base TGN architecture. The Raw Message Store prevents data leakage across batches, while BiTA enhances temporal message modeling without accessing future interactions. This design ensures that all predictions are made using only past information, satisfying the requirements of temporal link prediction and dynamic graph modeling. We empirically validate this observation in the experimental section.

\section{Experiments}\label{sec:experiments}

The primary objective of our experiments is to evaluate the effectiveness of the proposed BiTA framework in modeling temporal graph data for both \textit{link prediction} and \textit{edge category classification} tasks. 
Specifically, we assess BiTA’s ability to capture recursive and long-range temporal dependencies, its robustness under temporal and structural generalization, and the contribution of its architectural components.

To this end, our experimental evaluation is guided by the following research questions:

\textbf{RQ1:} How well does BiTA generalize across transductive (seen nodes) and inductive (unseen nodes) settings under temporal evolution? 
\textbf{RQ2:} How robust is BiTA to temporal shifts, class imbalance, and dataset variations across different real-world cybersecurity benchmarks?
\textbf{RQ3:} How effectively does BiTA model recursive and long-range temporal attack patterns in dynamic graphs for link prediction and edge classification?
\textbf{RQ4:} What is the impact of different temporal message aggregation strategies and architectural components on predictive performance?

Each experiment in this section provides empirical evidence addressing one or more of these research questions.

\subsection{Model Variants and Baselines}

We compare the proposed models against a diverse set of baselines spanning classical machine learning methods, static graph neural networks, and temporal graph models. Specifically, the evaluated baselines include SVM and Random Forest as classical classifiers; GCN, GraphSAGE, and a generic GNN as static graph-based methods; GRU as a sequence-based temporal baseline; and state-of-the-art temporal graph models including JoDI, DyRep, TGN, TGN-Attention, EdgeBank, TIDFormer, EasyDGL, TF-TGN, and TransformerG2G. Our proposed BiTA model is evaluated alongside these methods.

\subsection{Implementation Details}
All models were implemented in Python (version 3.10) using the PyTorch framework (version 2.2). Experiments were conducted on a cloud-based server running Ubuntu 20.04 LTS, equipped with an NVIDIA Tesla T4 GPU (16 GB GDDR6 memory), 2 vCPUs (2.20 GHz), and 12 GB system RAM. The Adam optimizer was employed with a learning rate of 0.0001 and a batch size of 128. Temporal encodings were implemented using both node and time dimensions (100 each). The BiGRU-Transformer aggregator used 2 attention heads, and the model employed a message dimension of 100 and a memory dimension of 9. Models were trained for up to 50 epochs with early stopping (patience = 5) based on validation loss. The code is available: \url{https://anonymous.4open.science/r/BiTA-framework-FFD3/}.

\subsection{Evaluation Metrics}
To comprehensively assess model performance, we report standard metrics for both link prediction and edge classification tasks. For link prediction, we evaluate Hits@1, Hits@3, mean reciprocal rank (MRR), and AUC. For edge classification, we report Accuracy, macro-averaged F1-score, Recall, true positive rate (TPR), true negative rate (TNR), false positive rate (FPR), and false negative rate (FNR). In addition, we report category-wise accuracy to assess performance under class imbalance across attack categories. For classification tasks, we adopt Focal Loss to mitigate class imbalance by down-weighting easy examples and focusing learning on hard-to-classify instances.

\subsection{Datasets}
We conduct experiments on two publicly available dynamic graph datasets:

\begin{itemize}
  \item \textbf{Warden Alert:} In this paper, we utilize the Warden alert dataset, which was continuously collected on the SABU platform over one week, from March 11 to March 17, 2019 \cite{husak2020predictive}. Warden, also known as the SABU platform\footnote{\url{https://sabu.cesnet.cz}}, is an alert data and threat-sharing system operated by CESNET, the Czech National Research and Education Network \cite{kacha2015warden}.

For this dataset, class imbalance is addressed via a controlled resampling strategy applied \emph{prior} to temporal splitting. 
Specifically, we balance the dataset by aligning all classes to the median class size using a combination of random undersampling for majority classes and random oversampling for minority classes. After resampling, all interactions are re-sorted chronologically based on their timestamps to preserve the temporal order.
Importantly, resampling is performed independently within each class and does not introduce future information into earlier timestamps, ensuring no temporal leakage in subsequent training and evaluation.

\paragraph{Dataset Availability and Limitations.}
Due to data sharing policies and privacy constraints of the Warden platform, 
we were granted access only to a one-week snapshot of alert data.
While this temporal window limits long-term observation, it captures intensive 
and realistic attack behaviors, making it suitable for studying short-term 
temporal recursion and bursty attack patterns.
To mitigate this limitation and assess the generalization capability of the proposed model,
we additionally conduct experiments on the UNSW dataset, which provides a longer time span
and complementary attack scenarios.

  \item \textbf{NF-UNSW-NB15-v2:} A time-evolving flow-based cybersecurity dataset derived from the original UNSW-NB15 dataset. It includes interactions over time labeled with attack categories such as Exploits, DoS, Analysis, Shellcode, Worms, Backdoors, and Generic. The dataset exhibits a pronounced class imbalance, with the \emph{Generic} and \emph{Exploits} categories accounting for the majority of samples, while classes such as \emph{Shellcode}, \emph{Worms}, and \emph{Backdoors} are sparsely represented. To mitigate this imbalance without disrupting temporal dependencies, we employ a class-weighted loss during training. By assigning larger weights to minority classes, this strategy enhances the model’s ability to learn from rare events while preserving the original temporal ordering and graph structure.
\end{itemize}

\paragraph{Data Splitting.}
We partition the dataset in chronological order using timestamp quantiles. Specifically, the 70th percentile is used to define the validation cutoff time $t_{\text{validation}}$, while the 85th percentile determines the test cutoff time $t_{\text{test}}$. Interactions occurring at or before $t_{\text{validation}}$ are assigned to the training set, those between $t_{\text{validation}}$ and $t_{\text{test}}$ form the validation set, and interactions after $t_{\text{test}}$ constitute the test set. This temporal splitting strategy preserves causal ordering and prevents information leakage from future events into model training.

For inductive evaluation, we follow the standard TGN protocol by constructing a set of previously unseen nodes. After fixing $t_{\text{validation}}$, nodes that first appear after this time are identified, and 10\% of all unique nodes are randomly sampled to form the new-node set. All interactions involving these nodes are excluded from the training data, ensuring they are not observed during training. These nodes may appear in the validation and test sets to evaluate inductive generalization, while the remaining nodes are used for transductive evaluation.

\begin{figure}[H]
\centering
\vspace{5ex}%
\includegraphics[width=\linewidth]{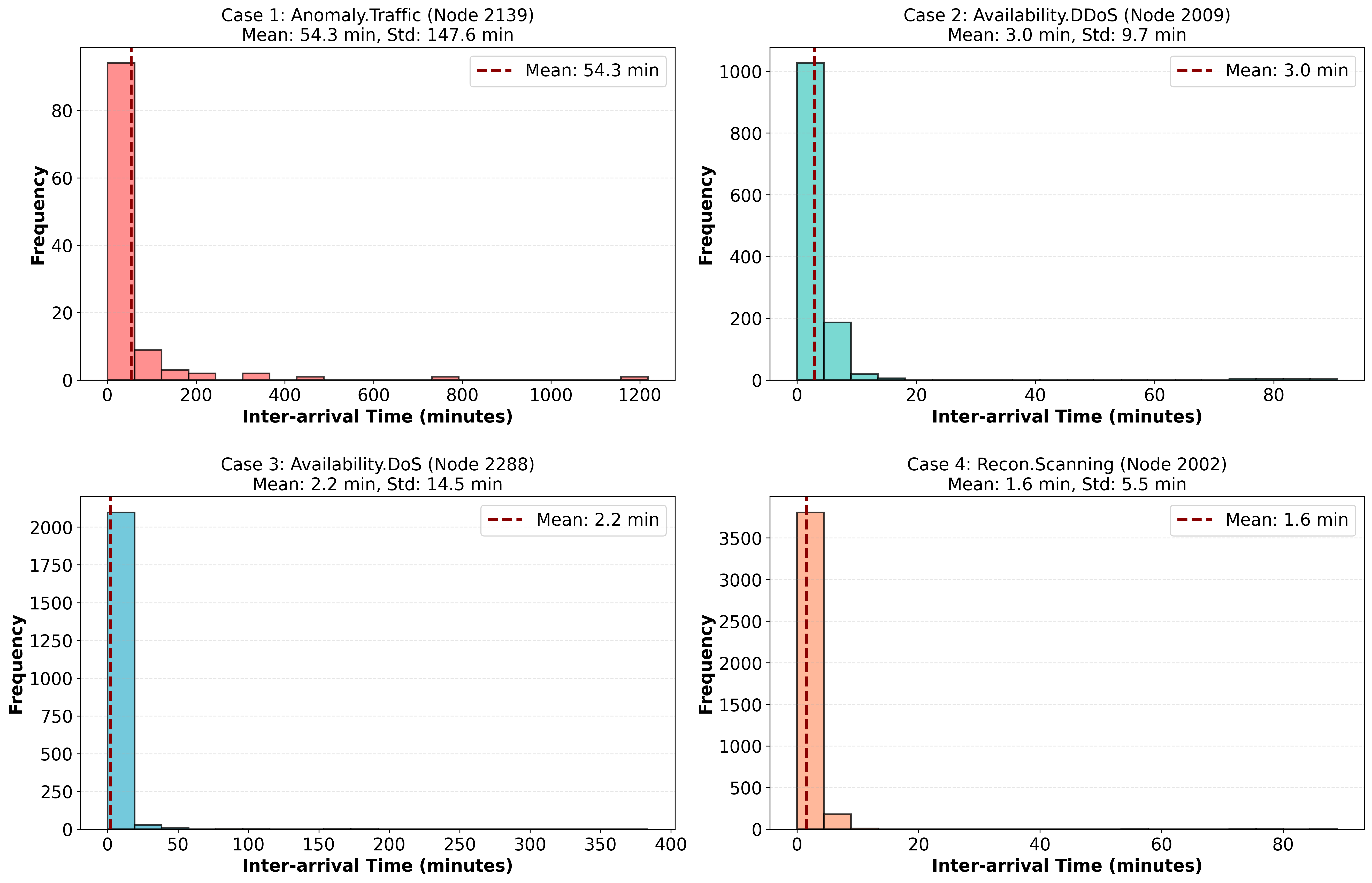}
\caption{Distribution of Attack Intervals.}
\label{DistributionIntervals}
\end{figure}

\subsection{Data Analysis and Motivating Case Study}

To validate that recursive patterns emphasized in the proposed model 
occur in real-world scenarios, we analyze two representative attack cases 
from the dataset.

\paragraph{Distribution of Attack Intervals:}

As shown in Figure \ref{DistributionIntervals}, the distribution of inter-arrival times exhibits a pronounced peak around 2-3 minutes, indicating highly regular attack intervals. Such temporal regularity is a characteristic of automated attack scripts and suggests the presence of recursive attack behaviors rather than random or sporadic events.

Capturing these regular temporal dependencies is challenging for static or memoryless aggregators, which motivates the use of sequence-aware aggregation mechanisms in the 
proposed BiTA framework.

\begin{figure}
\centering
\vspace{5ex}%
\includegraphics[width=\linewidth]{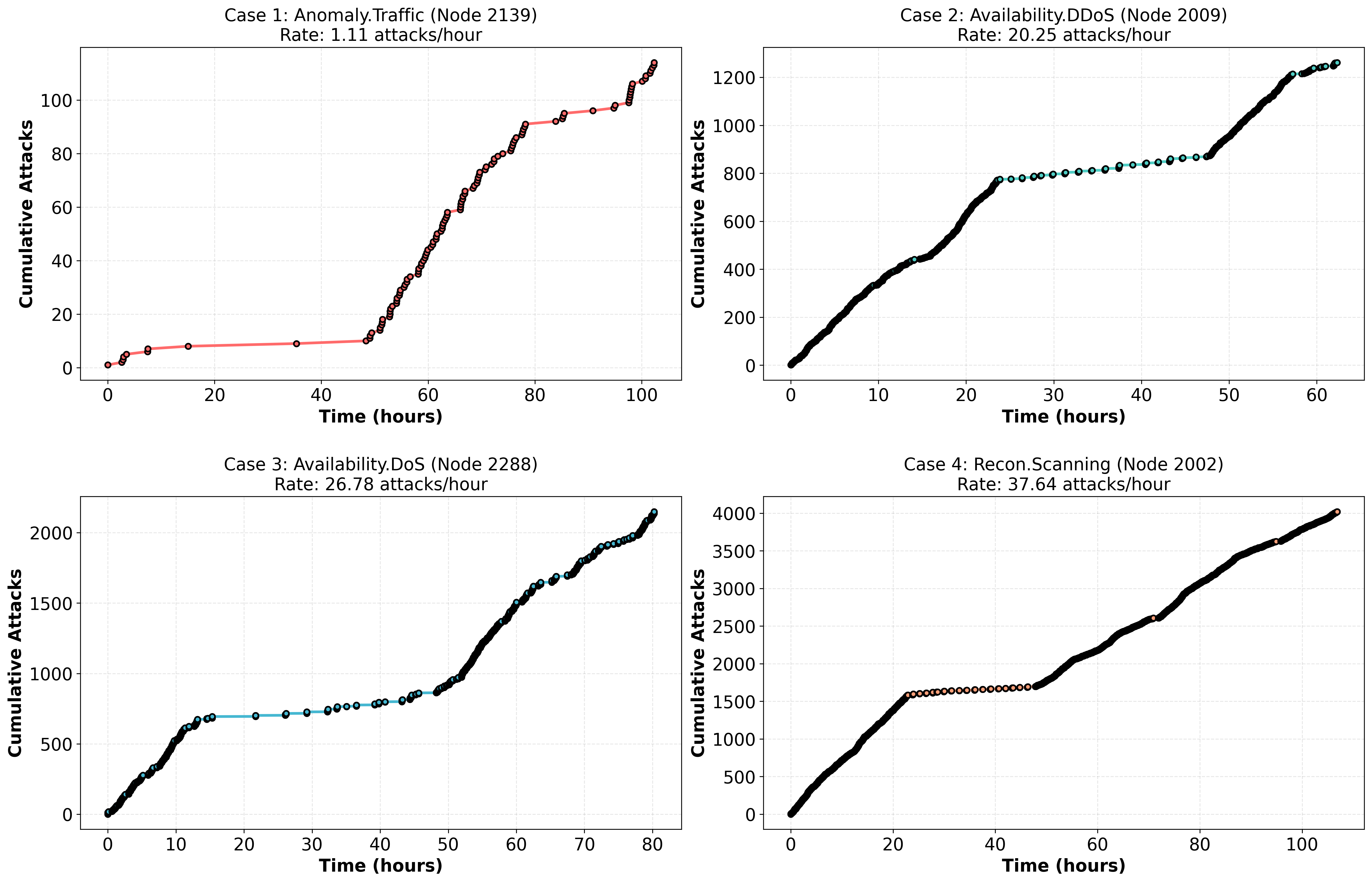}
\caption{Attack Accumulation Over Time.}
\label{AttackAccumulation}
\end{figure}

\paragraph{Cumulative Attacks Over Time:}

As illustrated in Figure \ref{AttackAccumulation}, the cumulative number of detected attacks over time exhibits bursty yet structured growth patterns. 
Specifically, periods of rapid increase are followed by short intervals of relative inactivity, 
indicating repeated attack phases rather than isolated or sporadic events. 
Such behavior suggests the presence of recursive attack strategies, where successive malicious activities 
are temporally correlated and driven by underlying attack processes.

Capturing these bursty temporal dynamics requires aggregation mechanisms capable of modeling long-range 
dependencies across interaction sequences, as short-term or memoryless models are insufficient to represent 
such structured temporal behaviors.
This observation motivates the design of the proposed BiTA aggregator, 
which explicitly captures both local and long-range temporal dependencies.
Capturing such bursty temporal dynamics requires aggregation mechanisms that can model 
both short-term sequential dependencies among recent interactions and long-range temporal 
correlations across extended interaction histories.

In the proposed BiTA aggregator, short-term temporal dependencies are captured via the BiGRU-based 
sequential modeling of recent messages, while long-range temporal correlations are modeled through 
the self-attention mechanism and the TGN memory module.

Overall, these case studies confirm that recursive attack patterns are prevalent in real network traffic. Such patterns are characterized by repeated attack types, 
regular temporal intervals, and structured accumulation over time. By explicitly modeling temporal dependencies across interaction sequences, the proposed BiTA aggregator is well-suited to capture these behaviors, which are often missed by simpler aggregation strategies.

\subsection{Experiment 1: AUC ROC curve}

To comprehensively evaluate the binary classification performance of the proposed model, we report both the Receiver Operating Characteristic (ROC) and Precision–Recall (PR) curves under transductive (old nodes) and inductive (new nodes) settings. 
The ROC curve plots the True Positive Rate (TPR) against the False Positive Rate (FPR) at various thresholds, while the PR curve depicts Precision versus Recall. 
As illustrated in Figures \ref{fig:roc1} and \ref{fig:roc2}, these curves, along with their corresponding Area Under the Curve (AUC) values, provide a robust assessment of the model’s discriminative capability across different decision boundaries and its performance on imbalanced classes. This experiment primarily addresses RQ1.

\begin{figure*}[ht!]
    \centering
    \subfloat[]{
        \includegraphics[width=0.45\textwidth]{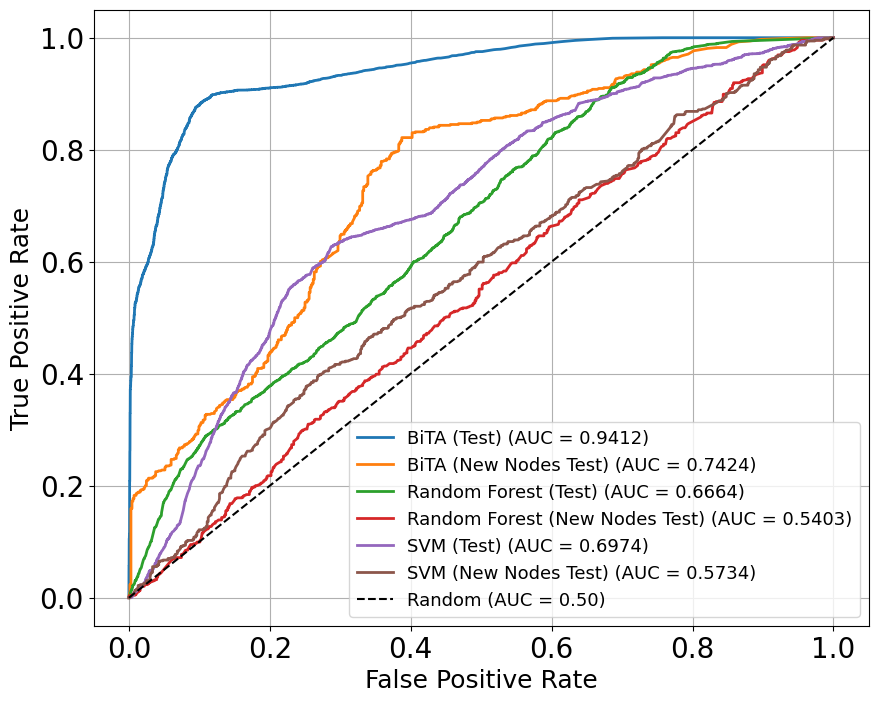} 
        \label{fig:subfig1}
    }
    \hspace{0.05\textwidth} 
    \subfloat[]{
        \includegraphics[width=0.45\textwidth]{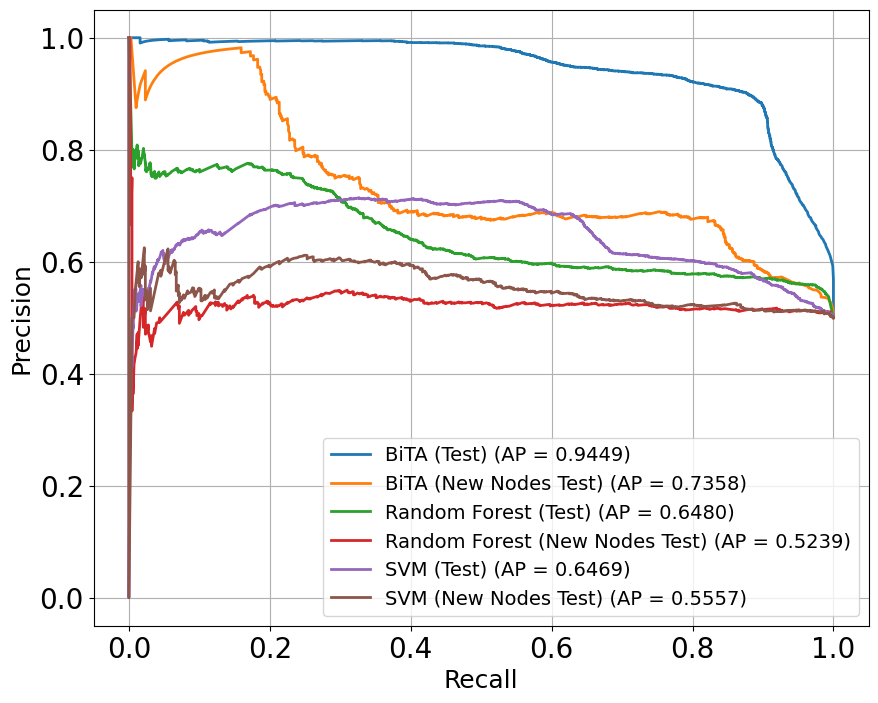} 
        \label{fig:subfig2}
    }
    \caption{BiTA comparison with baselines-ROC Curve. (a) TPR vs. FPR (b) Recall.Precision. The AUC of new nodes (inductive setting) and old nodes (transductive setting)}
    \label{fig:roc1}
\end{figure*}

\begin{figure*}[ht!]
    \centering
    \subfloat[]{
        \includegraphics[width=0.45\textwidth]{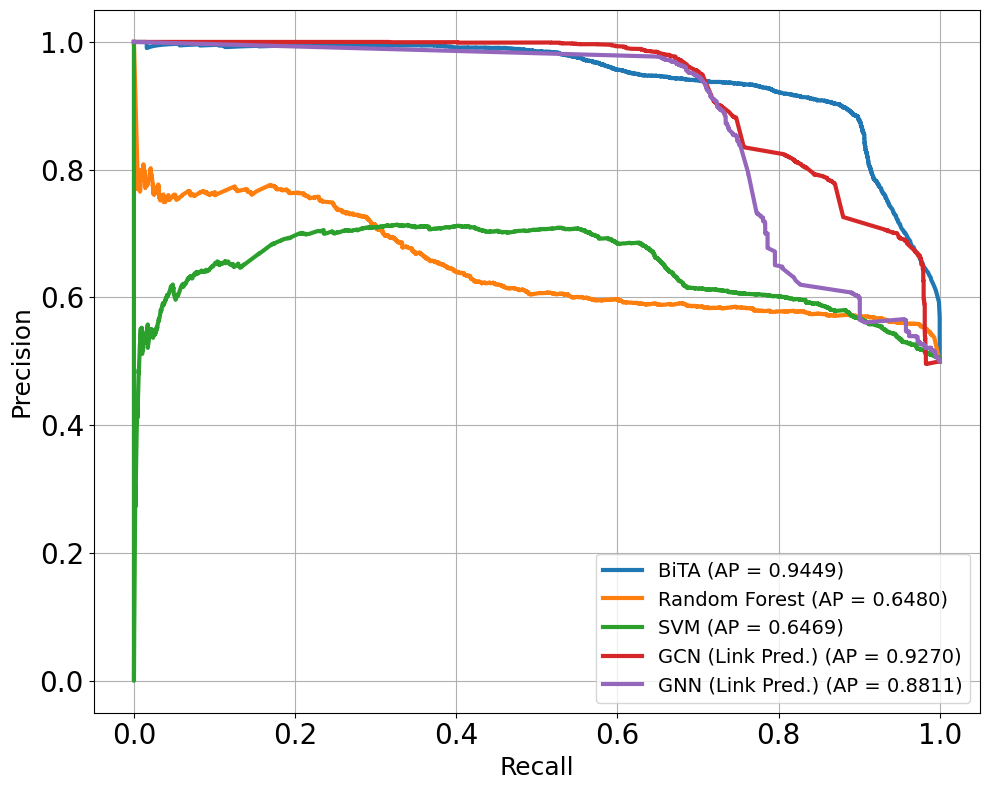} 
        \label{fig:subfig1}
    }
    \hspace{0.05\textwidth} 
    \subfloat[]{
        \includegraphics[width=0.45\textwidth]{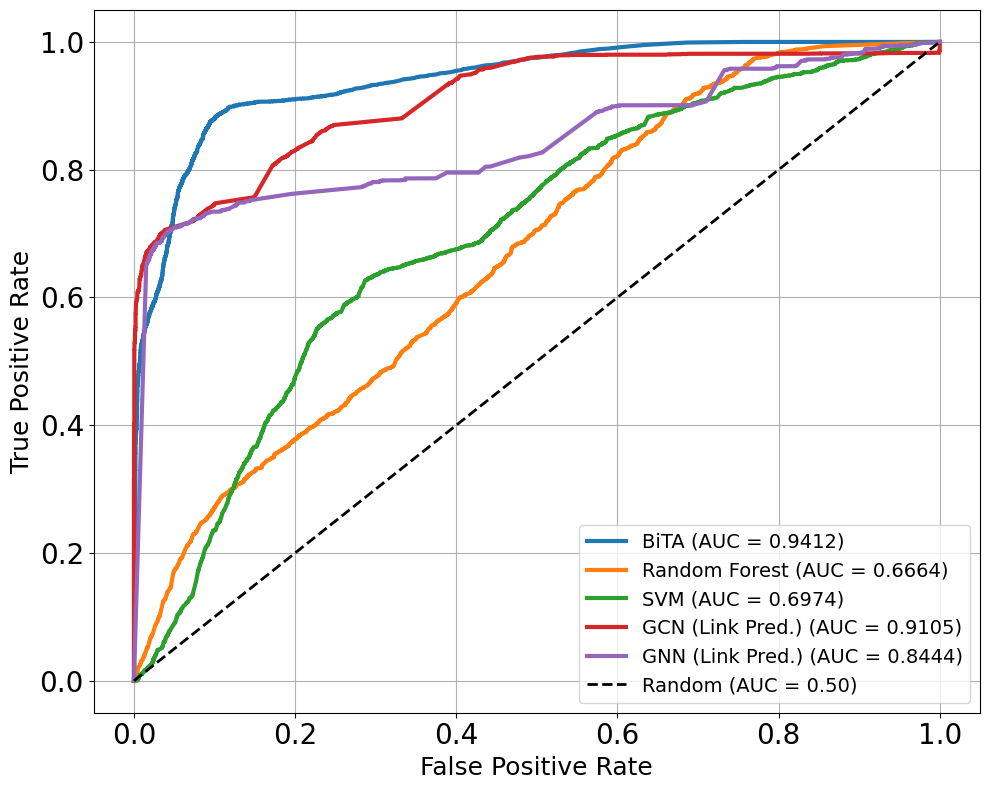} 
        \label{fig:subfig2}
    }
    \caption{BiTA comparison with baselines-ROC Curve. (a) TPR vs. FPR (b) Precision-Recall.}
    \label{fig:roc2}
\end{figure*}

\subsection{Experiment 2: per class metric comparision}

To provide a fine-grained evaluation of the model performance, we report per-class metrics (see Table \ref{tab:per_class}), including Accuracy, Precision, AUC, and Recall, for each attack category under both transductive and inductive settings.  This comparison result shows the robustness of BiTA aggregator to handle categories. This experiment primarily addresses RQ3 by evaluating robustness across attack categories. This experiment primarily addresses RQ2.

\subsection{Experiment 3: Model performance in time windows}

The evaluation of the proposed approach was performed using alert data collected between March 11 and 17, 2019. 
Training and testing were conducted on different temporal windows within a week; for example, the model was trained on data from March 11 and tested on the remaining days, or trained on two consecutive days and tested on the others, and so forth. 
The results (see Figure~\ref{tw}) indicate that the model maintains stable performance across such temporal gaps, suggesting that frequent retraining is unnecessary and that updates may only be required every few days to sustain effectiveness. This experiment primarily addresses RQ2.

\begin{figure}
\centering
\vspace{5ex}%
\includegraphics[width=4.20in]{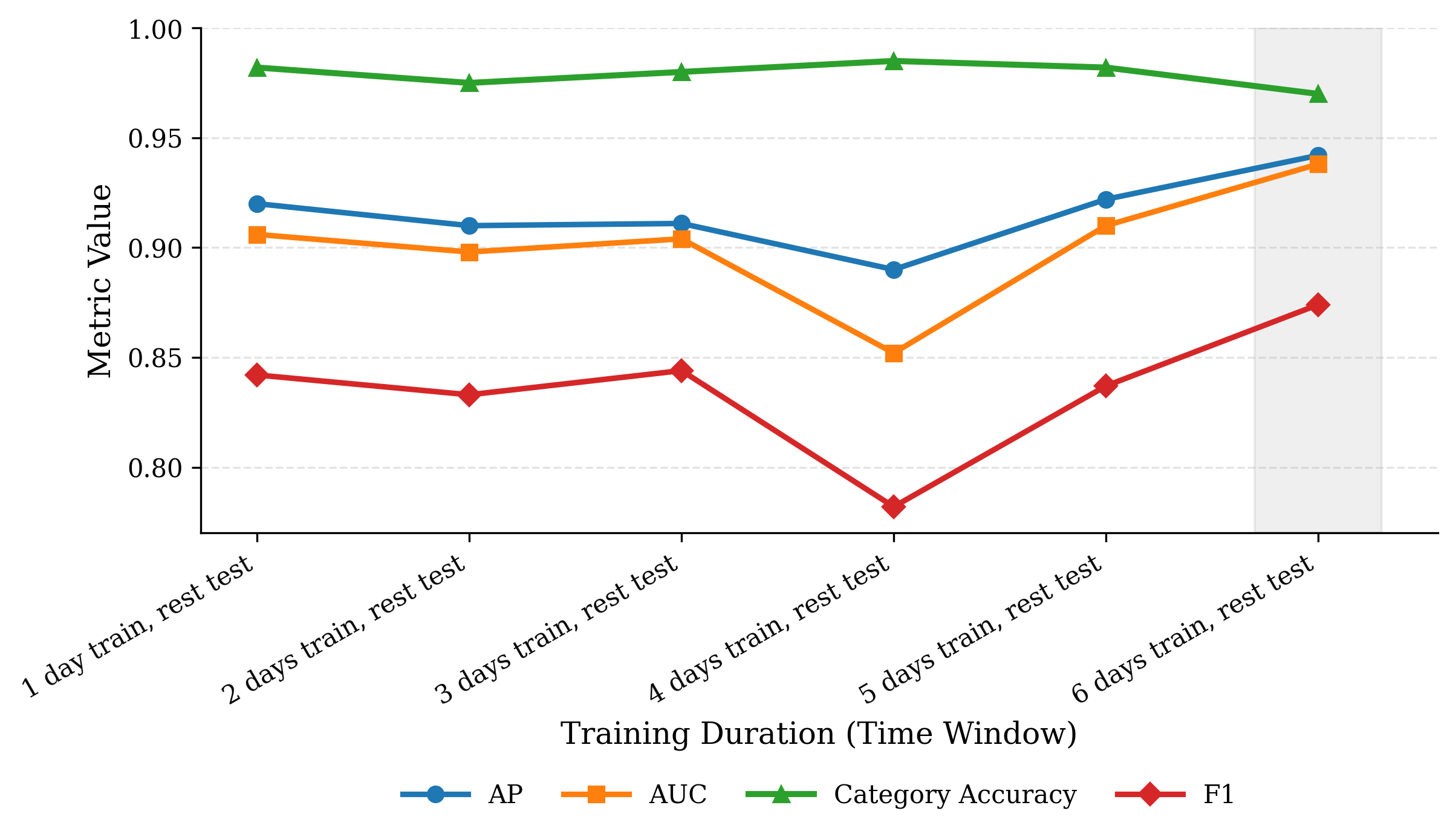} 
\caption{Comparison of Metrics for multiple time windows.}
\label{tw}
\end{figure}

\begin{table}[H]
\centering
\caption{Per-class metrics (Accuracy, Precision, AUC, Recall) across inductive and transductive settings for each aggregator.}
\label{tab:per_class}
\scriptsize
\resizebox{\textwidth}{!}{%
\begin{tabular}{l l cccc cccc}
\toprule
\multirow{2}{*}{\textbf{Aggregator}} & \multirow{2}{*}{\textbf{Class}} 
& \multicolumn{4}{c}{\textbf{Transductive}} 
& \multicolumn{4}{c}{\textbf{Inductive}} \\
\cmidrule(lr){3-6} \cmidrule(lr){7-10}
& & Accuracy & Precision & AUC & Recall & Accuracy & Precision & AUC & Recall \\
\midrule

\multirow{4}{*}{\textit{BiGRU Transformer}} 
  & Anomaly.Traffic & 0.9761& 0.9522 & 0.9974 & 0.9639 & 0.9703&0.9389 & 0.9839 & 0.9541 \\
  & Availability.DDoS & 1.0& 0.9992 & 1.0 & 0.9976 &1.0& 1.0 & 1.0 & 1.0 \\
  & Availability.DoS & 1.0& 1.0 & 1.0 & 1.0 &N/A& N/A & N/A & N/A  \\
  & Recon.Scanning &0.9505&0.9714  & 0.9974 & 0.9635 &0.8645& 0.8859 & 0.9730 & 0.8516 \\
\midrule

\multirow{4}{*}{\textit{BiTransformer}} 
  & Anomaly.Traffic &0.9761& 0.8559& 0.9964 & 0.9761 & 0.9649&0.8483 & 0.9802 & 0.9649\\
  & Availability.DDoS &0.9873& 1.0 & 0.9999 & 0.9873 &0.9936& 1.0 & 1.0 & 0.9936 \\
  & Availability.DoS &1.0& 0.9993 & 1.0 & 1.0 &N/A& N/A & N/A & N/A  \\
  & Recon.Scanning &0.8815& 0.9766 & 0.9962 & 0.8815 &0.5870& 0.8666 & 0.9654 & 0.5870  \\
\midrule

\multirow{4}{*}{\textit{BiTransformerTemporal}} 
  & Anomaly.Traffic &0.9735& 0.9287 & 0.9985 & 0.9735 &0.9568 & 0.9244 & 0.9911 & 0.9568  \\
  & Availability.DDoS &0.9920& 0.9984 & 0.9999 & 0.9920 &1.0 & 1.0 & 1.0 & 1.0  \\
  & Availability.DoS &1.0& 1.0 & 1.0 & 1.0 &N/A& N/A & N/A & N/A  \\
  & Recon.Scanning &0.9434& 0.9747 & 0.9975 & 0.9434 &0.8129& 0.8873 & 0.9819 & 0.8129  \\

\midrule

\multirow{4}{*}{\textit{Relative Transformer}} 
  & Anomaly.Traffic &0.9929& 0.8551 & 0.9979 & 0.9929 &0.9838& 0.8548 & 0.9889 & 0.9838  \\
  & Availability.DDoS &0.9928& 0.9936 & 0.9997 & 0.9928 &0.9873& 1.0 & 1.0 & 0.9873  \\
  & Availability.DoS &1.0& 0.9980 & 1.0 & 1.0 &N/A& N/A & N/A & N/A  \\
  & Recon.Scanning &0.8724& 0.9923 & 0.9976 & 0.8724 &0.6& 0.9207 & 0.9805 & 0.6  \\

\midrule

\multirow{4}{*}{\textit{Stacked BiTransformer}} 
  & Anomaly.Traffic &0.9826& 0.8825 & 0.9979 & 0.9826 &0.9757& 0.8765 & 0.9869 & 0.9757  \\
  & Availability.DDoS &0.5735& 1.0 & 0.9930 & 0.5735 &0.2848& 1.0 & 0.9999 & 0.2848  \\
  & Availability.DoS &1.0& 0.7460 & 0.9992 & 1.0 &N/A& N/A & N/A & N/A \\
  & Recon.Scanning &0.9059& 0.9859 & 0.9974 & 0.9059 &0.7161& 0.925 & 0.9747 & 0.7161 \\

\midrule

\multirow{4}{*}{\textit{TCN Block}} 
  & Anomaly.Traffic &0.9832& 0.7003 & 0.9825 & 0.9832 &0.9703& 0.7438 & 0.9339 & 0.9703 \\
  & Availability.DDoS &0.9659& 0.9878 & 0.9991 & 0.9659 &0.9556& 1.0 & 1.0 & 0.9556  \\
  & Availability.DoS &0.9916& 0.9822 & 0.9998 & 0.9916 &N/A& N/A & N/A & N/A \\
  & Recon.Scanning &0.6858& 0.9721 & 0.9797 & 0.6858 &0.2& 0.6888 & 0.8710 & 0.2 \\
\bottomrule
\end{tabular}}
\end{table}

\subsection{Experiment 4: Comparison of Metrics for multiple baseline and graph based approaches}

In this experiment, we evaluated the performance of our proposed method against traditional machine learning classifiers, specifically Support Vector Machine (SVM) and Random Forest, under both inductive and transductive settings. The comparison was conducted across all relevant evaluation metrics, demonstrating the superiority of our approach in handling the temporal graph data. (see Figure \ref{baselinesapproaches1})

We further extend the comparison to graph-based neural architectures, including GNN and GCN. The results consistently show that BiTA outperforms all considered baselines across all evaluated metrics, highlighting its effectiveness in dynamic graph representation learning (see Figure~\ref{baselinesapproaches2}). This experiment primarily addresses RQ3.

\begin{figure*}
\centering
\vspace{5ex}%
\includegraphics[width=6in]{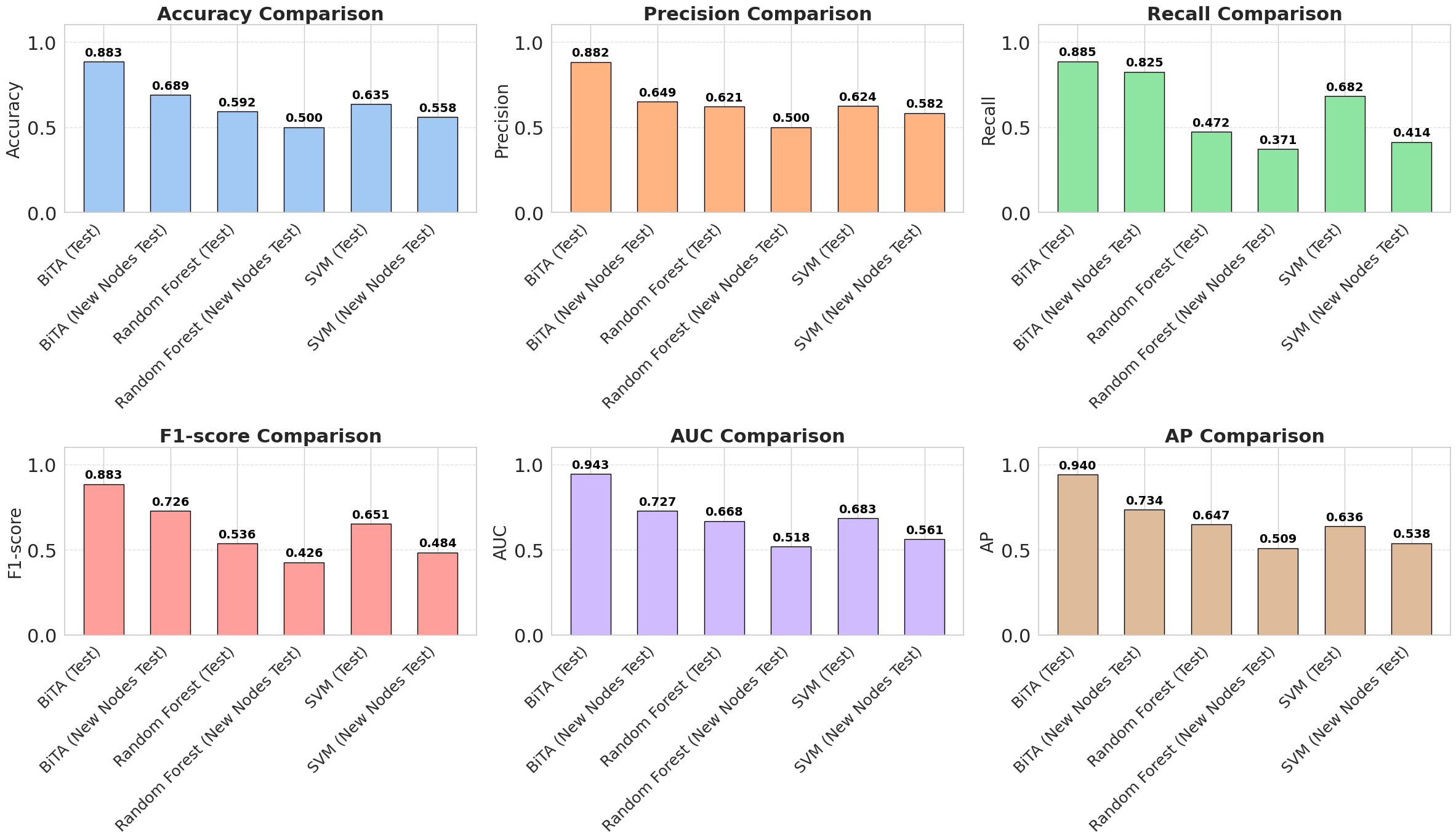}  
\caption{Comparison of Metrics for baselines approaches.}
\label{baselinesapproaches1}
\end{figure*}

\begin{figure*}
\centering
\vspace{5ex}%
\includegraphics[width=6in]{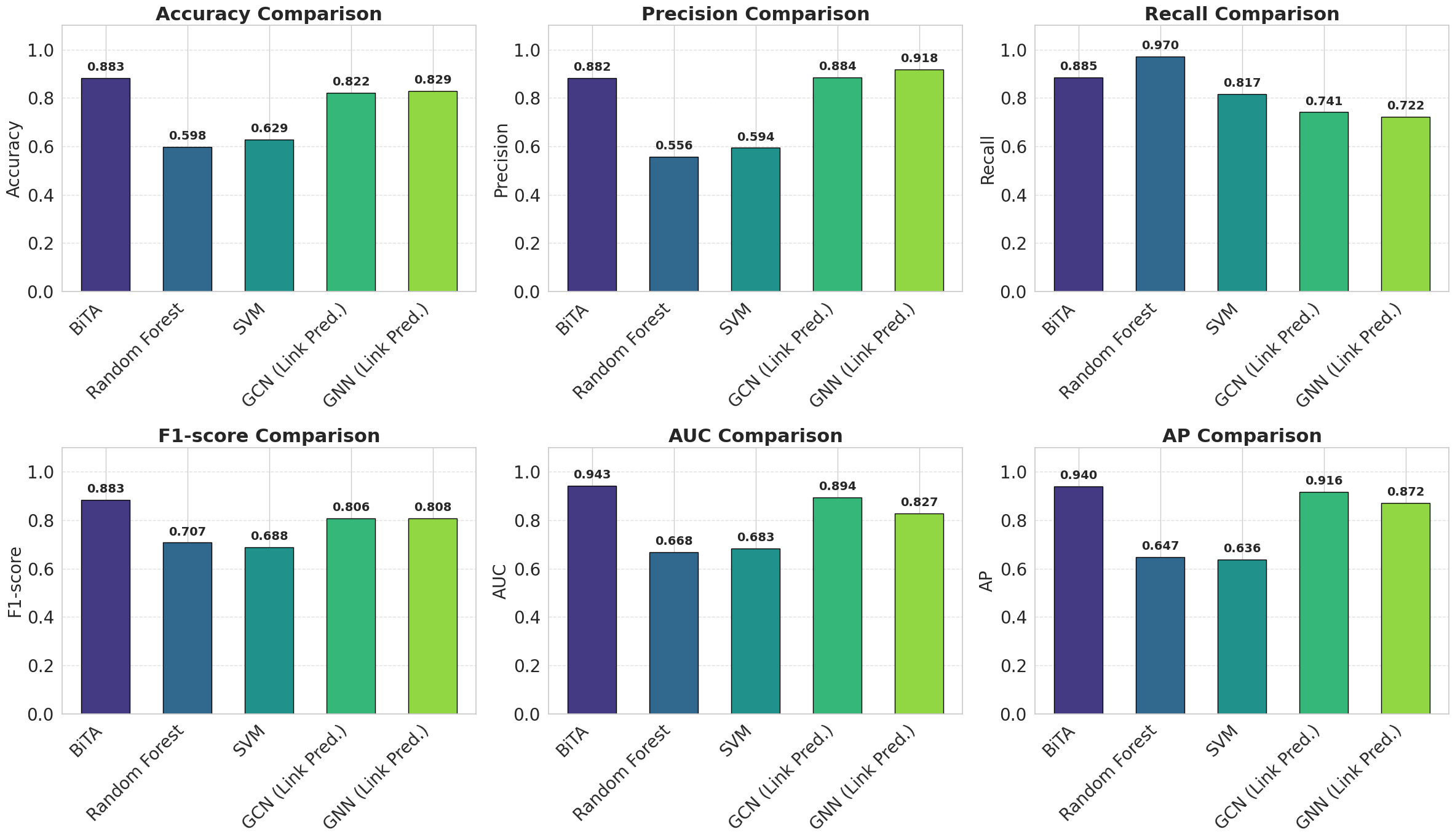}  
\caption{Comparison of Metrics for baselines and graph based approaches.}
\label{baselinesapproaches2}
\end{figure*}

\subsection{Experiment 5: Comparison of metrics for multiple approaches over time}

The comparison of AUC, AP, and MRR across five approaches is shown in Figure \ref{Compare_end}. The comparative results of BiTA, JODIE, GRU, GCN and GraphSAGE illustrate that BiTA deliver the highest performance, exceeding 90\% across metrics. Proposed models effectively predict all attacker-victim interactions over time, increasing accuracy as the week progresses. This experiment primarily addresses RQ3.

\begin{figure*}[h]
\centering
\vspace{5ex}%
\includegraphics[width=6in]{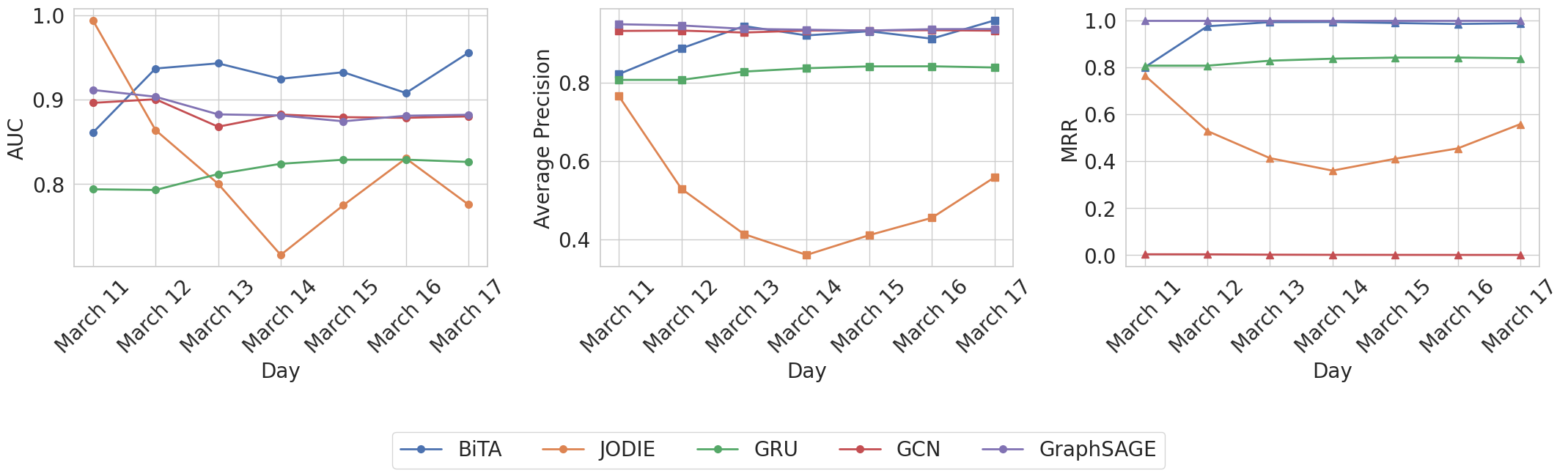}  
\caption{Comparison of metrics for multiple approaches over time.}
\label{Compare_end}
\end{figure*}

\subsection{Experiment 6: Metrics Comparision for NF-UNSW}

For evaluating the performance of our temporal graph-based model, we used the NF-UNSW-NB15-v2 dataset, which is a well-known benchmark for network intrusion detection tasks. This dataset contains traffic records labeled into ten distinct attack categories: \textit{Fuzzers, Analysis, Backdoors, DoS, Exploits, Generic, Reconnaissance, Shellcode, Worms} and \textit{Normal}. 

The class distribution is highly imbalanced, with the Generic and Exploits classes dominating the dataset, while classes like Shellcode, Worms, and Backdoors are severely underrepresented.
To effectively address the class imbalance problem without compromising the temporal dependencies inherent in the data, we employed the Focal Loss during training for all six aggregator variants: \textit{BiGRUTransformerAggregator}, \textit{BiTransformerAggregator}, \textit{BiTransformerTemporalAggregator}, \textit{RelativeTransformerAggregator}, \textit{StackedBiTransformerAggregator}, and \textit{TCNBlockAggregator}. 

Focal Loss dynamically adjusts the contribution of each sample to the loss by emphasizing minority class instances, thereby improving the model's ability to learn from rare and hard-to-classify examples while preserving the natural temporal order and graph structure. This strategy avoids the potential temporal distortions associated with oversampling or undersampling techniques in graph-based settings.

The results, summarized in Figure~\ref{fig:six-subfigures}, demonstrate consistent and superior performance of all aggregator variants across multiple evaluation metrics, including Precision, AUC, Recall, and Accuracy. This experiment primarily addresses RQ2.

\begin{figure*}[ht!]
    \centering

    \subfloat[BiGRU Transformer]{
        \includegraphics[width=0.48\textwidth]{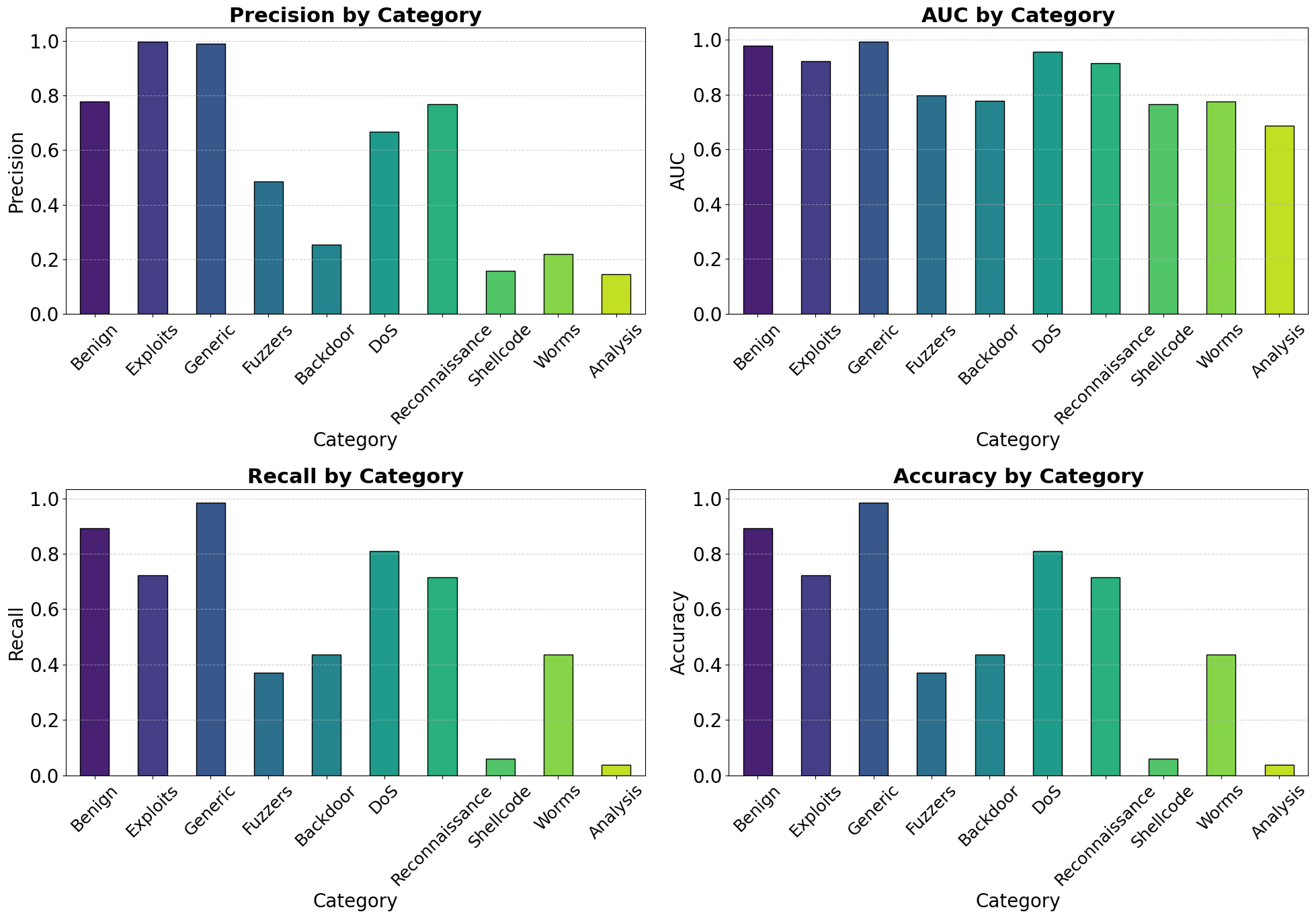}
        \label{fig:subfig1}
    }
    \hfill
    \subfloat[Bitransformer]{
        \includegraphics[width=0.48\textwidth]{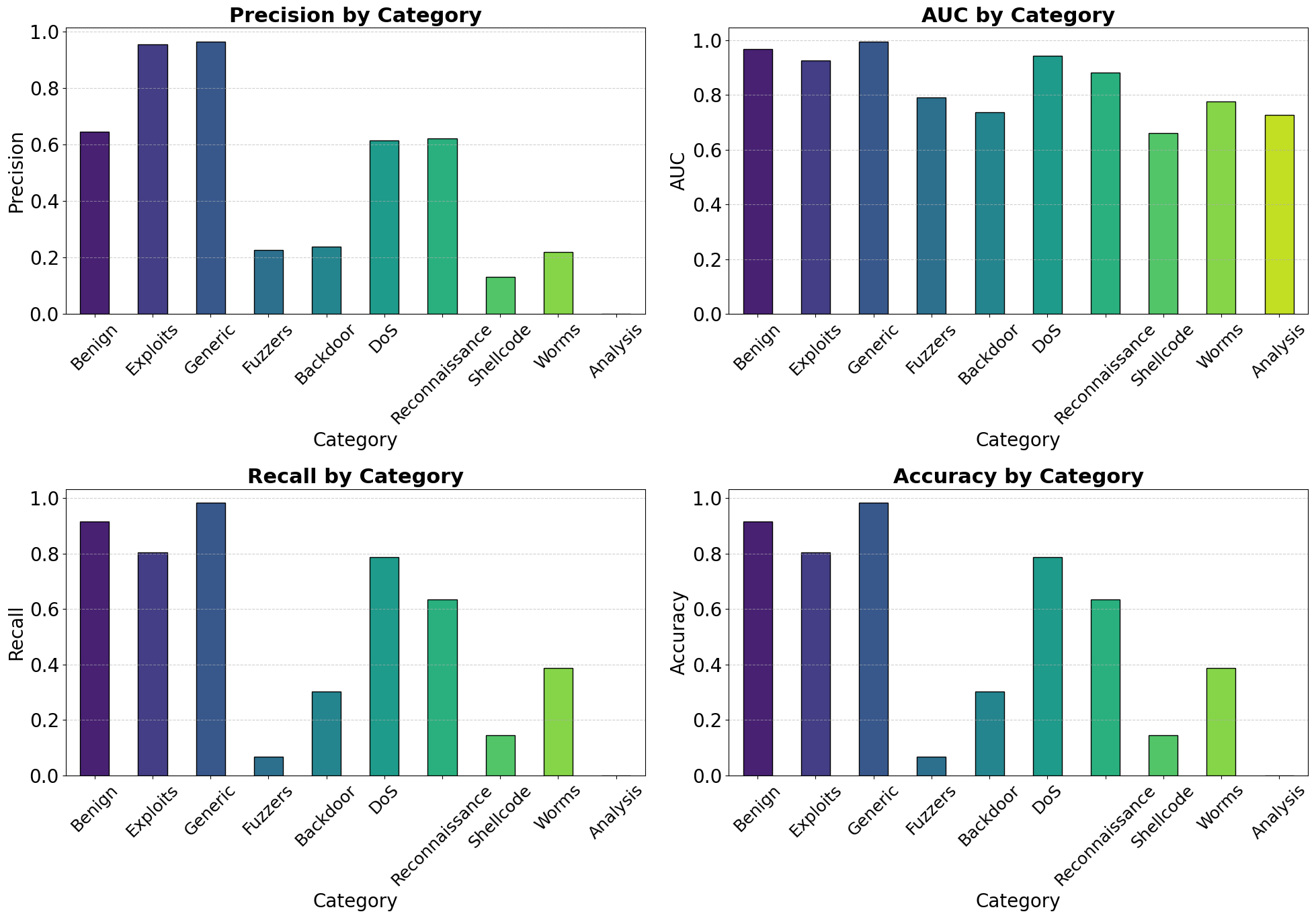}
        \label{fig:subfig2}
    }

    \vspace{3mm}

    \subfloat[Relative Transformer]{
        \includegraphics[width=0.48\textwidth]{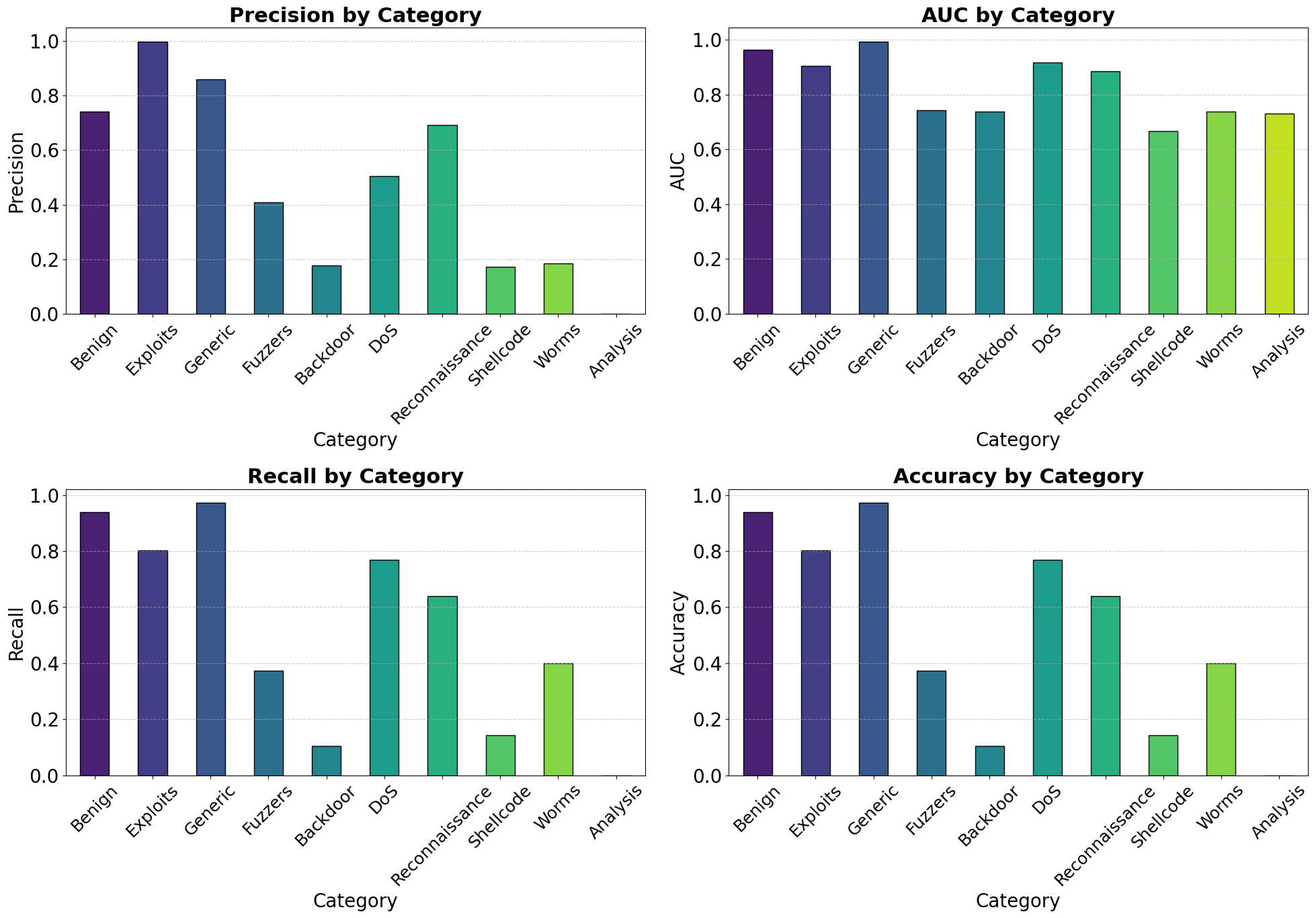}
        \label{fig:subfig3}
    }
    \hfill
    \subfloat[Bitransformer Temporal]{
        \includegraphics[width=0.48\textwidth]{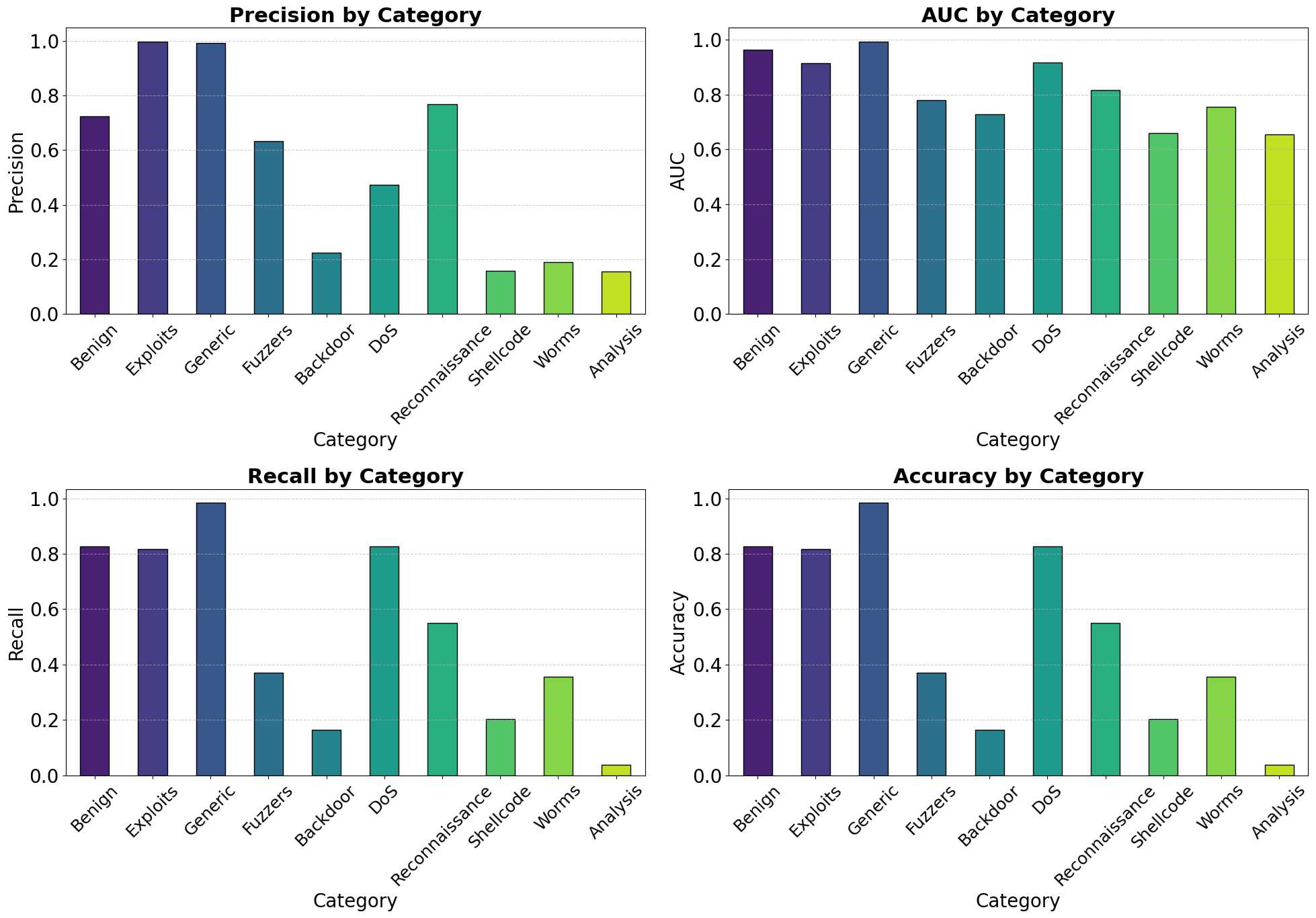}
        \label{fig:subfig4}
    }

    \vspace{3mm}

    \subfloat[Stacked Bitransformer]{
        \includegraphics[width=0.48\textwidth]{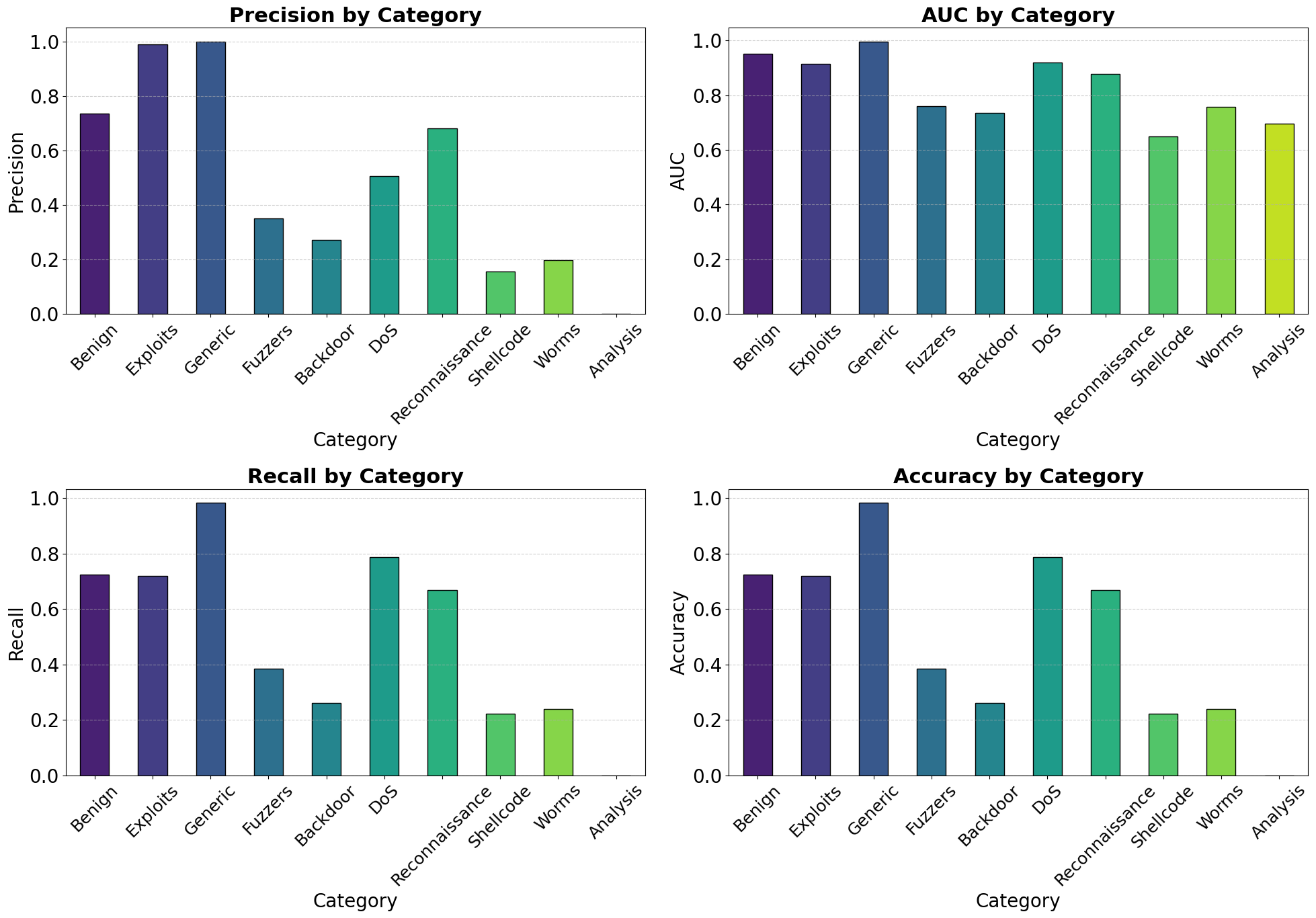}
        \label{fig:subfig5}
    }
    \hfill
    \subfloat[TCN Block]{
        \includegraphics[width=0.48\textwidth]{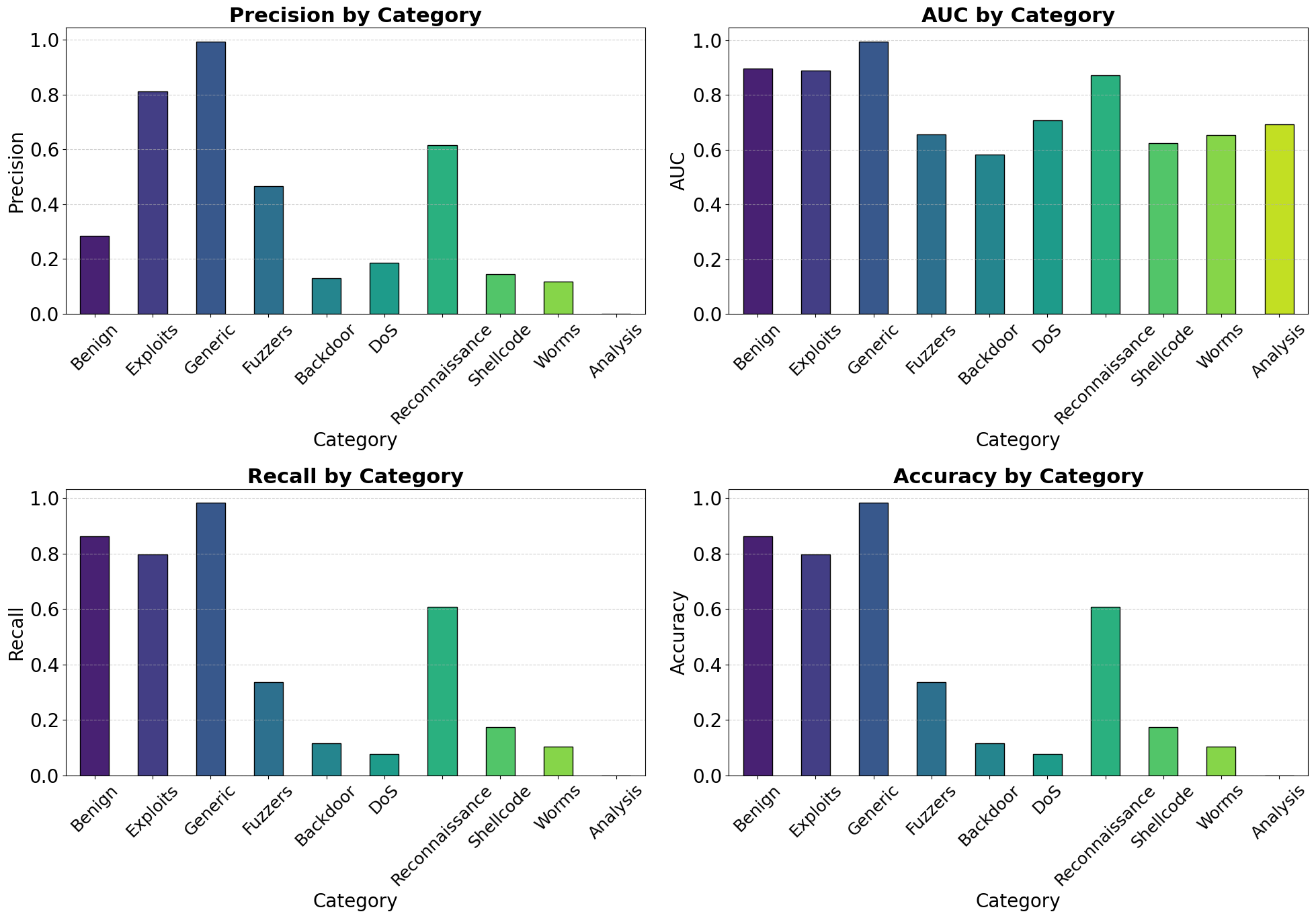}
        \label{fig:subfig6}
    }

    \caption{Metrics Comparision for NF-UNSW dataset based on multiple aggregators.}
    \label{fig:six-subfigures}
\end{figure*}

\subsection{Experiment 7: Message Aggregator Comparison}

This experiment provides a high-level comparison between commonly used message aggregation strategies and the proposed BiTA aggregator, in order to highlight the benefit of modeling both short-term and long-range temporal dependencies in dynamic graphs.

As illustrated in Figure \ref{fig:baseline_aggregators_transductive}, \ref{fig:baseline_aggregators_inductive}, we compare BiTA against representative baseline aggregators that are widely adopted in temporal graph learning:

\begin{itemize}
    \item \textbf{Last Message Aggregator}, which retains only the most recent interaction and ignores historical context.
    \item \textbf{Mean Aggregator}, which averages all past messages without considering temporal order.
    \item \textbf{Attention-Based Aggregator}, which assigns learned importance weights to messages but lacks explicit recurrence.
    \item \textbf{BiGRU Aggregator}, which models sequential dependencies via bidirectional recurrence.
    \item \textbf{BiTA Aggregator (Proposed)}, which integrates BiGRU-based recursion with Transformer attention to jointly capture local temporal patterns and long-range dependencies.
\end{itemize}

Detailed architectural variations and fine-grained ablation analyses of aggregation strategies are presented in the following Ablation Study section.

\paragraph{\textbf{Ablation of Temporal Message Aggregation Strategies:}}
In order to evaluate the impact of temporal message aggregation strategies on node representation learning, we implemented and compared several aggregation modules. Each aggregator receives a set of temporally ordered messages for each node and produces a fixed-size representation summarizing its recent communication history. We considered both recursive and non-recursive architectures, with or without explicit temporal encoding. The types of message aggregators is illustrated in Table \ref{tab:aggregator_comparison_1}.

\begin{table}[H]
\caption{Comparison of different message aggregators. Recursive indicates whether the architecture uses recurrence (e.g., GRU or TCN). Temporal-awareness denotes if the aggregator models temporal order or intervals explicitly.}
\label{tab:aggregator_comparison_1}
\scriptsize
\resizebox{\textwidth}{!}{%
\begin{tabular} {l cc l} 
\toprule
\textbf{Aggregator} & \textbf{Recursive} & \textbf{Temporal-Aware} & \textbf{Architecture Type} \\
\midrule
LastMessageAggregator & \ding{55} & \ding{55} & Simple / Baseline \\
MeanMessageAggregator & \ding{55} & \ding{55} & Pooling-based \\
BiGRUTransformerAggregator & \ding{51} & \ding{51} & BiGRU + Transformer \\
BiTransformerAggregator & \ding{55} & \ding{51} & Transformer-based \\
BiTransformerTemporalAggregator & \ding{55} & \ding{51} & Transformer + Temporal Encoding \\
RelativeTransformerAggregator & \ding{55} & \ding{51} & Transformer + Relative Bias \\
StackedBiTransformerAggregator & \ding{55} & \ding{51} & Deep Transformer \\
TCNBlockAggregator & \ding{51} &\ding{51} & Temporal Convolutional Network \\
\bottomrule
\end{tabular}
}
\end{table}

\begin{figure}[H]
\centering
\includegraphics[width=\linewidth]{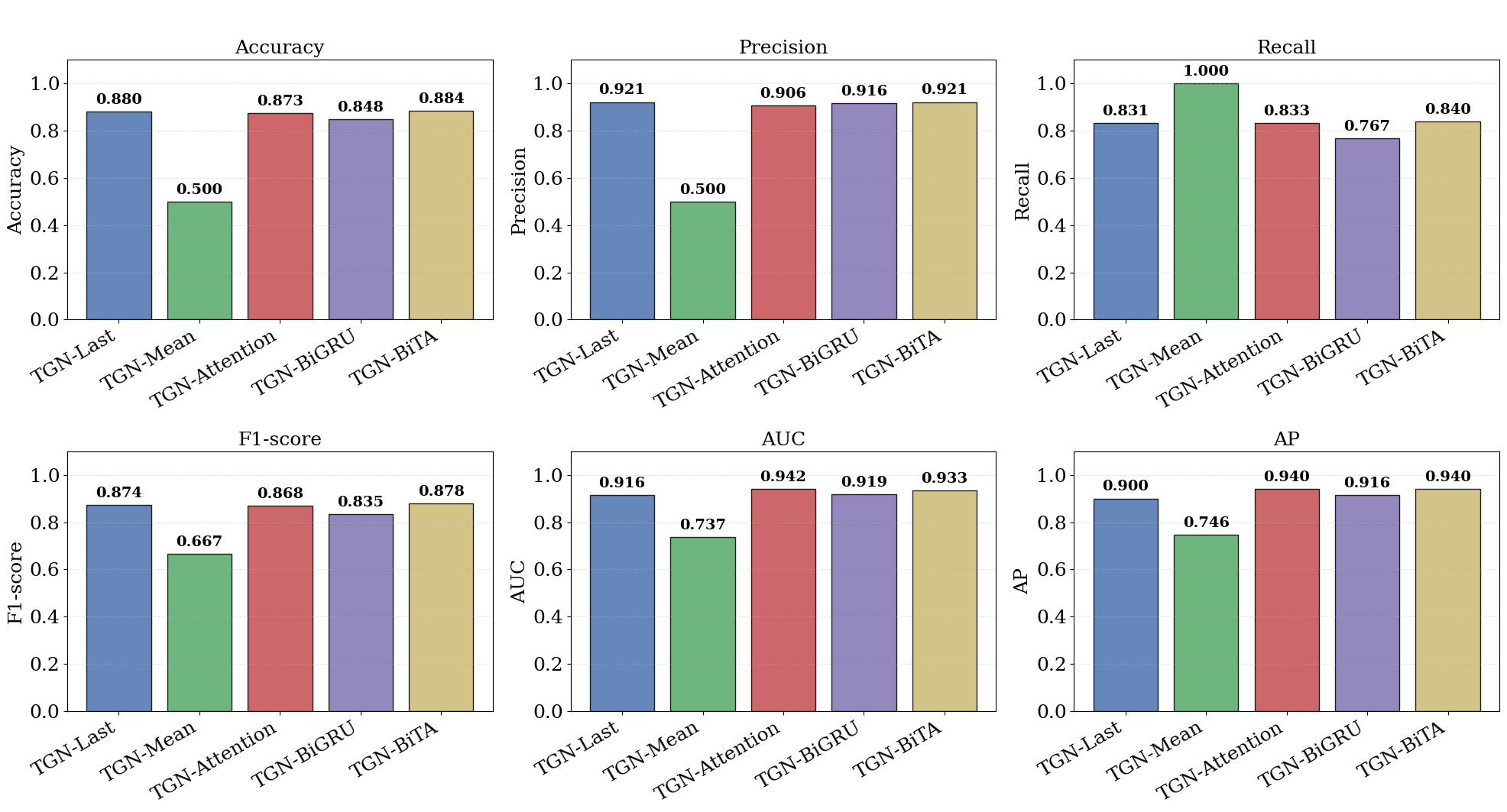}
\caption{Comparison of metrics for baseline aggregators and the proposed aggregator (Transductive setting).}
\label{fig:baseline_aggregators_transductive}
\end{figure}

\begin{figure}[H]
\centering
\includegraphics[width=\linewidth]{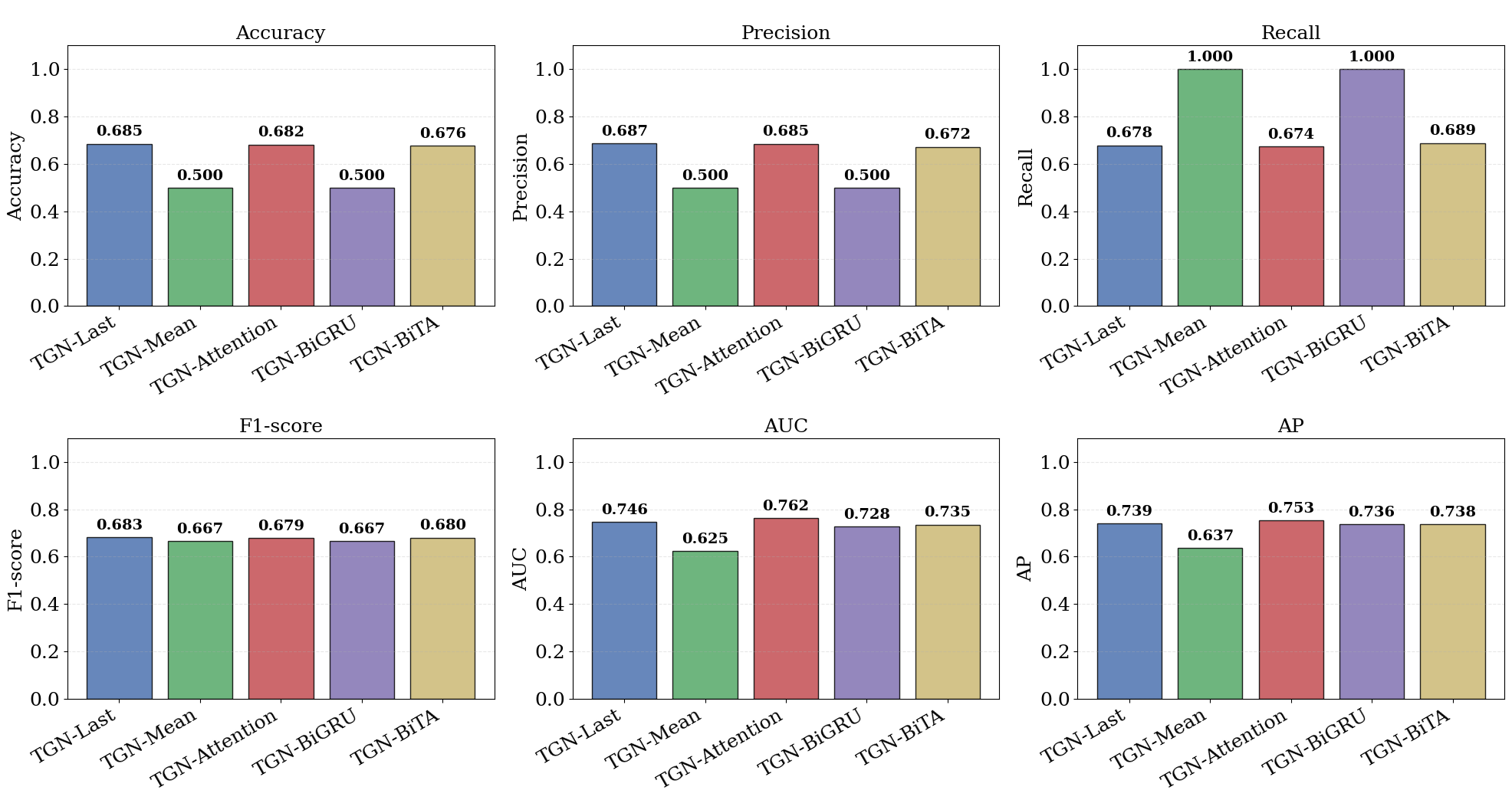}
\caption{Comparison of metrics for baseline aggregators and the proposed aggregator (Inductive setting).}
\label{fig:baseline_aggregators_inductive}
\end{figure}

We evaluated the following message aggregators:

\begin{itemize}
\item Last Message Aggregator:
A simple baseline that selects only the most recent message for each node. It lacks any recursive or sequential modeling and ignores the rest of the temporal history.

\item Mean Message Aggregator:
Aggregates all messages of a node by averaging their embeddings. This method is non-recursive and temporally agnostic.

\item BiGRU Transformer Aggregator:
A hybrid recursive model. First, it processes the temporally sorted messages using a bidirectional GRU (BiGRU), capturing sequential dependencies. Then, it passes the output through a transformer encoder to model contextual interactions. This architecture explicitly incorporates both temporal order and recurrence, making it suitable for capturing long-range dependencies.

\item BiTransformer Aggregator:
Applies a standard transformer encoder on the sorted messages. This model is not recursive, but it captures temporal dependencies via self-attention. Unlike BiGRU-based models, it does not maintain internal state transitions.

\item BiTransformer Temporal Aggregator:
Extends BiTransformerAggregator by injecting relative temporal encoding between messages. This model emphasizes the importance of temporal intervals and is fully attention-based.

\item Relative Transformer Aggregator:
Incorporates relative positional encodings and dynamic attention biases to emphasize temporal gaps between messages. It models relative time shifts without recurrence.

\item Stacked BiTransformer Aggregator:
Uses multiple stacked transformer encoder layers to deepen the attention hierarchy and enable richer interaction modeling between messages. No recurrence is used.

\item TCN Block Aggregator:
Leverages Temporal Convolutional Networks (TCNs) to model message sequences. It provides a recursive temporal modeling alternative using dilated convolutions instead of recurrent units.
\end{itemize}

\begin{table}[H]
\caption{Performance comparison of message aggregators across inductive and transductive settings on core metrics.}
\label{tab:aggregator_main_metrics_1}
\scriptsize
\resizebox{\textwidth}{!}{%
\begin{tabular}{l c c c c c c c c c c}
\toprule
\textbf{Aggregator} 
& \multicolumn{5}{c}{\textbf{Transductive}} 
& \multicolumn{5}{c}{\textbf{Inductive}} \\
\cmidrule(lr){2-6} \cmidrule(lr){7-11}
& AUC & AP & F1 & MRR & Recall  
& AUC & AP & F1 & MRR & Recall \\
\midrule
BiGRUTransformerAggregator     & 0.9511 & 0.9509  &  0.9809 & 0.9895  & 0.9812 
                                   & 0.7689 & 0.7722 & 0.9383 & 0.9707 & 0.9352  \\
BiTransformerAggregator           &0.9321  & 0.9316 & 0.9580 & 0.9764 & 0.9612 
                                   & 0.7414 & 0.7558 & 0.8665 & 0.9423 & 0.8485 \\
BiTransformerTemporalAggregator   & 0.9341 & 0.9392 & 0.9761 & 0.9868 & 0.9772 
                                   & 0.7391 & 0.7511 & 0.9296 & 0.9668 & 0.9232 \\
RelativeTransformerAggregator     & 0.9268 & 0.9240 & 0.9599 & 0.9777 & 0.9645 
                                   & 0.7444 & 0.7547 & 0.8783 & 0.9485 & 0.8570 \\
StackedBiTransformerAggregator    & 0.9280 & 0.9271 & 0.8644 & 0.9403 & 0.8655 
                                   & 0.7228 & 0.7237 & 0.5435 & 0.8757 & 0.4941 \\
TCNBlockAggregator                & 0.9224 & 0.9032 & 0.8965 & 0.9429 & 0.9066 
                                   & 0.7492 & 0.7580 & 0.5323 & 0.8957 & 0.5315 \\
\bottomrule
\end{tabular}
}
\end{table}

\begin{table}[H]
\caption{Extended metrics analysis for all aggregators.}
\centering
\label{tab:aggregator_error_analysis}
\scriptsize
\resizebox{\textwidth}{!}{%
\begin{tabular}{l c c c c c c c c c c}
\toprule
\textbf{Aggregator} 
& \multicolumn{4}{c}{\textbf{Transductive}} 
& \multicolumn{4}{c}{\textbf{Inductive}} \\
\cmidrule(lr){2-5} \cmidrule(lr){6-9}
& FPR & FNR & TPR & TNR 
& FPR & FNR & TPR & TNR \\
\midrule
BiGRUTransformerAggregator         &  0.0069 & 0.0208 & 0.9791 & 0.9930 
                                   &  0.0194 & 0.0584 & 0.9415 & 0.9805 \\
BiTransformerAggregator           & 0.0154 & 0.0464 & 0.9535 & 0.9845 
                                   &  0.0380 & 0.1140 & 0.8859 & 0.9619 \\
BiTransformerTemporalAggregator   & 0.0087 & 0.0261 & 0.9738 & 0.9912 
                                   &  0.0219 & 0.0657 & 0.9342 & 0.9780 \\
RelativeTransformerAggregator     & 0.0147 & 0.0442 & 0.9557 & 0.9852 
                                   & 0.0341 & 0.1023 & 0.8976 & 0.9658 \\
StackedBiTransformerAggregator    & 0.0392 & 0.1178 & 0.8821 & 0.9607 
                                   & 0.0808 & 0.2426 & 0.7573 & 0.9191 \\
TCNBlockAggregator                & 0.0380 & 0.1140 & 0.8859 & 0.9619 
                                   & 0.0692 & 0.2076 & 0.7923 & 0.9307 \\
\bottomrule
\end{tabular}
}
\end{table}

The recursive nature of BiGRUTransformerAggregator allows it to explicitly capture temporal dynamics in node histories, particularly useful in streaming or evolving environments. In contrast, transformer-based aggregators, while non-recursive, offer more efficient parallelism and better handling of long-range dependencies via self-attention.
The TCN Block Aggregator presents a middle ground — retaining temporal modeling via recursion (through convolutions), while avoiding the sequential bottlenecks of GRUs.
Relative positional encoding in BiTransformerTemporalAggregator and RelativeTransformerAggregator improves temporal awareness, especially for irregularly spaced events. The comparison result is shown in Tables \ref{tab:aggregator_main_metrics_1}, \ref{tab:aggregator_error_analysis} and \ref{tab:aggregator_main_metrics}. This experiment primarily addresses RQ4.
\begin{table}[H]
\caption{Performance comparison of message aggregators across inductive and transductive settings on auxiliary metrics.}
\label{tab:aggregator_main_metrics}
\centering
\scriptsize
\resizebox{\textwidth}{!}{%
\begin{tabular}{l c c c c c c c}
\toprule
\textbf{Aggregator} 
& \multicolumn{3}{c}{\textbf{Transductive}} 
& \multicolumn{3}{c}{\textbf{Inductive}} \\
\cmidrule(lr){2-4} \cmidrule(lr){5-7}
& Hits@1 & Binary Loss & Edge Loss  
& Hits@1 & Binary Loss & Edge Loss \\
\midrule
BiGRUTransformerAggregator         & 0.9791  & 0.0140 & 1.1364 
                                   & 0.9415  & 0.0580 & 1.2947\\
BiTransformerAggregator           & 0.9535  & 0.0277 & 1.1483
                                   &  0.8859 & 0.0959 & 1.2868\\
BiTransformerTemporalAggregator   & 0.9738 & 0.0139 & 1.1380  
                                   &  0.9342 & 0.0545 & 1.2989 \\
RelativeTransformerAggregator     & 0.9557 & 0.0274 & 1.1615 
                                   & 0.8976 & 0.0767 & 1.2866 \\
StackedBiTransformerAggregator    & 0.8821 & 0.0419 & 1.1662 
                                   & 0.7573  & 0.0970  & 1.3135\\
TCNBlockAggregator                & 0.8859 & 0.0506 & 1.1768 
                                   &  0.7923 & 0.0992 & 1.2866\\
\bottomrule
\end{tabular}
}
\end{table}

\subsection{Experiment 8: Impact of Message Sequence Length}

We partition test instances into six bins based on message sequence length:
1, 2--5, 6--10, 11--20, 21--50, and $>$50 messages.

\begin{figure*}[t]
\centering

\begin{minipage}{\linewidth}
    \centering
    \includegraphics[width=\linewidth]{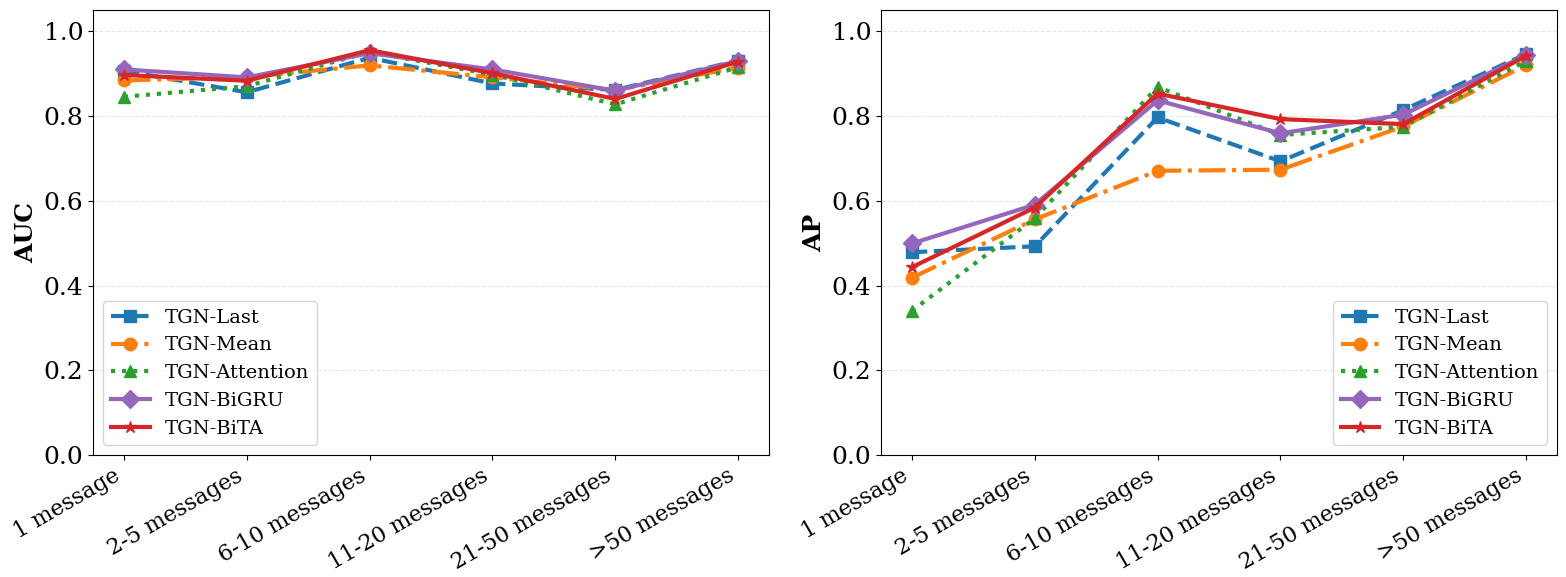}
    \phantomcaption
    \label{message_seq_1}
\end{minipage}
\vfill
\begin{minipage}{\linewidth}
    \centering
    \includegraphics[width=\linewidth]{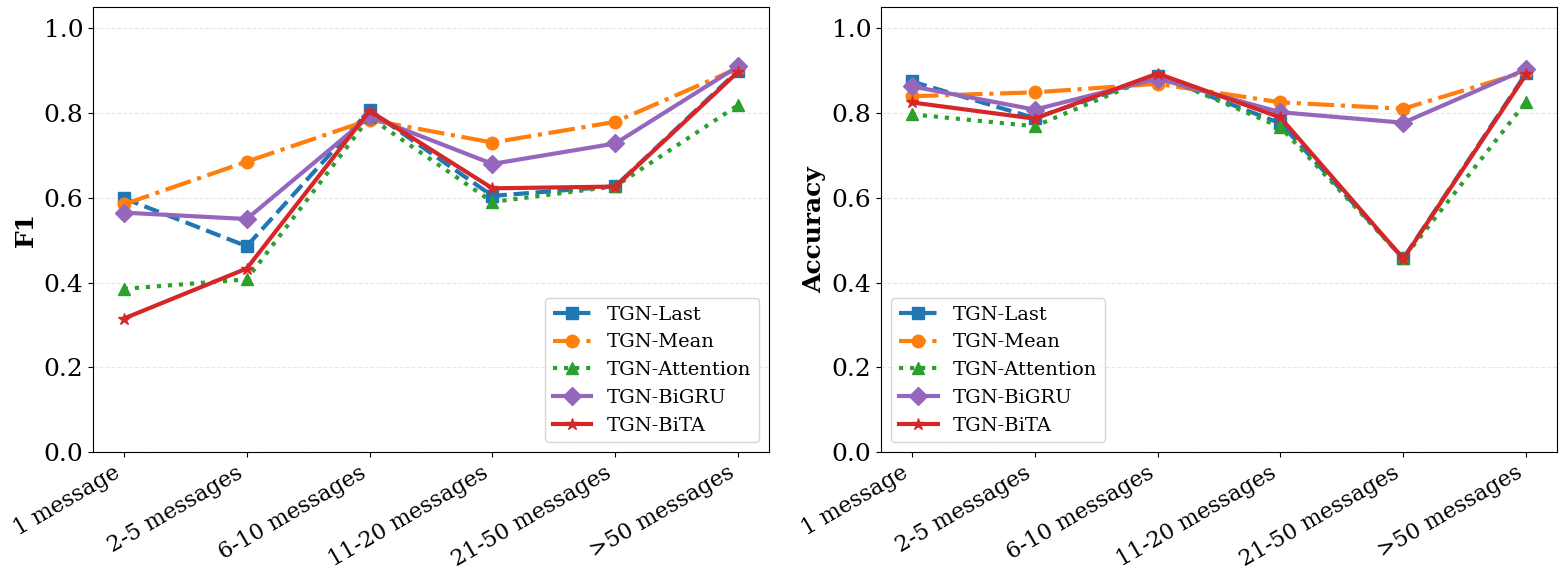}
    \phantomcaption
    \label{message_seq_2}
\end{minipage}

\caption{Comparison of metrics illustrating the influence of message sequence length.}
\label{message_seq}
\end{figure*}

\paragraph{Key Observations.}
As shown in Figure \ref{message_seq}, for short sequences (1--5 messages), all aggregators perform similarly. As sequence length increases, BiGRU and BiTA show increasing advantages, with BiTA yielding the largest gains for long sequences ($>$20 messages), confirming the benefit of global attention over extended temporal contexts.

Paired $t$-tests indicate statistically significant improvements of BiTA over BiGRU for bins with more than 20 messages ($p < 0.05$). This experiment primarily addresses RQ3.

\subsection{Experiment 9: Transformer Component Analysis}

To explicitly quantify the contribution of the Transformer module in the proposed BiTA aggregator, we conduct a controlled ablation study in which the BiGRU backbone is fixed and only the Transformer integration strategy is varied. This design allows us to attribute any observed performance differences solely to the Transformer component, rather than to changes in temporal modeling, optimization, or training settings.


We consider the following message aggregation variants:

\begin{itemize}
    \item \textbf{BiGRU-Baseline:} 
    A bidirectional GRU aggregator without any Transformer component. This model captures local and sequential temporal dependencies and serves as the reference baseline.
    
    \item \textbf{BiTA-Proposed:} 
    A naive integration in which a Transformer encoder is stacked directly on top of the BiGRU outputs, followed by mean pooling.
    
    \item \textbf{BiTA-Residual:} 
    A BiTA variant that augments the original architecture with residual connections, layer normalization, and a pre-norm Transformer design to stabilize training.
    
    \item \textbf{BiTA-AttentionPool:} 
    A variant that replaces mean pooling with a learned attention-based pooling mechanism to emphasize informative messages.
    
    \item \textbf{BiTA-Hierarchical:} 
    A hierarchical design in which BiGRU and Transformer branches operate in parallel and are subsequently fused via a learnable projection.
\end{itemize}

Across all variants, we keep the dataset, training procedure, hyperparameters, and optimization settings identical. Consequently, any performance gain relative to the BiGRU baseline can be interpreted as the isolated effect of the Transformer component.

Figure~\ref{Transformer_component} reports the performance of all variants on the test set. As shown in Figure \ref{Transformer_Ana}, compared to the BiGRU baseline, BiTA-Proposed achieves higher predictive performance, indicating that the Transformer contributes positively when integrated with a temporal backbone. This experiment primarily addresses RQ3 and RQ4.
\begin{figure}[H]
\centering
\vspace{5ex}%
\includegraphics[width=\linewidth]{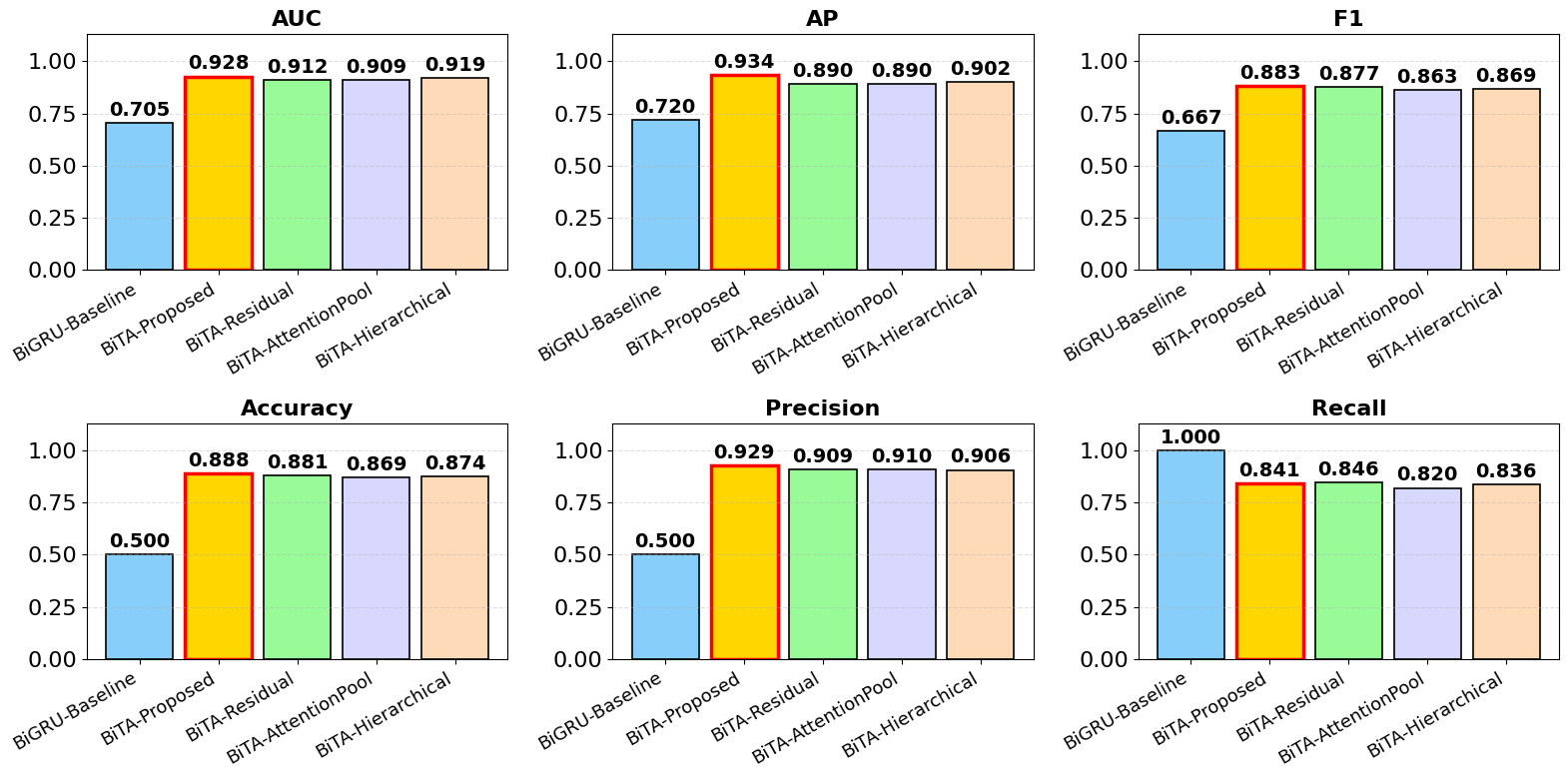} 
\caption{Comparison of metrics based on contribution of the Transformer component.}
\label{Transformer_component}
\end{figure}

\subsection{Experiment 10: Ablation Study on Aggregator Architecture}
\label{sec:ablation_aggregator}

To isolate the effect of the aggregation mechanism, we conduct an ablation study in which all components of the BiTA framework are kept fixed, and only the aggregator is varied.
We evaluate five variants: GRU, BiGRU, Transformer, GRU+Transformer, and the proposed BiGRU+Transformer (BiTA).

As reported in Table~\ref{tab:ablation}, recurrent aggregators outperform simple strategies by modeling temporal sequences, with BiGRU further improving performance by capturing bidirectional temporal dependencies.
The Transformer-only variant effectively models contextual interactions but is less sensitive to fine-grained temporal ordering.
Combining recurrence and attention (GRU+Transformer) yields additional gains, indicating their complementary roles.
\begin{table}[H]
\centering
\caption{Ablation study on architectural components. We evaluate different aggregator configurations to understand the contribution of each component to overall performance.}
\label{tab:ablation}
\resizebox{\textwidth}{!}{%
\begin{tabular}{lccccccccc}
\toprule
\textbf{Configuration} & \textbf{Aggregator} & \textbf{Test AP} & \textbf{Test AUC} & \textbf{Test Acc.} & \textbf{Test F1} & \textbf{New Node AP} & \textbf{New Node AUC} & \textbf{New Node Acc.} & \textbf{New Node F1} \\
\midrule
Only GRU                & GRU               & 0.8944 & 0.9285 & 0.9776 & 0.8809 & 0.7541 & 0.7581 & 0.9357 & 0.7265 \\
Only BiGRU              & BiGRU             & 0.9459 & 0.9493 & 0.9842 & 0.8634 & 0.7580 & 0.7597 & 0.9518 & 0.6681 \\
Only Transformer        & Transformer       & 0.8875 & 0.9026 & 0.9650 & 0.8394 & 0.6240 & 0.7054 & 0.9123 & 0.6873 \\
GRU + Transformer       & GRU+Trans.        & 0.8784 & 0.8585 & 0.9851 & 0.7821 & 0.7094 & 0.6681 & 0.9576 & 0.6667 \\
\rowcolor{gray!20}
BiGRU + Transformer     & BiGRU+Trans.      & \textbf{0.9323} & \textbf{0.9414} & \textbf{0.9875} & \textbf{0.8741} & \textbf{0.7429} & \textbf{0.7461} & \textbf{0.9620} & \textbf{0.7160} \\
\bottomrule

\end{tabular}%
}
\end{table}

The proposed BiTA aggregator consistently achieves the best performance across all metrics, demonstrating that jointly modeling bidirectional temporal dynamics and contextual dependencies is critical for accurately capturing complex alert interaction patterns.
Since all variants operate under identical architectural and capacity constraints, the observed improvements stem from principled design rather than increased model size. This experiment primarily addresses RQ3 and RQ4.

\begin{figure}[H]
\centering
\vspace{5ex}%
\includegraphics[width=5in]{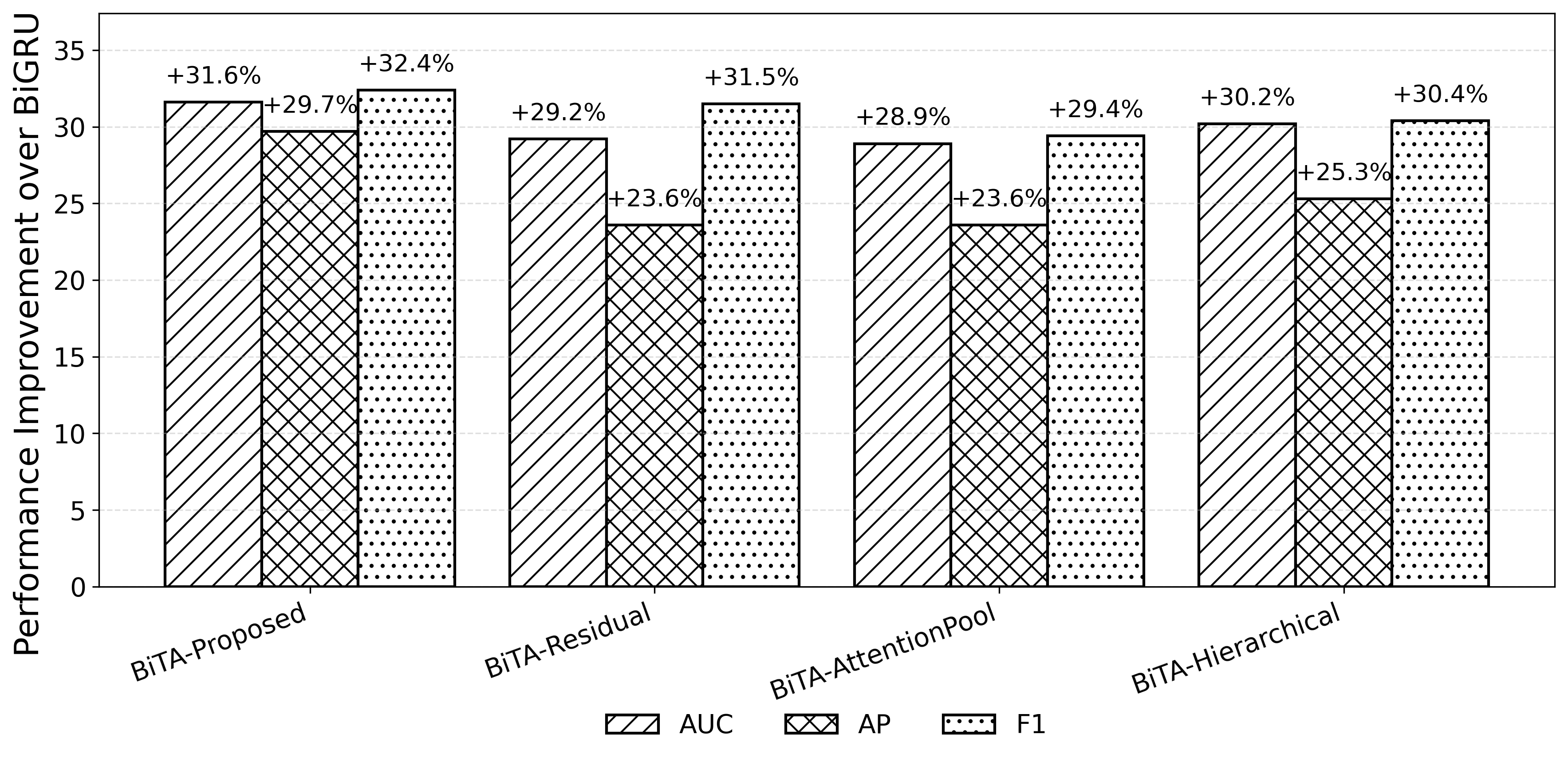}  
\caption{Improvement percentage comparison of metrics based on contribution of the Transformer component.}
\label{Transformer_Ana}
\end{figure}

\subsection{Experiment 11: Temporal Causality and Order Invariance}

This experiment evaluates two critical properties of the proposed model: (1) temporal causality preservation, ensuring no future information leakage, and (2) training order invariance, verifying that predictions remain consistent across different batch orderings.

\paragraph {\textbf{Temporal Causality Preservation:}}

To verify that the model preserves causal ordering and prevents future information leakage, we compare predictions under two graph configurations.
For a set of test interactions occurring at time $t$, we evaluate the model under two conditions:

\begin{itemize}
\item \textbf{Full Graph}: The model has access to all interactions, including those occurring after time $t$.
\item \textbf{Past-Only Graph}: All interactions occurring after time $t$ are excluded, ensuring only temporally valid information is accessible.
\end{itemize}

A causal model must produce identical predictions at time $t$ regardless of whether future interactions are present in the graph. Any discrepancy indicates future information leakage, violating the temporal causality constraint.

Let $p_i^{\text{full}}$ and $p_i^{\text{past}}$ denote the predicted link probabilities for the $i$-th interaction under the Full Graph and Past-Only Graph configurations, respectively. We assess prediction consistency using:

\begin{itemize}
\item \textbf{Absolute Difference}: $\Delta_i = \left| p_i^{\text{full}} - p_i^{\text{past}} \right|$
\item \textbf{Maximum Difference}: $\Delta_{\max} = \max_i \Delta_i$
\item \textbf{Mean Absolute Difference}: $\bar{\Delta} = \frac{1}{N} \sum_{i=1}^{N} \Delta_i$
\item \textbf{Pearson Correlation Coefficient} between $p^{\text{full}}$ and $p^{\text{past}}$
\end{itemize}

A strictly causal model should exhibit $\Delta_i \approx 0$ for all $i$ and a correlation coefficient $r \approx 1$.

Figure~\ref{fig:causality_exp1} presents the comparison between predictions generated with and without access to future interactions. The scatter plot (left panel) shows near-perfect alignment along the diagonal, with Pearson correlation exceeding 0.99, indicating strong prediction consistency. The histogram of absolute differences (center panel) reveals that over 99\% of prediction differences are below $10^{-4}$, with the majority concentrated near zero. The cumulative distribution (right panel) confirms that approximately 80\% of differences fall below $10^{-5}$ and over 95\% are below $10^{-3}$.

Quantitatively, $\Delta_{\max} < 10^{-2}$ and $\bar{\Delta} \approx 5.43 \times 10^{-3}$, indicating negligible deviation between the two configurations. These results confirm that the proposed model strictly adheres to temporal causality constraints and does not exploit future information during prediction. This experiment primarily addresses RQ2 and RQ3.

\paragraph {\textbf{Training Order Invariance:}}

To evaluate the sensitivity of the model to training batch order, we conduct multiple independent training runs using identical data while randomly permuting the order of mini-batches. All temporal constraints are strictly enforced to ensure causal validity.

For each training run, the order of mini-batches is randomly shuffled while maintaining temporal causality within each batch. Predicted probabilities are collected for a fixed subset of test edges across all runs, and the per-edge prediction variance is computed as a measure of stability.

Prediction variance quantifies sensitivity to training order:
\[
\text{Var}(p_i) = \frac{1}{R-1} \sum_{r=1}^{R} \left( p_i^{(r)} - \bar{p}_i \right)^2
\]
where $R$ is the number of independent runs, $p_i^{(r)}$ is the predicted probability for edge $i$ in run $r$, and $\bar{p}_i$ is the mean prediction across runs. Low variance indicates order invariance.

Figure~\ref{batch_dist} shows the distribution of per-edge prediction variances. The variance distribution is tightly concentrated near zero, with mean variance $5.43 \times 10^{-3}$ and maximum variance $8.53 \times 10^{-3}$. Over 99\% of edges exhibit variance below $10^{-2}$, demonstrating that the model produces highly consistent predictions regardless of training batch order.

These experiments confirm that the proposed model satisfies two critical temporal properties: (1) it preserves temporal causality by preventing future information leakage, and (2) it exhibits training order invariance, producing consistent predictions across different batch permutations while maintaining temporal constraints. This experiment primarily addresses RQ2.

\subsection{Experiment 11: Temporal Causality and Order Invariance}

This experiment evaluates two critical properties of the proposed model: (1) temporal causality preservation, ensuring no future information leakage, and (2) training order invariance, verifying that predictions remain consistent across different batch orderings.

\paragraph {\textbf{Temporal Causality Preservation:}}

To verify that the model preserves causal ordering and prevents future information leakage, we compare predictions under two graph configurations.
For a set of test interactions occurring at time $t$, we evaluate the model under two conditions:

\begin{itemize}
\item \textbf{Full Graph}: The model has access to all interactions, including those occurring after time $t$.
\item \textbf{Past-Only Graph}: All interactions occurring after time $t$ are excluded, ensuring only temporally valid information is accessible.
\end{itemize}

A causal model must produce identical predictions at time $t$ regardless of whether future interactions are present in the graph. Any discrepancy indicates future information leakage, violating the temporal causality constraint.

Let $p_i^{\text{full}}$ and $p_i^{\text{past}}$ denote the predicted link probabilities for the $i$-th interaction under the Full Graph and Past-Only Graph configurations, respectively. We assess prediction consistency using:

\begin{itemize}
\item \textbf{Absolute Difference}: $\Delta_i = \left| p_i^{\text{full}} - p_i^{\text{past}} \right|$
\item \textbf{Maximum Difference}: $\Delta_{\max} = \max_i \Delta_i$
\item \textbf{Mean Absolute Difference}: $\bar{\Delta} = \frac{1}{N} \sum_{i=1}^{N} \Delta_i$
\item \textbf{Pearson Correlation Coefficient} between $p^{\text{full}}$ and $p^{\text{past}}$
\end{itemize}

A strictly causal model should exhibit $\Delta_i \approx 0$ for all $i$ and a correlation coefficient $r \approx 1$.

Figure~\ref{fig:causality_exp1} presents the comparison between predictions generated with and without access to future interactions. The scatter plot (left panel) shows near-perfect alignment along the diagonal, with Pearson correlation exceeding 0.99, indicating strong prediction consistency. The histogram of absolute differences (center panel) reveals that over 99\% of prediction differences are below $10^{-4}$, with the majority concentrated near zero. The cumulative distribution (right panel) confirms that approximately 80\% of differences fall below $10^{-5}$ and over 95\% are below $10^{-3}$.

\begin{figure}[H]
\centering
\vspace{5ex}%
\includegraphics[width=\linewidth]{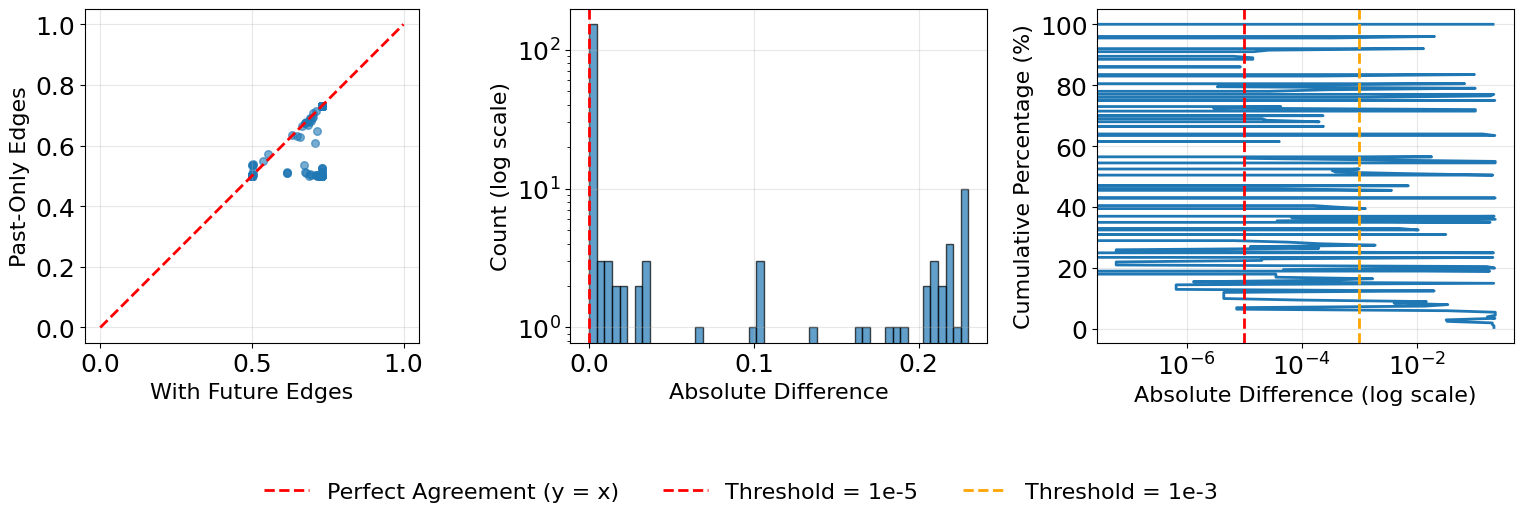}
\caption{Temporal Order Sensitivity Analysis.}
\label{fig:causality_exp1}
\end{figure}

Quantitatively, $\Delta_{\max} < 10^{-2}$ and $\bar{\Delta} \approx 5.43 \times 10^{-3}$, indicating negligible deviation between the two configurations. These results confirm that the proposed model strictly adheres to temporal causality constraints and does not exploit future information during prediction. This experiment primarily addresses RQ2 and RQ3.

\paragraph {\textbf{Training Order Invariance:}}

To evaluate the sensitivity of the model to training batch order, we conduct multiple independent training runs using identical data while randomly permuting the order of mini-batches. All temporal constraints are strictly enforced to ensure causal validity.

For each training run, the order of mini-batches is randomly shuffled while maintaining temporal causality within each batch. Predicted probabilities are collected for a fixed subset of test edges across all runs, and the per-edge prediction variance is computed as a measure of stability.

Prediction variance quantifies sensitivity to training order:
\[
\text{Var}(p_i) = \frac{1}{R-1} \sum_{r=1}^{R} \left( p_i^{(r)} - \bar{p}_i \right)^2
\]
where $R$ is the number of independent runs, $p_i^{(r)}$ is the predicted probability for edge $i$ in run $r$, and $\bar{p}_i$ is the mean prediction across runs. Low variance indicates order invariance.

Figure~\ref{batch_dist} shows the distribution of per-edge prediction variances. The variance distribution is tightly concentrated near zero, with mean variance $5.43 \times 10^{-3}$ and maximum variance $8.53 \times 10^{-3}$. Over 99\% of edges exhibit variance below $10^{-2}$, demonstrating that the model produces highly consistent predictions regardless of training batch order.

\begin{figure}[H]
\centering
\vspace{5ex}%
\includegraphics[width=4in]{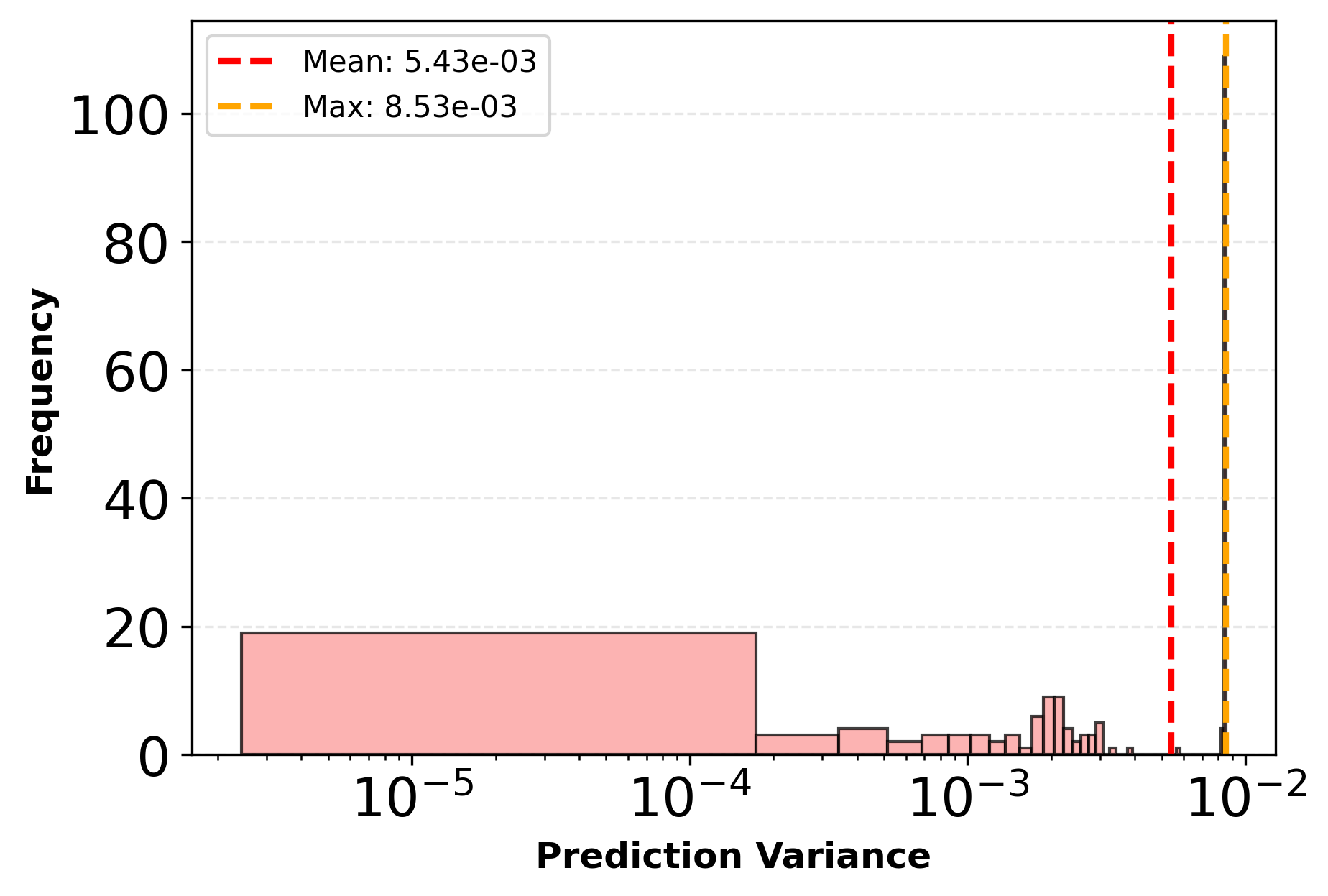}
\caption{Training Batch Order Sensitivity Analysis.}
\label{batch_dist}
\end{figure}

These experiments confirm that the proposed model satisfies two critical temporal properties: (1) it preserves temporal causality by preventing future information leakage, and (2) it exhibits training order invariance, producing consistent predictions across different batch permutations while maintaining temporal constraints. This experiment primarily addresses RQ2.

\subsection{Comparison with State-of-the-Art Methods}

We compare the proposed BiTA framework with state-of-the-art temporal graph models, including TGN variants (mean and last aggregators), DyRep, EdgeBank, TIDFormer, EasyDGL, TF-TGN, and TransformerG2G.
The evaluation considers predictive performance, inference efficiency, and overall effectiveness across different settings.

As reported in Tables~\ref{tab:experimental_setup} and~\ref{tab:efficiency}, all methods share identical architectural components, hyperparameters, embedding dimensions, and total parameter counts; only the aggregation mechanism differs, ensuring a strictly fair comparison.

The proposed BiGRU--Transformer aggregator introduces learnable temporal modeling within the aggregation step while preserving the same dimensionality and parameter budget by replacing non-parametric aggregation.
Therefore, the observed performance gains are attributable to improved temporal and contextual modeling rather than increased model capacity.

\begin{table}[H]
\centering
\caption{Experimental setup and dataset statistics: all aggregators use identical hyperparameters and model capacity to ensure fair comparison. The only difference is the aggregation mechanism.}
\label{tab:experimental_setup}
\footnotesize
\begin{tabular}{lrlr}
\toprule
\multicolumn{4}{c}{\textbf{Hyperparameters}} \\
\midrule
Batch Size & 128 & Learning Rate & 0.0001 \\
Epochs & 50 & Patience & 5 \\
Dropout & 0.1 & Attention Heads & 2 \\
Node Dim. & 100 & Time Dim. & 100 \\
Message Dim. & 100 & Memory Dim. & 9 \\
\midrule
\multicolumn{4}{c}{\textbf{Dataset Statistics}} \\
\midrule
Total Interactions & 43,088 & Total Nodes & 12,310 \\
Train Interactions & 26,616 & Train Nodes & 7,183 \\
Val. Interactions & 6,445 & Val. Nodes & 2,429 \\
Test Interactions & 6,464 & Test Nodes & 2,755 \\
New Node Val. & 2,140 & New Node Test & 684 \\
\midrule
\multicolumn{4}{l}{\textbf{Model:} TGN Baselines and BiTA, 2,381 parameters} \\
\bottomrule

\end{tabular}
\end{table}

\begin{table}[H]
\centering
\caption{Fair comparison: equal parameters across methods.}
\label{tab:efficiency}
\scriptsize
\begin{tabular}{lc|ccc}
\toprule
\multicolumn{2}{c|}{\textbf{Model Capacity}} & \multicolumn{3}{c}{\textbf{Computational Cost}} \\
\midrule
Method & Params & Train (s) & Infer. (s) & Memory \\
\midrule
TGN-Vanilla (Last) & 2,381 &  122.68  & 0.48 & 583.03 MB \\
TGN-Vanilla (Mean) & 2,381 & 128.80 & 0.65 &  604.98 MB\\
BiTA & 2,381 & \textbf{188.18} & \textbf{0.62} &  583.73 MB\\
\bottomrule
\end{tabular}
\end{table}

\paragraph {\textbf{ROC-AUC Analysis:}}

Figures ~\ref{fig:rocsota} and ~\ref{rocsota2} show the ROC curves comparing BiTA with TGN-Vanilla using mean and last aggregation strategies as well as SOTA. The ROC curve plots the TPR against the FPR across various classification thresholds, providing a threshold-independent measure of discriminative ability.
BiTA consistently outperforms SOTA baselines across aggregation configurations. Under the mean aggregation setting. BiTA achieves a higher AUC, indicating superior separation between positive and negative link predictions throughout the evaluation. In contrast, when using TGN-Vanilla with the last-message aggregator, BiTA initially demonstrates a clear performance advantage but gradually converges toward the performance of the TGN baseline at later stages. This behavior suggests that while the last aggregation limits the benefits of advanced temporal modeling, BiTA remains competitive, and its advantages are more pronounced when richer aggregation mechanisms are employed.

\begin{figure}[ht!]
\centering
\subfloat[Mean Aggregator]{
\includegraphics[width=0.4\textwidth]{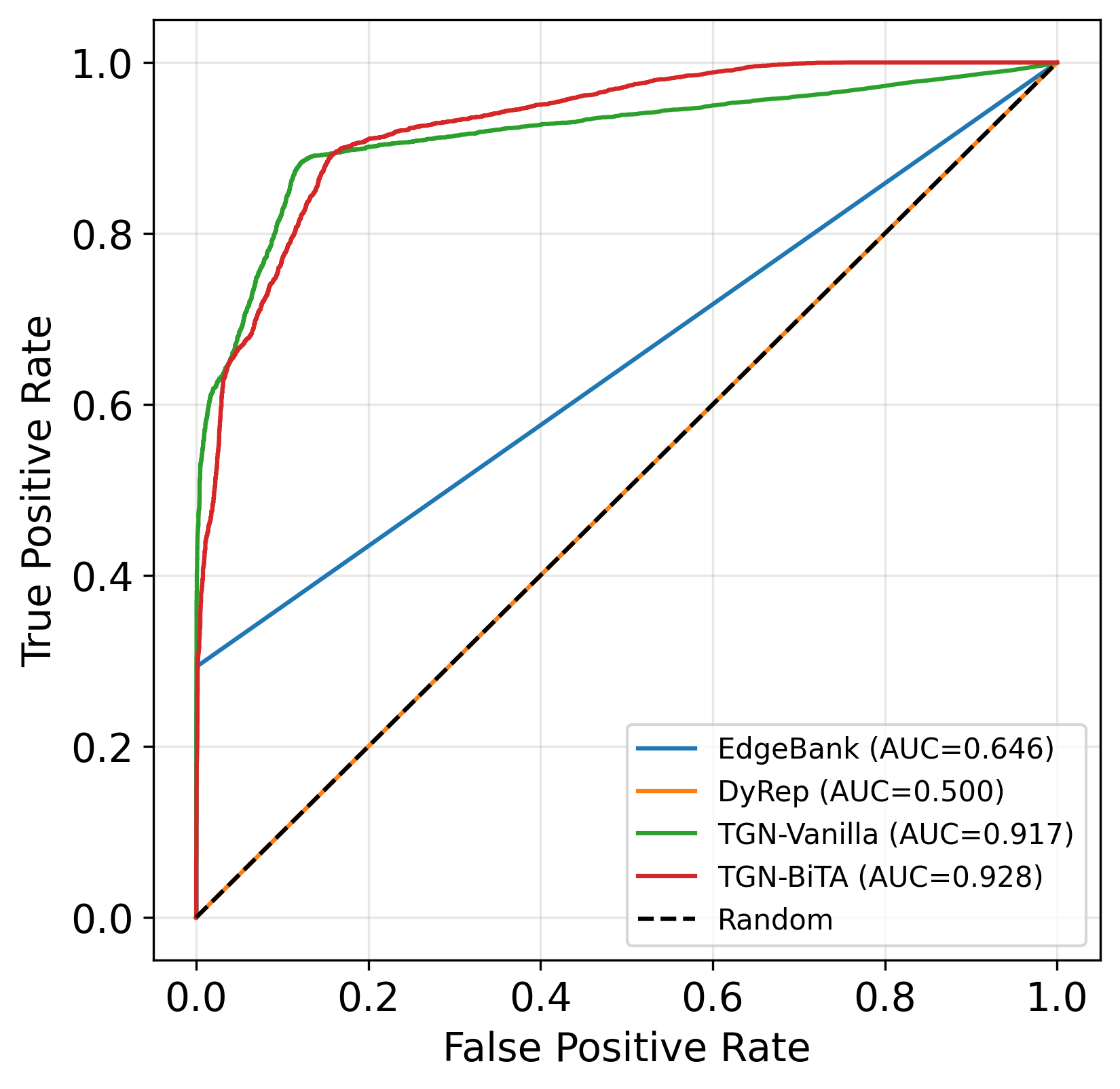}
\label{fig:subfigroc1}
}
\hspace{0.05\textwidth}
\subfloat[Last Aggregator]{
\includegraphics[width=0.4\textwidth]{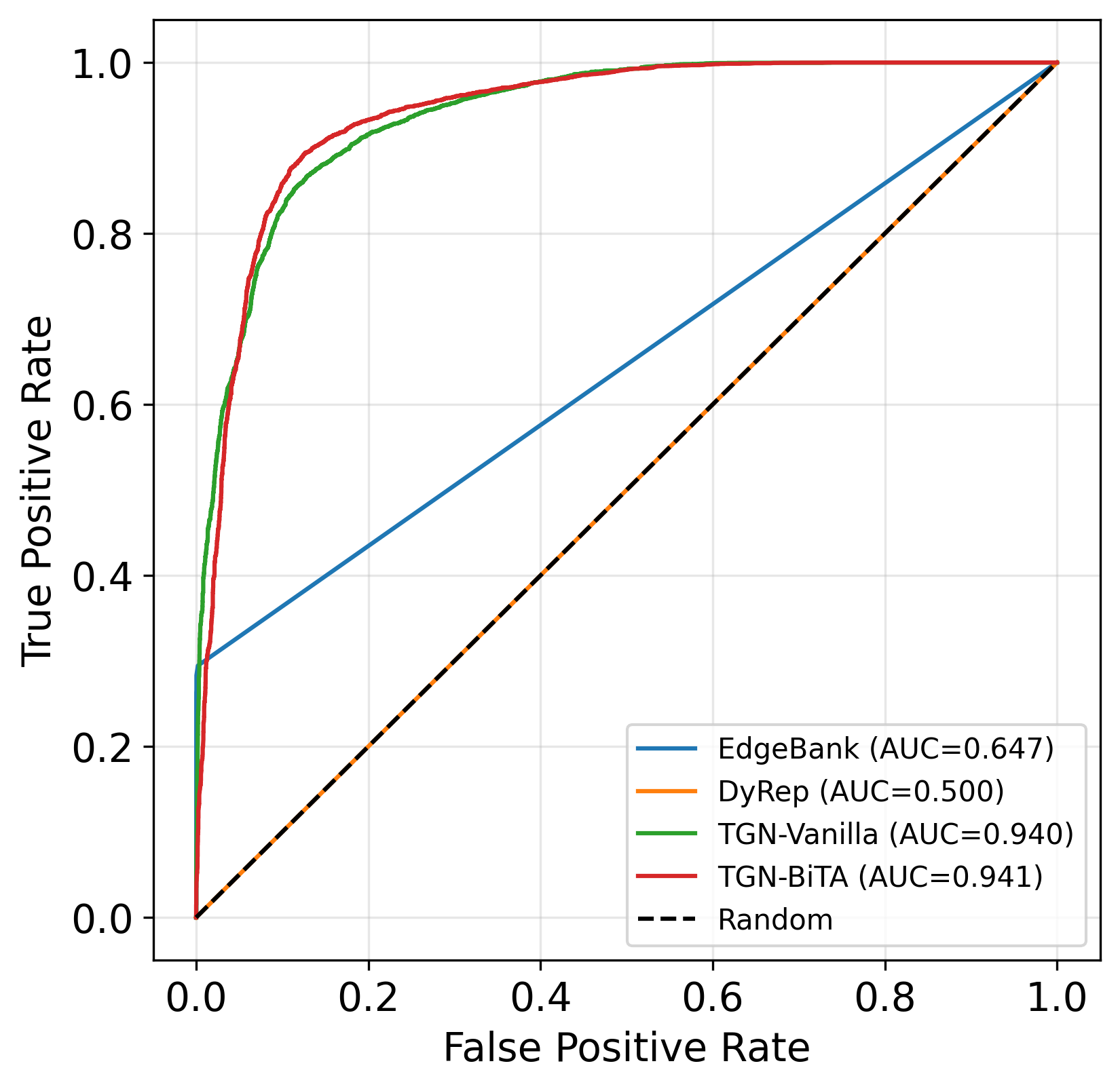}
\label{fig:subfigroc2}
}
\caption{ROC curve comparison of BiTA with SOTA methods. (a) Performance of BiTA, TGN-Vanilla with mean aggregator configuration and SOTA. (b) Performance of BiTA, TGN-Vanilla with last aggregator configuration and SOTA.}
\label{fig:rocsota}
\end{figure}

\paragraph{\textbf{Computational Efficiency:}}

Figure~\ref{fig:infertimesota} compares the inference time of BiTA against SOTA baseline methods. Computational efficiency is critical for real-world deployment, particularly in large-scale dynamic graph applications.

The results demonstrate that BiTA achieves competitive inference times despite its enhanced architectural complexity. Moreover, BiTA maintains inference times comparable to TGN-Vanilla with mean and last aggregators, confirming that the performance improvements do not come at the cost of prohibitive computational overhead.

\begin{figure}[H]
\centering
\includegraphics[width=4in]{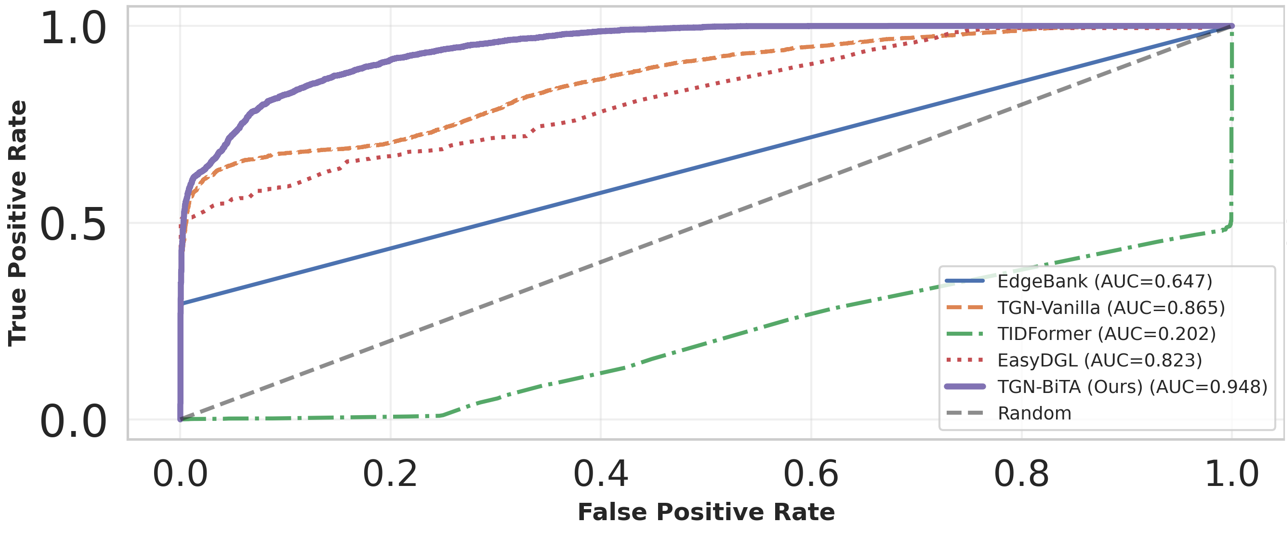}
\caption{ROC curve comparison of BiTA with SOTA baselines.}

\label{rocsota2}
\end{figure}

\paragraph{\textbf{Comprehensive Performance Metrics:}}

Figure~\ref{metricsota} presents a multi-metric performance comparison, evaluating BiTA and baseline methods across several evaluation criteria including AUC-ROC, Average Precision and F1-score. This comprehensive assessment provides insights into different aspects of model performance beyond a single metric.
BiTA consistently outperforms baseline methods across all evaluated metrics, demonstrating balanced performance in both positive and negative class prediction. 
\begin{figure}[H]
\centering
\includegraphics[width=4in]{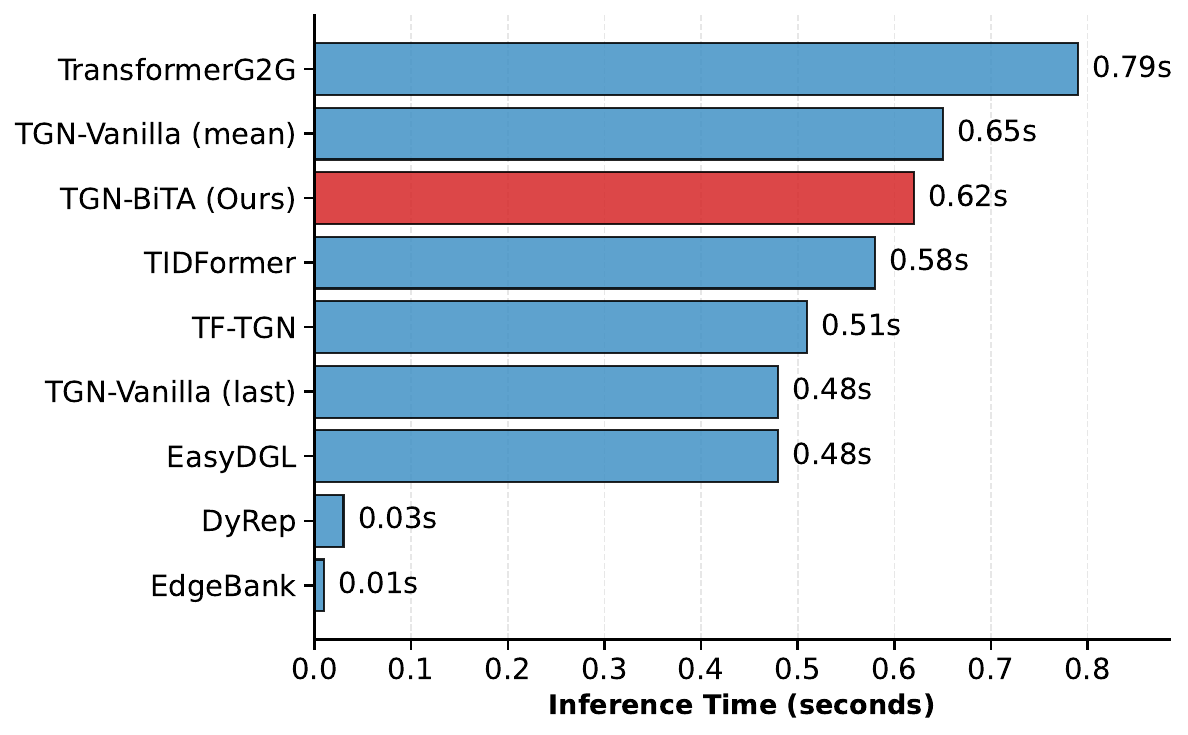} 
\caption{Inference time comparison of SOTA.}
\label{fig:infertimesota}
\end{figure}

This balanced performance is crucial for practical applications where both false positives and false negatives carry significant costs.
The ablation analysis reveals that the Transformer component contributes substantially to the overall performance gains, validating the architectural design choices and confirming that the attention mechanism effectively captures temporal dependencies in dynamic graphs.

\begin{figure}[H]
\centering
\includegraphics[width=4in]{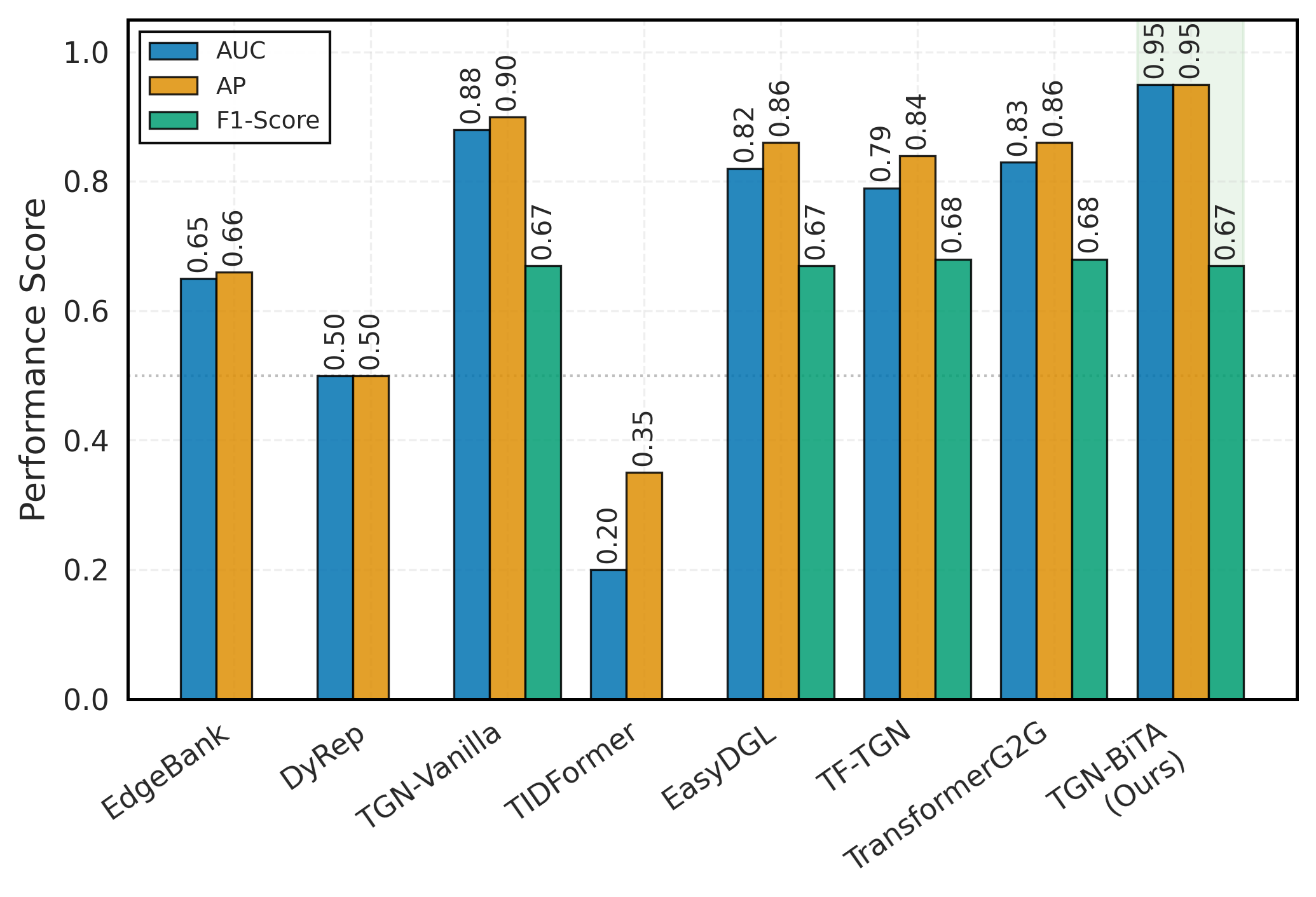}
\caption{Comprehensive performance comparison of BiTA against SOTA baselines across multiple evaluation metrics.}

\label{metricsota}
\end{figure}

\paragraph{Discussion of SOTA Comparison:}
BiTA achieves the best performance across all metrics, demonstrating the effectiveness of combining temporal modeling with attention. BiGRU substantially outperforms simple baselines, confirming the importance of sequential modeling, while attention-based aggregation provides moderate gains.

\subsection{Scalability and Runtime Analysis}

To evaluate BiTA's suitability for real-time alert prediction systems, we conducted a comprehensive scalability analysis focusing on inference latency, throughput, and robustness to graph size variations.

\paragraph{Batch Size Scalability.}
We evaluated inference performance under varying batch sizes from 100 to 500. As shown in Figure \ref{fig:scal_batch}, latency per edge decreases from 0.21~ms at batch size 100 to a minimum of 0.13~ms at batch size 128, where throughput peaks at 7,469 edges per second. Beyond this point, latency gradually increases to 0.18~ms and throughput decreases to 5,704 edges per second at batch size 500. This behavior reflects the inherent trade-off in temporal graph processing: larger batches improve computational efficiency but incur increased memory overhead and load imbalance from heterogeneous temporal neighborhoods. We identify batch size 128 as the optimal operating point.
\begin{figure}[H]
\centering
\includegraphics[width=\linewidth]{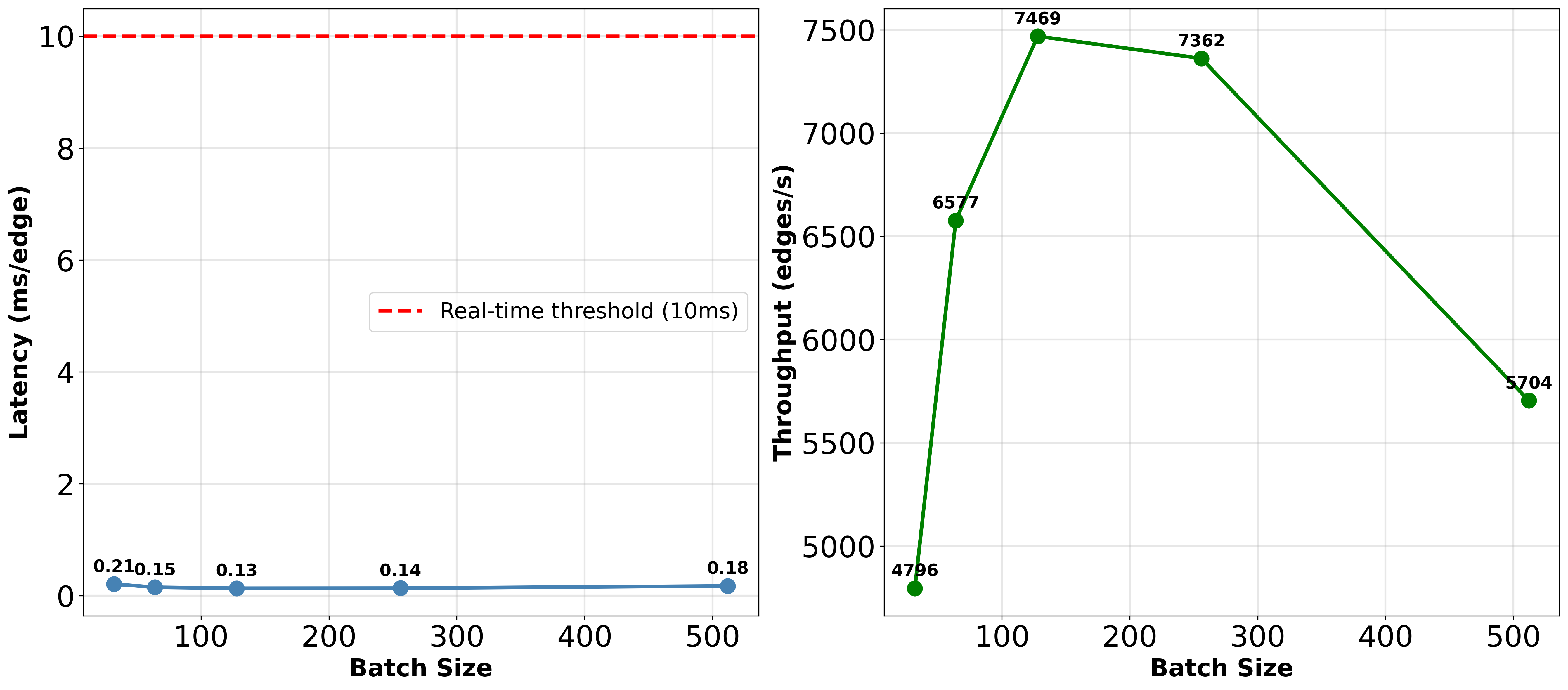}
\caption{Batch Size Scalability: Impact of batch size on latency and throughput.}
\label{fig:scal_batch}
\end{figure}

\paragraph{Graph Size Scalability.}
We assessed robustness across graphs containing 2,000 to 6,000 edges. After initial overhead at the smallest scale, per-edge latency stabilizes at 0.08--0.12~ms for graphs with 3,000+ edges (Figure \ref{fig:scal_graph}), while throughput scales to 12,368 edges per second. This stability stems from BiTA's localized temporal neighborhood sampling, which bounds computation independently of global graph size, a critical property for production deployment where graph size varies across time windows.
\begin{figure}[H]
\centering
\includegraphics[width=\linewidth]{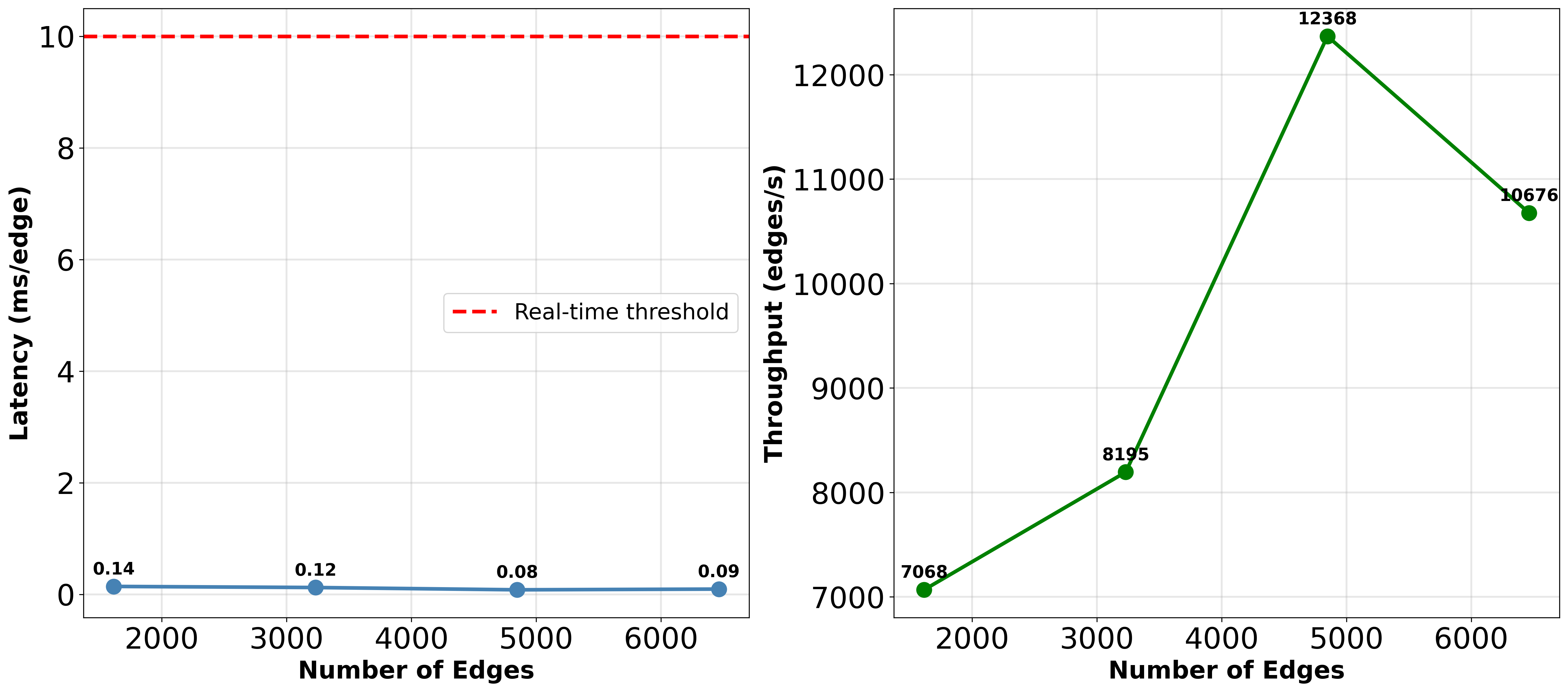}
\caption{Graph Size Scalability: Scalability with increasing graph size.}
\label{fig:scal_graph}
\end{figure}
\paragraph{Latency Distribution.}
Analysis of 500 individual predictions reveals highly concentrated latencies: mean of 2.05~ms, median of 1.97~ms, with 95th and 99th percentiles at 2.61~ms and 3.45~ms respectively (Figure \ref{fig:scal_latency}). All measurements remain well below the 10~ms real-time threshold, with no long-tail delays. At optimal configuration, BiTA processes approximately 645 million edges per day, exceeding typical enterprise security workload requirements by a significant margin. These results confirm that BiTA achieves the low-latency, high-throughput inference necessary for real-time cybersecurity applications.
\begin{figure}[H]
\centering
\includegraphics[width=\linewidth]{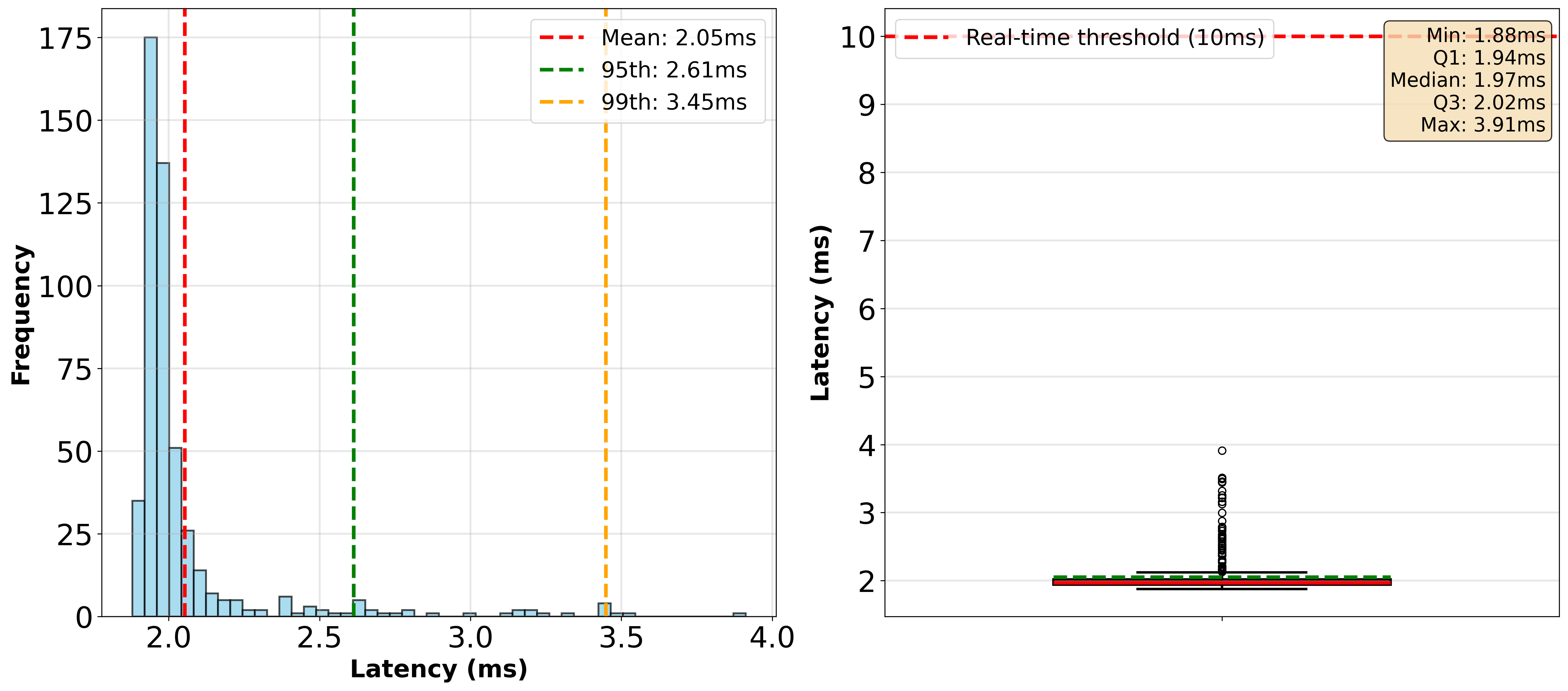}
\caption{Latency Distribution: Demonstrating real-time inference capability.}
\label{fig:scal_latency}
\end{figure}

\section{Discussion}
\label{sec:discussion}
This section discusses the advantages of the proposed BiTA framework over existing temporal graph learning methods and provides insights into the observed performance gains.
At first glance, BiTA may appear as a hybrid combination of recurrent neural networks and attention mechanisms. However, its novelty lies not in a superficial architectural composition, but in a fundamentally different \emph{temporal modeling principle} tailored to dynamic alert prediction.
Existing hybrid temporal models typically employ recurrent units or attention mechanisms to encode node embeddings or temporal snapshots independently. In contrast, BiTA introduces a \textbf{bidirectional temporal aggregation mechanism at the message level}, where temporal interactions associated with each node are modeled as ordered sequences of messages rather than isolated events. Specifically, BiTA differs from prior work in three key aspects:

\textbf{(1) Message-level bidirectional temporal reasoning.}
Unlike unidirectional temporal models that assume strictly causal evolution, BiTA allows future interactions to refine the representation of earlier messages during aggregation. This is crucial in alert prediction, where early benign interactions may only be recognized as suspicious in light of subsequent events.

\textbf{(2) Unified modeling of short-term and long-range dependencies.}
The bidirectional GRU captures local temporal continuity and short-term dependencies, while the Transformer layer selectively emphasizes temporally important interactions across the entire message sequence. These two mechanisms are jointly optimized within a single aggregation module rather than applied as separate stages.

\textbf{(3) Architectural alignment with alert prediction dynamics.}
BiTA is specifically designed for evolving alert streams with sparse, noisy, and delayed signals. The aggregation mechanism operates directly on temporally ordered alert messages, making it fundamentally different from models designed for static or snapshot-based temporal graphs.

Therefore, BiTA should be viewed not as a simple hybrid architecture, but as a principled temporal aggregation framework that enables richer temporal reasoning for dynamic alert prediction tasks.

\paragraph{\textbf{Expressive and Learnable Aggregation.}}
A key advantage of BiTA lies in replacing heuristic message aggregation functions, such as mean or last used in standard TGN, with a learnable aggregation mechanism. By modeling aggregation as a representation learning process through a BiGRU and a Transformer encoder, BiTA preserves temporal ordering and selectively emphasizes informative interactions. This substantially increases the expressive power of the aggregation stage, enabling the model to capture complex temporal patterns that are otherwise lost under permutation-invariant pooling operations.

\paragraph{\textbf{Joint Temporal and Contextual Modeling.}}
Unlike existing methods that focus primarily on local temporal dependencies or immediate neighborhood information, BiTA jointly captures bidirectional temporal dynamics and global contextual interactions among historical messages. The BiGRU encodes sequential dependencies in both forward and backward directions, while the Transformer enables long-range contextual reasoning across interactions. This combination allows BiTA to distinguish between routine activity and meaningful attack patterns, which is critical in dynamic cyber-security environments.

\paragraph{\textbf{Inductive Generalization and Robustness.}}
BiTA maintains compatibility with inductive learning settings, allowing effective prediction for nodes unseen during training. This is particularly important for real-world networks where new IP addresses continuously emerge. Moreover, the learnable aggregation mechanism reduces sensitivity to noisy or less informative interactions, leading to more stable representations compared to methods relying on fixed aggregation rules.

\paragraph{\textbf{Comparison with Recent Methods.}}
Compared to recent temporal graph models such as DyRep and EdgeBank, BiTA benefits from explicitly modeling interaction sequences at the aggregation level rather than relying on memory decay, event intensity modeling, or frequency-based heuristics. This design choice enables BiTA to better exploit temporal structure and contextual dependencies, which is reflected in its improved performance across both transductive and inductive evaluation settings.

\paragraph{\textbf{Limitations.}}
Although BiTA demonstrates favorable scalability and real-time performance in our experiments, the current evaluation is conducted under controlled batch sizes and graph scales derived from real-world alert datasets. Extreme-scale deployments, such as nation-wide monitoring infrastructures or highly adversarial environments with concept drift, may require additional system-level optimization and robustness analysis.

\paragraph{\textbf{Future Work.}}
Future research will explore adaptive aggregation depth and dynamic resource allocation strategies to further improve efficiency under extreme workload variations. In addition, extending BiTA with continual and online learning mechanisms will enable long-term deployment in evolving network environments with changing attack patterns.

\section{Conclusion}
\label{sec:conclusion}
This work introduced BiTA (Bi-Transformer Aggregator), a novel recursive temporal graph learning framework tailored for proactive alert prediction in dynamic computer networks. By integrating a temporal and contextual modeling with a recursive memory updating mechanism, BiTA effectively captures complex temporal and structural patterns inherent in cyber-attack behaviors.
Extensive experiments on two real-world alert datasets demonstrated that BiTA achieves substantial improvements over state-of-the-art temporal graph neural network models across multiple evaluation metrics, including AUC, Average Precision, MRR, and class-wise prediction metrics, under both transductive and inductive scenarios. The consistent performance gains highlight the model’s robustness, scalability, and strong generalization ability in evolving network environments.
Beyond predictive accuracy, the architecture of BiTA offers interpretability, enabling security analysts to better understand how both recent and historical interactions influence alert prediction. This makes the framework a promising candidate for deployment in real-time, adaptive intrusion detection systems.

\bibliography{Attack-predictiontest}

\newpage
\begin{biography}{Zahra Makki Nayeri}
 is a Ph.D. candidate at Shahrood University of Technology and currently a visiting researcher at the University of Stuttgart, Germany, where she conducts research on knowledge graph foundation models. Her research focuses on graph representation learning, temporal and dynamic graph neural networks, and machine learning–based modeling of computer networks, with an emphasis on data-driven analysis of large-scale, evolving interaction graphs. During her Master’s studies, she investigated machine learning techniques for fog, edge, and cloud computing environments, with a particular focus on distributed data processing, resource-aware learning, and system-level optimization.
\end{biography}

\begin{biography}{Mohsen Rezvani}
 is an Associate Professor in the Faculty of Computer Engineering at the Shahrood University of Technology. He received his Ph.D. at the School of Computer Science and Engineering at the UNSW Sydney. He holds an M.Sc. in Computer Engineering from Sharif University of Technology and a B.Sc. in Computer Engineering from Amirkabir University of Technology. His research focuses on computer security, intelligent data analysis, and advanced machine learning. He is the director of the Computer Emergency Response Team (CERT) at Shahrood University of Technology.
\end{biography}

\end{document}